\documentclass[12pt]{iopart}
\usepackage{amsmath}
\usepackage{graphicx,epstopdf}
\usepackage{caption}
\usepackage{subcaption}
\usepackage{lscape}
\usepackage{hyperref}
\usepackage{microtype}
\usepackage{color,soul}
\usepackage[export]{adjustbox}
\usepackage[table,xcdraw]{xcolor}

\usepackage{placeins}

\graphicspath{ {./figs/} }

\DeclareMathAlphabet{\mathbbold}{U}{bbold}{m}{n}

\usepackage{iopams}

\begin{document}

\title[Temporal Basis Function Models]{Temporal Basis Function Models for Closed-Loop Neural Stimulation}
\author{Matthew J Bryan$^{1,2,3}$, Felix Schwock$^{2,3,5}$, Azadeh Yazdan-Shahmorad$^{2,3,4,5}$, Rajesh P N Rao$^{1,2,3}$}

\address{$^{1}$ Neural Systems Laboratory, Paul G. Allen School of Computer
Science \& Engineering, University of Washington, Seattle, WA, USA}
\address{$^{2}$ Center for Neurotechnology, University of Washington, Seattle, WA, USA}
\address{$^{3}$ Computational Neuroscience Center, University of Washington, Seattle, WA, USA}
\address{$^{4}$ Department of Bioengineering, University of Washington, Seattle, WA, USA}
\address{$^{5}$ Department of Electrical and Computer Engineering, University of Washington, Seattle, WA, USA}

\ead{\{mmattb,rao\}@cs.washington.edu}
\vspace{10pt}
\begin{indented}
\item[]May 2025
\end{indented}

\begin{abstract}
\textit{Objective} Closed-loop neural stimulation provides novel therapies for neurological diseases such as Parkinson's disease (PD), but it is not yet clear whether artificial intelligence (AI) techniques can tailor closed-loop stimulation to individual patients or identify new therapies. Further advancements are required to address a number of difficulties with translating AI to this domain, including sample efficiency, training time, and minimizing loop latency such that stimulation may be shaped in response to changing brain activity.
\textit{Approach} We propose temporal basis function models (TBFMs) to address these difficulties, and explore this approach in the context of excitatory optogenetic stimulation. We demonstrate the ability of TBF models to provide a
single-trial, spatiotemporal forward prediction of the effect of optogenetic stimulation on local field potentials
(LFPs) measured in two non-human primates. We further use simulations to demonstrate the use of TBF models for closed-loop stimulation, driving neural activity towards target patterns.
The simplicity of TBF models allow them to be sample efficient, rapid to train ($<$$5min$), and low latency ($<$$0.2ms$) on desktop CPUs.
\textit{Main results} We demonstrate the model on $40$ sessions of previously published excitatory optogenetic stimulation data. For each session, the model required $<$$20$ minutes of data collection to
successfully model the remainder of the session. It achieved a prediction accuracy comparable to a
baseline nonlinear dynamical systems model that requires hours to train, and superior accuracy to a linear state-space model requiring 90 minutes to train. In our simulations, it also successfully allowed a closed-loop stimulator to control a neural circuit.
\textit{Significance} By optimizing for sample efficiency, training time, and latency, our approach begins to bridge the gap between complex AI-based approaches to modeling dynamical systems and the vision of using such forward prediction models to develop novel, clinically useful closed-loop stimulation protocols. 
\end{abstract}


\maketitle

\section{Introduction}
\label{sec:introduction}

Advances in neural stimulation over the past few decades have allowed targeted restoration of sensory and motor function, ranging from cochlear and retinal implants to restore hearing and vision, to deep brain stimulation to reduce symptoms of Parkinson's disease (PD) \cite{niparko.cochlear, weiland.retinal, tomlinson.propr, tabot.tact, tyler.tact, dadarlat.tact, sharlene.tact, cronin.tact}. Closed-loop stimulation uses sensed brain activity and other sensor data in a feedback loop to adapt stimulation for applications such as fine-grained real-time control of neural circuits and prosthetic devices (e.g., \cite{nicolelis.bmbi, bryan.coproc, bolus.opto, kahana.biomarker, berger.closedloop, tafazoli.acls}). This feedback loop provides increased efficacy, energy efficiency \cite{castano.pd}, and reduced side effects \cite{little.park} compared to open-loop stimulation. Additionally, evidence suggests that stimulation responses can be dependent on the neural activity at the time stimulation is applied - a property we call ``state-dependence'' \cite{bradley.statedep, zanos.closedloop}. State-dependence can appear in neural responses \cite{kabir.statedep}, behavioral responses \cite{bradley.statedep}, and plasticity effects \cite{bloch.statedep, zanos.beta}. As a result, modeling stimulation responses as a function of measured brain activity may enable more effective and efficient stimulation-based therapies in the future.

The use of AI and artificial neural networks (ANNs) to enable adaptive closed-loop stimulation - a concept known as ``neural co-processing'' \cite{rao.braincoproc, rao.coproc, bryan.coproc} - may allow us to better tailor stimulation to individual diseases and patients. Neural co-processors seek to couple ANNs to biological neural networks in the brain, leveraging machine learning to map complex neural activity patterns to stimulation which drives neural activity or behavioral outcomes to desired regimes for therapy and rehabilitation.

However, significant barriers exist for the implementation of such ML-based approaches to neural stimulation. One barrier is {\em sample efficiency}, which is defined as the quantity of data required for a model to learn effectively. Time and safety considerations limit the quantity of stimulation we can attempt on a subject's brain, and consequently any co-processor must be designed for sample efficiency, for example by leveraging low dimensional structures in the neural activity like neural manifolds \cite{gallego.manifold}. A second barrier is {\em training efficiency}. A learning algorithm which requires many hours or days of training time will significantly restrict the range of experiments and treatments it could enable. Finally, computation-intensive methods can exhibit a high {\em execution latency}, causing a large gap in time between the sensing of brain activity and applying stimulation in response to that activity. That may in turn inhibit effective shaping of the activity or behavior due to the limited predictability of the brain: closed-loop stimulation may perform no better than open-loop if the controller's latency exceeds the forecast horizon in which neural activity can be reasonably predicted \cite{furht.realtime, huang.realtime}.

We propose a new approach to closed-loop neural stimulation which attempts to overcome these barriers in the context of optogenetic stimulation \cite{azadeh.data}. We present:
\begin{enumerate}
    \item A spatiotemporal forward predictive model which predicts the state-dependent effect of stimulation on neural activity while accounting for the stimulation loop's latency. We provide a $PyTorch$ implementation and demo code, available under an open source license (see `Code and data availability' Section \ref{sec:cad}).
    \item Two example controllers which leverage the forward model to determine an optimal stimulation policy given target neural dynamics, showing in simulation that our model can successfully shape neural activity (Section \ref{sec:methods-demo1}, \ref{sec:methods-demo2})
    \item Comparison of our model to a more complex nonlinear dynamical systems model, as well as a simpler linear state space model (LSSM). We show that our model's prediction accuracy exceeds that of the LSSM model, and surprisingly also that of the more complex model as well, across existing data for $40$ experimental sessions involving two rhesus macaque monkeys. At the same time, it exhibits significantly better sample efficiency, training time, and latency. On average it requires $<$20min of data collection and $<$5min of training time on our data; and a single prediction requires $<$0.2ms on a commodity CPU.
    \item Statistical evidence showing that the excitatory optogenetic stimulation response is state-dependent in this setting (Section \ref{sec:results.state}).
    \item Evidence that state-dependent models perform better on these data.
\end{enumerate}

\section{Background}
\label{sec:background}

Model-based approaches to closed-loop stimulation involve building a ``forward model'' that predicts the effect of the stimulation, and using that model to specify an AI agent which controls the stimulation. When that model is built using artifical neural networks (ANNs), it is atype of brain-computer interface called a neural co-processor \cite{rao.coproc,rao.braincoproc,bryan.coproc, pan.coproc}. The forward model may predict external behavioral variables such as hand position, gait or speech, or internal variables such as brain activity or other physiological measurements. Given a forward model, at least two approaches are available: 1.) model-based control, such as model predictive control (MPC) \cite{xu.control} or learning an inverse model \cite{bryan.coproc}; and 2.) model-based reinforcement learning (MBRL) \cite{barto.rl}. In both cases we may impose cost or reward functions which balance goals such as minimizing disease symptoms, prolonging implant battery life, and reducing side effects. Such model-based approaches are sometimes favored for their sample efficiency \cite{barto.rl, pan.coproc}, which is one of the reasons we leverage them here. They also tend to generalize better across tasks, and can more easily be designed for safety \cite{moerland.mbrl}. The choice of model depends drastically on the goals and structure of the problem, and involves trade-offs between interpretability, robustness, memory efficiency, sample efficiency, and other concerns. 

Some existing approaches to stimulation modeling leverage detailed mathematical simulations of cellular biophysics, for example, The Virtual Electrode Recording Tool for EXtracellular Potentials (VERTEX) model \cite{vertex.stim}. These detailed models can recreate some features of brain activity and stimulation response, but tend not to be usable for network-level modeling of individual subjects data due to scalability and difficulties fitting the model.

Other existing forward models of neural stimulation consider only the stimulation parameters and not the initial brain state (i.e. state-dependence). For example in Yang et al.'s approach to model-based control of stimulation \cite{shanechi.stimmodel}, the stimulation neural dynamics in rhesus macaques is estimated from randomized electrical stimulation over multi-second time scales. While their model is a linear state space model (LSSM) and therefore involves a latent state, that state is zero-initialized and the forecast is updated according to the stimulation parameters alone. Such an approach measures the ``input-driven dynamics'' but ignores the effect of state-dependence - a reasonable approach since intrinsic neural dynamics are likely difficult to forecast over multiple seconds. The authors additionally provide a simulated demonstration of their model for closed-loop control. Here their model provides single-step predictions and leverages a linear quadratic regulator (LQR) approach to control but does not provide a multi-step prediction.

\section{Methods}
\label{sec:methods}

\begin{figure}
	\centering
	\begin{subfigure}[c]{0.99\textwidth}
		\centering
		\includegraphics[width=\textwidth]{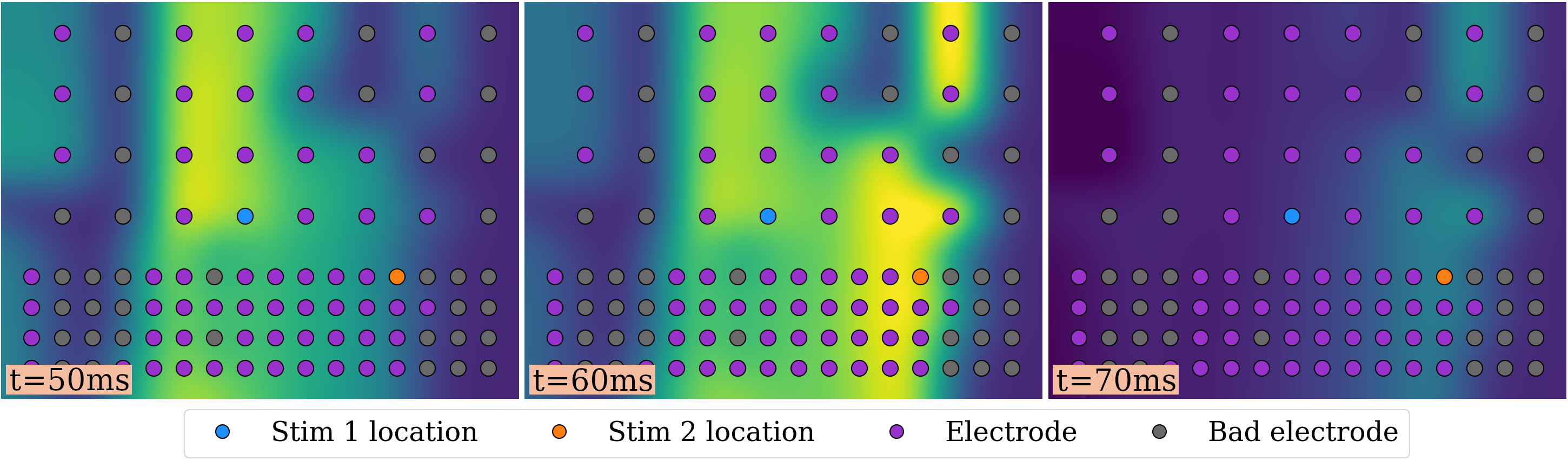}
		\caption{}
	\end{subfigure}
	\hfill
  	\begin{subfigure}[c]{0.33\textwidth}
		\centering
		\includegraphics[width=\textwidth]{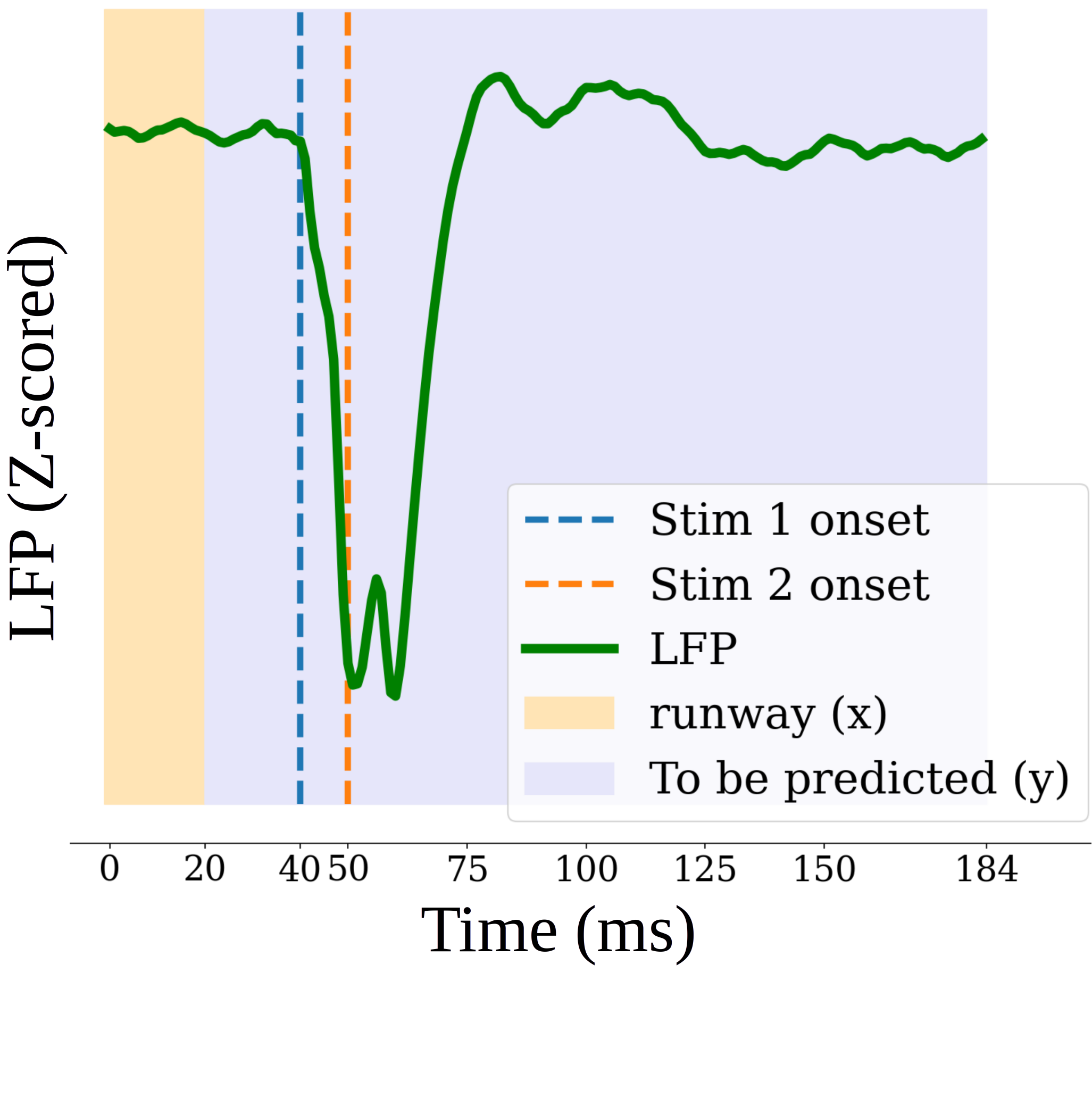}
		\caption{}
	\end{subfigure}
	\hfill
      	\begin{subfigure}[c]{0.64\textwidth}
		\centering
		\includegraphics[width=\textwidth]{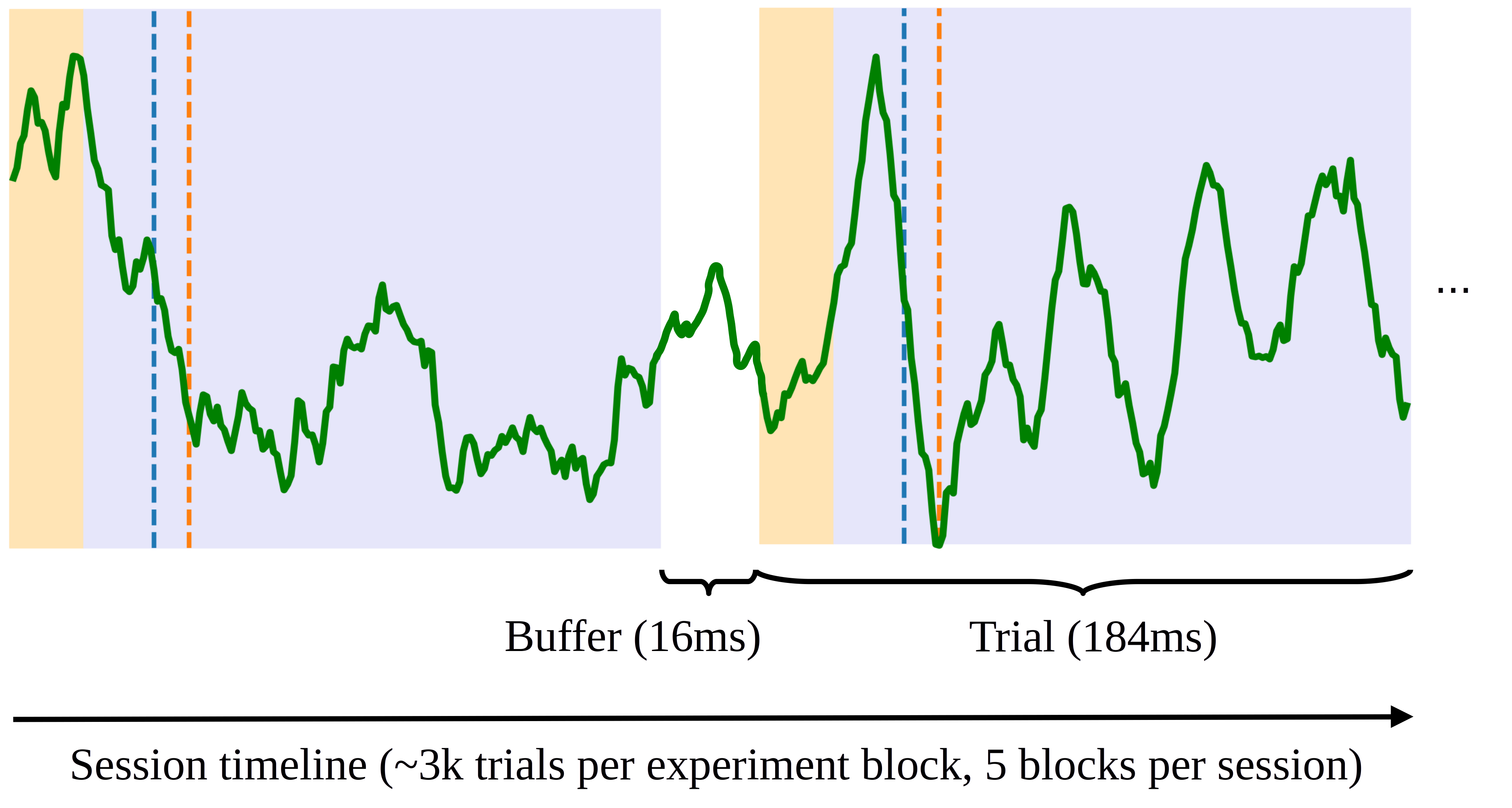}
		\caption{}
	\end{subfigure}
	\hfill
\caption{\small \textbf{Spatiotemporal electrocorticographic (ECoG) responses after paired pulse optical stimulation}. (a) \textbf{Dynamics of ECoG responses after stimulation.} The three panels show spatially-smoothed time domain ECoG responses across the surface of monkey sensorimotor cortex (S1, M1), measured using an array of $\mu$ECoG electrodes (circles). The precise placement of the array on the cortex varied between sessions. The ECoG responses in the three panels were computed as values Gaussian-smoothed across the electrodes after the first (t=$50ms$) and second (t=$60ms$) stimulation pulses, and at a time later in the trial (t=$70ms$). Some electrodes were not usable in a given session (dark grey circles). These electrodes were assumed to have values of 0 for the plots.
(b) \textbf{ECoG response for a single electrode and the forward prediction problem}. Trial-averaged ECoG response over time. Our model forward predicts the single-trial response based on a $20ms$ window of input data ($\boldsymbol{x}$, the orange region) which we term the ``runway.'' The time range of the model's prediction ($\boldsymbol{y}$) is depicted as the blue region which starts $20ms$ before the first pulse and extends $164ms$ in legth. The ECoG recording shown is from the site of the first stimulation pulse (Stim 1 location) in (a). This $184ms$ timespan is referred to as a ``trial.''
(c) \textbf{Session timeline} Each trial's beginning/end are always positioned such that the first pulse onset is at t=$40ms$. Note this subfigure depicts a single channel (not trial-averaged), but the trial involves all usable channels. A buffer between trials allowed us to experiment with runway length, but isn't integral to our design. A typical session contains $\approx$$15k$ trials.
}
\label{fig:trial}
\end{figure}

\subsection{Optogenetic stimulation dataset}
We demonstrate our approach on a previously-published \cite{azadeh.data} micro-electrocorticography ($\mu$ECoG) dataset from 40 optogenetic stimulation sessions involving two non-human primates (NHPs; rhesus macaques). Excitatory neurons in the left primary somatosensory (S1) and motor (M1) cortices were optogenetically modified to express the red-shifted channelrhodopsin, C1V1. A semi-transparent 96 channel $\mu$ECoG array allowed for measurement of cortical activity while paired pulses of optical stimulation were delivered by two fiber optics (Figure \ref{fig:trial}(a)) \cite{azadeh.data2, azadeh.data3, azadeh.data4, azadeh.data5}. The fiber optic probes were co-located with two different ECoG electrodes. The choice of optical stimulation locations varied across sessions but were constant within a session. In any given session some number of electrodes were unusable for various reasons (mean $78.6$, stdev $13.2$). Once recorded, data were filtered for line noise. See Appendix \ref{apx:dataset} for additional details. We refer to these data as local field potentials (LFPs) throughout, since these $\mu$ECoG electrodes are thought to be small enough to measure LFPs \cite{bloch.opto}.

For a given session, pairs of optical stimulation pulses were delivered every $200ms$ across five 10 minute blocks, for a total of 15k pairs. Each pulse had a $5ms$ duration. The two pulses were spaced apart either $10, 30,$ or $100ms$ and this interval varied from session to session. The five stimulation blocks were interleaved with 10 minute resting blocks during which no stimulation was applied.

We primarily present results below for time domain data with no additional filtering applied except noise removal, and for beta bandpassed time domain data. Time domain neural data has previously been used to extract important and neurologically relevant features such as event-related potentials (ERPs) and low frequency ($<0.5hz$) local motor potentials (LMPs) from the motor cortex \cite{flint.lmps}. ERPs are associated with myriad phenomena such as attention and perception, and often exhibit low frequency structure operating over hundreds of milliseconds, such as the well-known P300. They have also been shown to be state-dependent in some settings \cite{busch.erpstatedep}. LMPs estimated from LFPs have been shown to be predictive of motor movement \cite{flint.lmps}. While we may apply low-pass filters to isolate these low frequency phenomena, such filters can add significant latency to the controller - an issue we are attempting to mitigate. Instead, we will rely on the well-known 1/f power scaling phenomena \cite{bedard.powerlaw} which causes our model to mostly focus on lower frequencies (analyzed further in Appendix \ref{apx:lowfreq}). We additionally present results below for beta-bandpassed (13-30hz) data. The bandpass was performed using a first order Butterworth filter. Beta time domain data has been used to decode both Parkinsonian symptoms and motor movement\cite{papadopoulos.betaburst}. Results for additional signal features such as time-frequency domain and bandpassed data are presented in Appendix \ref{apx:filtering} but omitted here for brevity.

\subsection{Data windowing}
\label{subsec:datawindowing}
For each experimental block we window the session timeline into $184ms$ trials (see Figure \ref{fig:trial}). The experiments provided $200ms$ between pulse pairs, and trials $184ms$ in size allowed us a buffer to easily experiment with window size. For each trial the first stimulation pulse begins at t=$40ms$. In a typical session this resulted in $\approx$15k such windows, or $\approx$7.5k for shorter sessions. To compare stimulation and non-stimulation conditions we also randomly crop $184ms$ windows of data from non-stimulation blocks.

Our model leverages the first $20ms$ of data to forward predict the rest of the trial. The first stimulation pulse always has onset at t=$20ms$ into the predicted portion of the trial to compensate for a $20ms$ loop latency. A $20ms$ latency resembles a reasonable worst-case latency relative to some other existing closed-loop systems; e.g. \cite{guggenmos.latency} which targets a $7.5ms$ latency. Similarly, \cite{zanos.beta} specifically measures and accounts for a control loop latency of $3$-$5ms$. We train a model for a specific loop latency to represent a worse case, and note that the system designer can insert artificial delays to account for faster systems. See Figure \ref{fig:trial}(b) for a visual depiction of stimulation and prediction timing. 

\subsection{Temporal basis function model}
\label{methods:tbfm}
The temporal basis function model (TBFM) provides a single-trial multi-step spatiotemporal forward prediction of the effect of neural stimulation applied in the future, given the most recently measured brain state (here, LFP values). In our dataset, stimulation was applied at fixed electrode sites, so the main degree of freedom for varying stimulation parameters is the timing of the stimulation relative to brain activity. The prediction extends to a fixed horizon $T_h$. Shorter horizons tend to lead to improved accuracy, and are sufficient for some tasks (see Section \ref{sec:methods-demo1} for an example).

\begin{figure}
    \raggedleft
    \includegraphics[width=\textwidth]{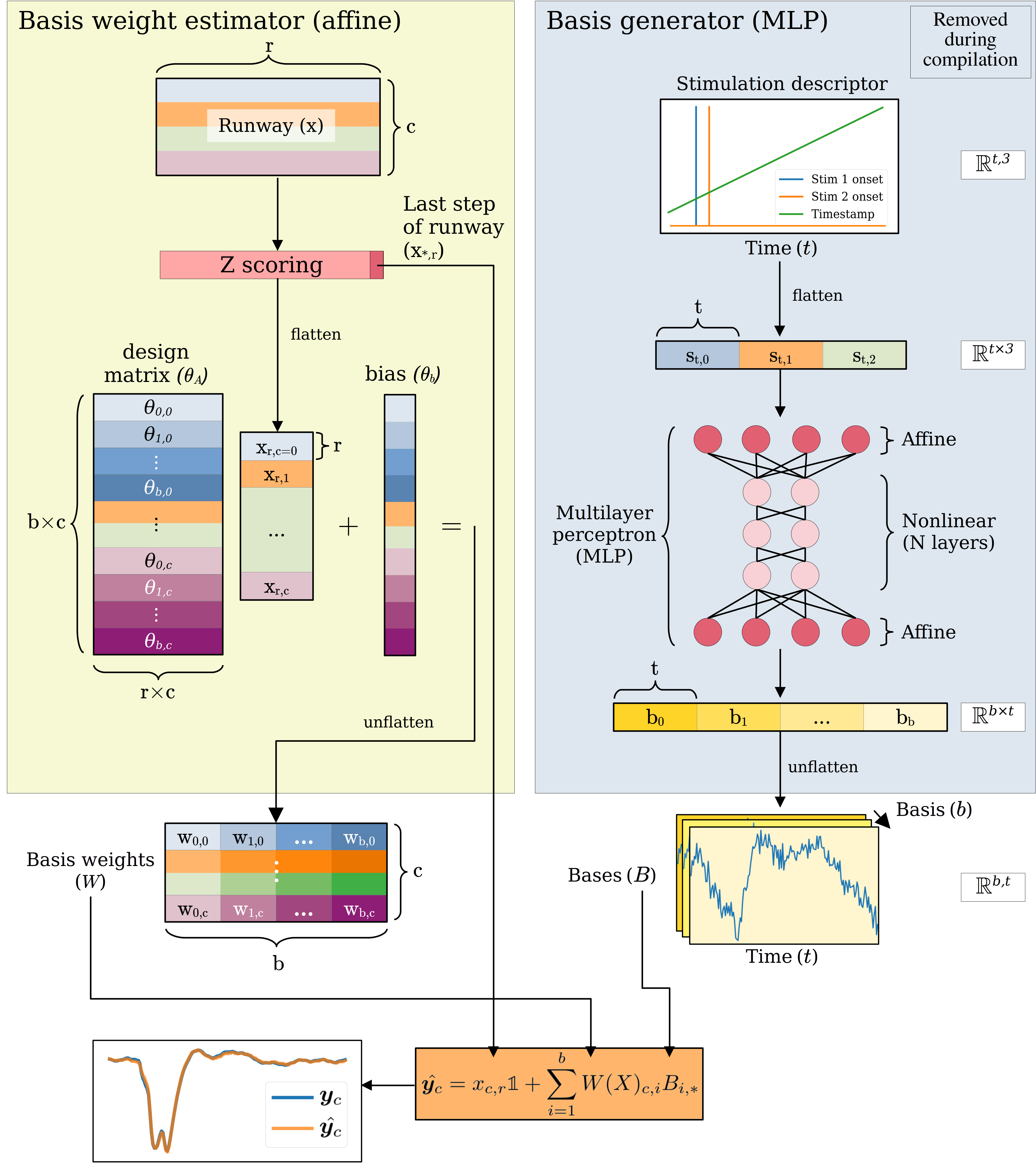}
    \captionsetup{singlelinecheck=false,format=hang,justification=raggedright}
    \caption{\textbf{Architecture of the temporal basis function model.}}
    \noindent\begin{minipage}{\textwidth}
        \small
        The model predicts the response $\boldsymbol{y}_c$ of a channel $c$ to stimulation using a weighted sum of temporal basis functions and the last
        measurement $x_{c,r}$ made before prediction. $r$ here is the runway length.
        A multilayer perceptron (MLP) generates $b$ temporal bases from a descriptor of the stimulation parameters (known as the ``stimulation descriptor''). The bases have length equal to the length of the prediction horizon $T_h$. 
        The weights $W$ for linearly combining the bases are estimated by an affine function using $20ms$ of data from before the prediction (known as the ``runway'' $X$). 
        The weights for a channel are estimated using the
        runway of all channels. The weights and bases are estimated 
        jointly using gradient descent to minimize a prediction error function (see Section \ref{sec:training} for training details).
    \end{minipage}
\label{fig:tfm_arch}
\end{figure}

The general architecture of the TBFM is illustrated in Figure \ref{fig:tfm_arch}. We describe the training procedure of the model in Section \ref{sec:training}, but first outline its architecture here. Additional details for each aspect of the architecture are in Appendix \ref{apx:arch} and a $PyTorch$ implementation of the temporal basis function model is available for download (see `Code and data availability' Section \ref{sec:cad}). The model consists of:

\begin{enumerate}
    \item \textbf{A ``stimulation descriptor'' of length $T_h$ which encodes the time varying stimulation parameters}. In our case this is a matrix $\mathbb{R}^{T_h,3}$ which encodes the timing of the two pulses, since no other parameter is varied. In future experiments where stimulation parameters such as pulse width, location, and timing vary this descriptor would be adapted to encode those parameters as well.
    
    \item \textbf{A basis function generator, which in our case is a multi-layer perceptron (MLP)}. It receives as input the stimulation descriptors, and outputs $B \in \mathbb{R}^{b,T_h}$: a matrix of temporal basis functions. Each row $B_{i,*} \in \mathbb{R}^{T_h}$ contains one temporal basis function. The number of basis functions ($b$) is a hyperparameter - usually 12-15, chosen in a manner similar to principal component analysis (PCA); see Section \ref{sec:additive} for details. Note that our model outputs the entire forward prediction as a single output, rather than using the slow process of ``unrolling'' through time as in recurrent neural networks (RNNs) and other recurrent models.
    
    \item \textbf{The ``runway $(x)$'' which is the portion of LFP data prior to the time point from which we forward predict}. Thus $x \in \mathbb{R}^{c,r}$ where $c$ is the number of usable channels and $r$ is the runway length.
    
    \item \textbf{Per-channel means and variances which are used to Z-score the incoming runway of data}. We calculate these from the training dataset.
    
    \item \textbf{An affine basis weight estimator}. The estimator receives the Z-scored runway as input and outputs a set of channel specific and time invariant weights $W \in \mathbb{R}^{c,b}$ for the basis functions. Note that while the weights are time invariant for a given trial, they vary between trials since they depend on the runway of that trial. We found that a nonlinear basis weight estimator did not significantly improve performance in our case but TBFMs can use a nonlinear function here.
    
    \item \textbf{Calculating the forward predictions for each channel $\hat{\boldsymbol{y}_c} \in \mathbb{R}^{T_h}$} as the weighted sum of temporal basis functions and the last LFP measurement for that channel $x_{c,r} \in \mathbb{R}$. For a single channel that becomes: $\hat{\boldsymbol{y}_c} = x_{c,r} \mathbbold{1} + \sum_{i=1}^{b} W(X)_{c,i} B_{i,*}$. We repeat that calculation across all channels to obtain the full spatiotemporal forward prediction.
\end{enumerate}

\subsection{Training method}
\label{sec:training}
The basis generator and basis weight estimator are learned jointly using supervised learning, error backpropagation, and stochastic gradient descent. First, we calculate the per-channel means and variances from a training dataset and Z-score that set. We retain those means and variances for unseen examples. Our loss function is based on $\mathcal{L}_2$ prediction error: $\mathcal{L}(\theta) = \mathcal{L}_2(\hat{y}, y) + \lambda ||\theta_A||_F$. The second term is a regularizing term based on the Frobenius norm, applied to the parameters of the weight estimator's design matrix $\theta_A$. We found this regularization decreased overfit of our model. Here $\lambda \in \mathbb{R}$ is the scalar for our regularizer.

\begin{figure}
	\centering
  	\begin{subfigure}[c]{0.99\textwidth}
		\centering
		\includegraphics[width=\textwidth]{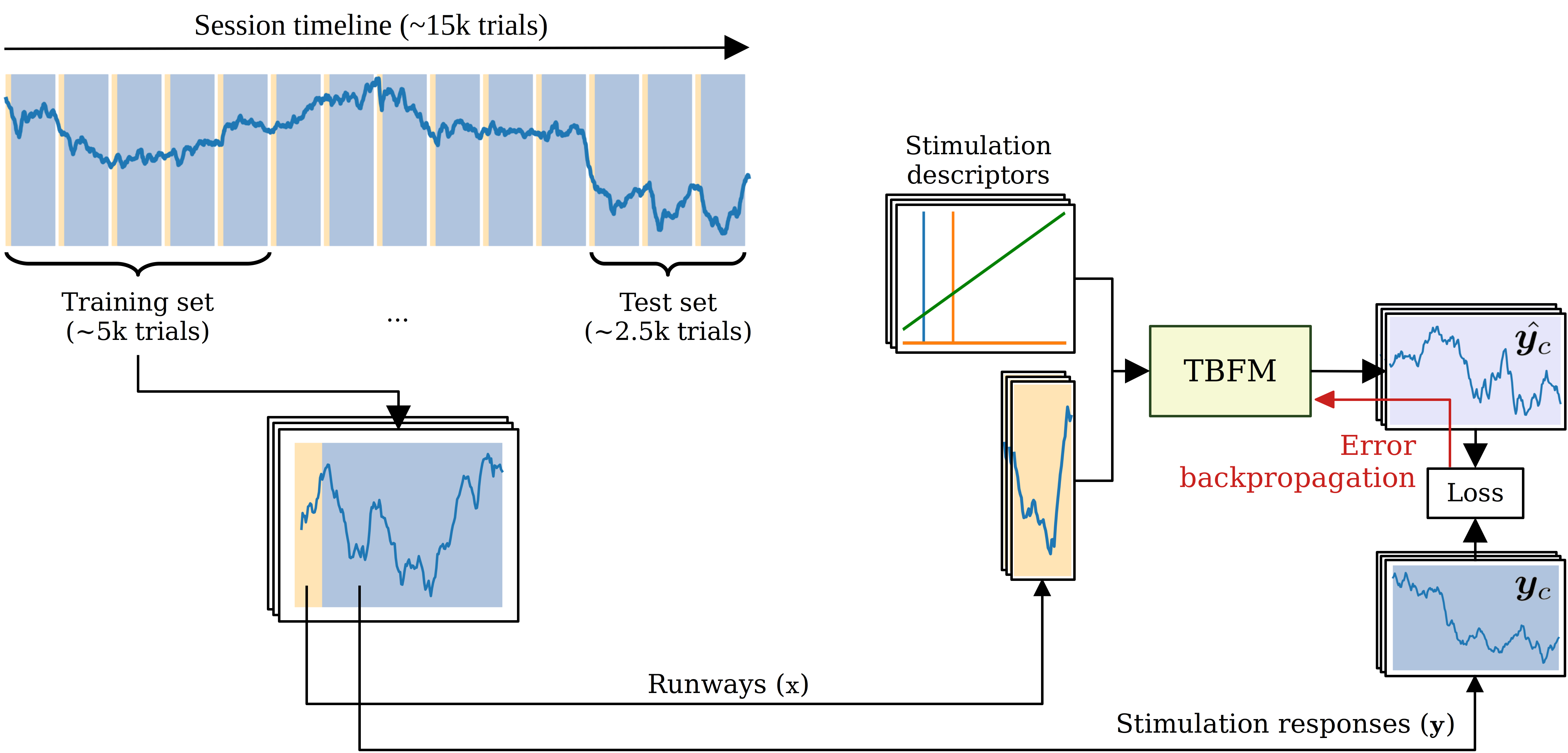}
	\end{subfigure}
	\hfill
\caption{\small \textbf{Training procedure.} Training set trials are sourced from early in the session, to simulate experimental conditions. The runways are passed to the Temporal Basis Function Model (TBFM) along with the stimulation descriptors. The TBFM predicts the local field potentials (LFPs) from the latter part of the trial. Note we depict individual channels here, but the TBFM is jointly estimating all channels for a given trial; i.e. it is making a spatiotemporal prediction.
}
\label{fig:training}
\end{figure}

We train the model with batches of training set data, consisting of trials from early in each session (Figure \ref{fig:training}). By sourcing from the first trials in the session, we can test the model's ability to generalize to trials later in the session. This is important since it mocks experimental conditions where we would gather data early in the session and then validate the model and controller later. We run the training until training loss no longer decreases (usually 15k training epochs). Additional details on the train/test split can be found in Appendix \ref{apx:dataset}.

Each trial in the batch is accompanied with a stimulation descriptor. For the results presented in this paper the stimulation descriptor will be identical for all batch elements within a given session, since each model will be trained on trials containing the same number of stimulation pulses, as explained above (Section \ref{methods:tbfm}). However, that constraint is not fundamental to TBFMs and can be relaxed.

\subsection{Forward stagewise additive modeling}
\label{sec:additive}
An additional method to learn the temporal basis functions is forward stagewise additive modeling (FSAM) \cite{murphy.ml}, which involves adding one basis function at a time and while we continue to optimize the basis weights. Our approach differs from traditional FSAM in that we don't freeze the weights for bases previously added. Similar to principal component analysis (PCA), this gives us a method by which to pick the number of basis functions we will retain: we add bases until accuracy on a validation set reaches a sufficient level or is no longer increasing. As we show in Section \ref{sec:results.additive} this can also cause the basis functions to be more interpretable. Additional technical details on FSAM can be found in Appendix \ref{apx:additivebasis}.

\subsection{The closed-loop controller}
\label{methods:controller}

We demonstrate the model's use for closed-loop neural stimulation using two simulated tasks based on our previously collected data: 1) applying stimulation only when certain brain states occur immediately before stimulation onset; and 2) stimulating to shape brain activity toward target trajectories. These simulations illustrate how a controller could be built from a TBFM, and that the TBFM's accuracy is sufficient for controllability.

Because our dataset came from stimulation experiments that did not vary stimulation parameters within a given session, we focus on simulated controllers which determine the timing of stimulation to achieve the goals in 1) and 2). In both tasks each trial presents a $20ms$ ``runway'' of data to the controller, and the controller decides whether or not to stimulate in order to optimize for the target outcome. We present additional technical details of the controller design and simulations in Appendix \ref{apx:demos}.

\subsubsection{Demonstration 1: stimulating based on predicted target brain states.}
\label{sec:methods-demo1}
Our first controller attempts to target stimulation to key events in the features of upcoming neural activity while countering the effects of latency. This resembles the task faced by some existing closed-loop controllers which attempt to trigger stimulation on particular neural features, e.g., particular beta band phases \cite{zanos.beta} or neural markers of tremor \cite{castano.pd}.

\begin{figure}
	\centering
     \begin{subfigure}[c]{0.49\textwidth}
		\centering
		\includegraphics[width=\textwidth]{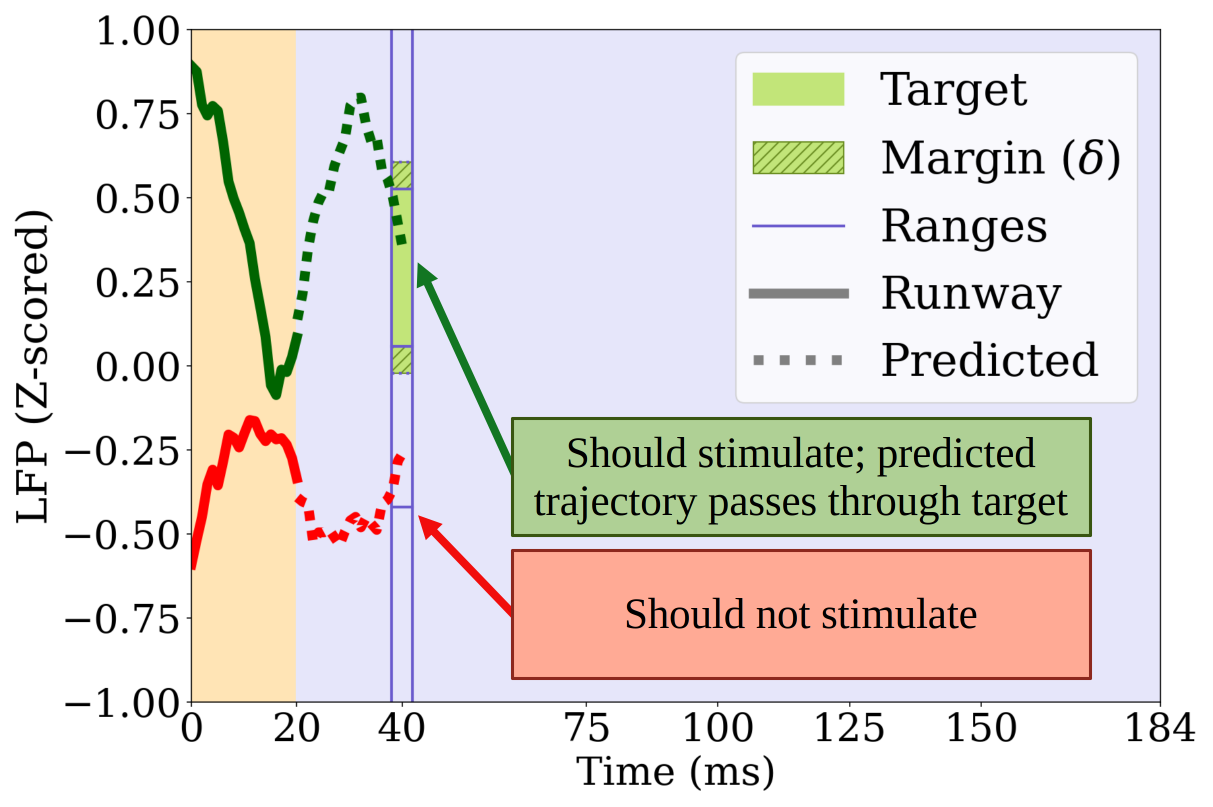}
		\caption{Demonstration 1}
	\end{subfigure}
	\hfill
   	\begin{subfigure}[c]{0.49\textwidth}
		\centering
		\includegraphics[width=\textwidth]{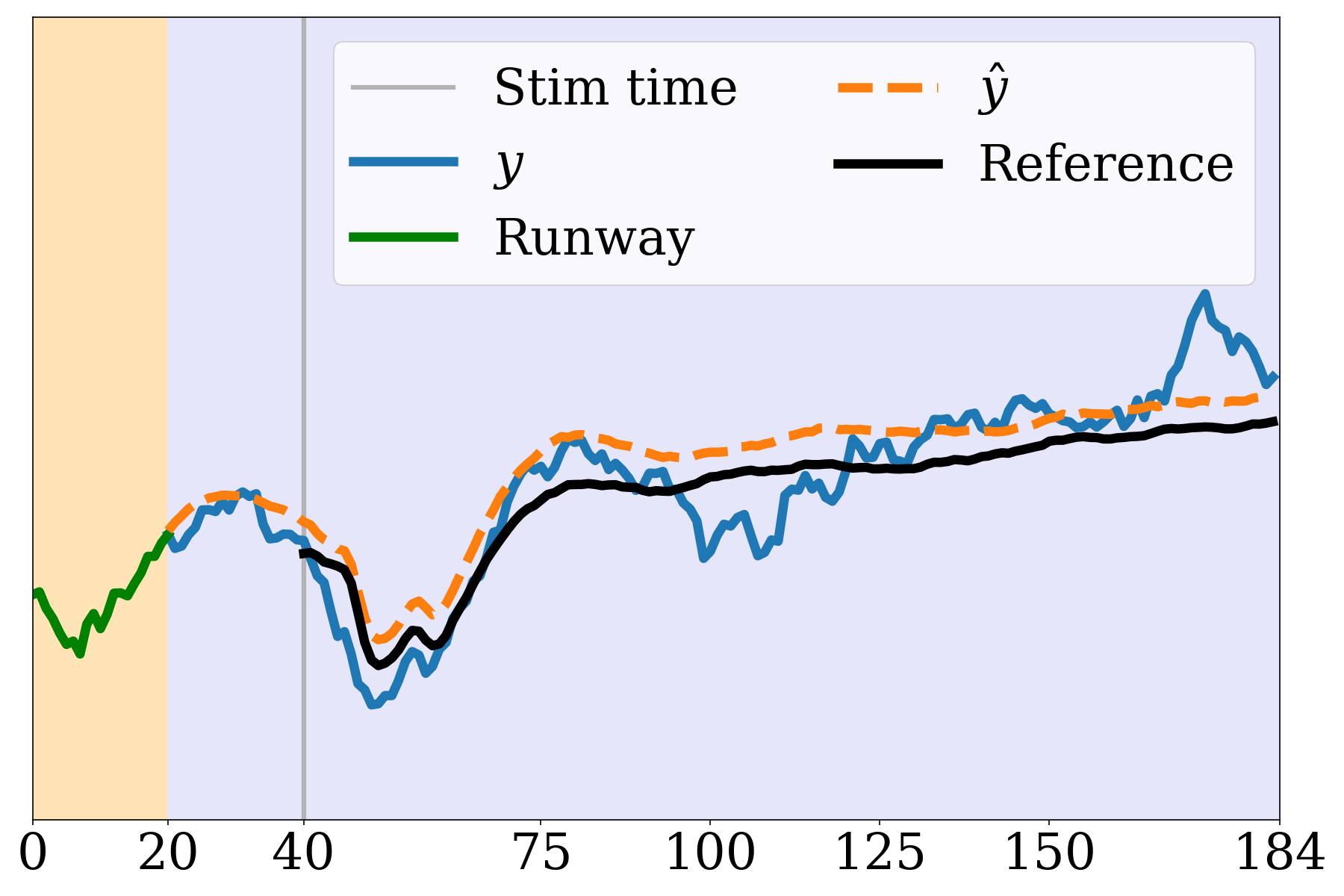}
		\caption{Demonstration 2}
	\end{subfigure}
	\hfill
\caption{\small \textbf{Demonstration of two closed-loop stimulators.} \textit{(a) Stimulation applied to target future brain states.} In this task, the controller must decide at t=$20ms$ to apply stimulation in the future at t=$40ms$, but only if a target brain state is forecast to occur. The controller leverages the stimulation temporal basis function model (TBFM) to forecast if neural activity across two channels will enter the target feature range (in this case a local field potentials (LFP) range, green color). If the model forecasts that the target will be hit, the controller will choose stimulation which is delivered at t=$40ms$ (example in dark green). If the model forecasts that neural activity will not cross the target range, the controller will not stimulate (example in red). Margins on the lower and upper of bounds of the target with size $\delta$ allow the experimenter to compensate for forecast uncertainty, biasing the controller in favor of over- or under-stimulation by expanding or contracting the range of the target.
\textit{(b) Stimulation to drive future brain state trajectories towards desired trajectories.} Here the controller stimulates only if the $L_2$ distance between its prediction (orange) and a reference trajectory (black) is below a threshold $\epsilon_p$. The threshold can be varied to bias in favor of more or less stimulation. The simulation evaluates the controller's performance by measuring the $L_2$ distance between the actual trajectory assuming stimulation (blue) and the reference, and comparing that distance to the same threshold $\epsilon_s$. This example depicts a true positive, i.e. where the predicted trajectory, actual stimulated trajectory, and reference are all similar.}
\label{fig:demos}
\end{figure}

While target features could be specified in any way, we use simple ranges for the purpose of our simulation. We target LFP ranges across two channels: the two where the stimulation pulses are applied. If the trajectory of brain activity is expected to pass through the target range on both channels at t=$40ms$ from trial onset, the controller should choose to stimulate. That is - it chooses at t=$20ms$ based on the ``runway'' data whether to apply stimulation in the future (at t=$40ms$) due to the assumed $20ms$ loop latency. Figure \ref{fig:demos}(a) depicts this visually for a single channel.

We allow the experimenter to bias the controller in favor of over- or under-stimulating by adjusting the target boundaries to be more or less permissive. We do that by ``inflating'' the target - adding some amount $\delta$ to each target boundary. In Figure \ref{fig:demos}(a) the margin is shown as the hashed green area $\delta$.

\subsubsection{Demonstration 2: stimulating to drive brain activity towards desired trajectories}
\label{sec:methods-demo2}
The goal of our second controller is to time stimulation in order to shape neural activities (LFP trajectories) to more closely resemble a desired target trajectory, called the ``reference trajectory''. It is assumed that the controller designer chooses reference trajectories that target particular healthy regimes or percepts. Figure \ref{fig:demos}(b) depicts this visually for a single channel. We explain how we pick reference trajectories for simulation purpose and provide other technical details in Appendix \ref{apx:demos}.

The controller triggers stimulation if its forecast is within some threshold $L_2$ distance of the reference trajectory $\epsilon$. As with Demonstration 1, varying $\epsilon$ allows for an under- versus over-stimulation tradeoff.

\subsection{Comparison models}
We compare our results to two existing model types: 1.) a simple linear state space model (LSSM) similar to that trained by a Kalman Filter; and 2.) a more complex nonlinear recurrent neural network model. We explicitly train both to perform multi-step state-dependent forecasts. The former appears in existing stimulation models, for both electrical and optogenetic stimulation \cite{bolus.opto, shanechi.stimmodel}; we refer to it as LSSM throughout. The latter - referred to as AE-LSTM - uses an autoencoder and long short-term memory (LSTM) layers \cite{murphy.ml}. It is strictly more expressive than LSSMs and is inspired by existing works which model complex nonlinear control problems \cite{kim.dyncorrespond, shi.deepkoopman, jung.lstmmpc}. We pick the AE-LSTM's latent state dimensionality and other hyperparameters to optimize its test set performance, rather than speed. Appendix \ref{apx:referencemodels} contains further technical details for both.

\section{Results}
\label{sec:results}

\subsection{Is the stimulation response state-dependent?}
\label{sec:results.state}
There exist two reasons why a model may provide better forward predictions by considering the initial brain state: 1.) in general future neural activity depends on past neural activity regardless of stimulation; and 2.) the stimulation response of the neural circuit depends on the neural activity at the time stimulation is applied \cite{bloch.statedep, bradley.statedep, kabir.statedep} - i.e. the stimulation response is be state-dependent. In this dataset we see clear dependence of stimulation responses on brain state across sessions and channels.

For a simple exploration of the state-dependence within a single channel, let us define the stimulation response as the difference between the observed trajectory and an estimate of what the trajectory would have been had we not stimulated. Specifically, since stimulation begins at t=$40ms$ we will refer to the channel's value immediately before ($x_{40}$) as the initial state. We estimate the stimulation response ($\boldsymbol{r}_t$) by subtracting the subsequent stimulated trajectory from a baseline (non-stimulation) trajectory passing through the same initial state. Then, we perform statistical analysis on the range $45ms$-$70ms$ into each trial in order to concentrate on the largest magnitude portion of the stimulation response. The test detects statistical dependence between $x_{40}$ and $\boldsymbol{r}_t$ - i.e. between the initial state and stimulation response. See Appendix \ref{apx:statedep} for more details.

\begin{figure}
	\centering
    \begin{subfigure}[c]{0.99\textwidth}
		\centering
		\includegraphics[width=\textwidth]{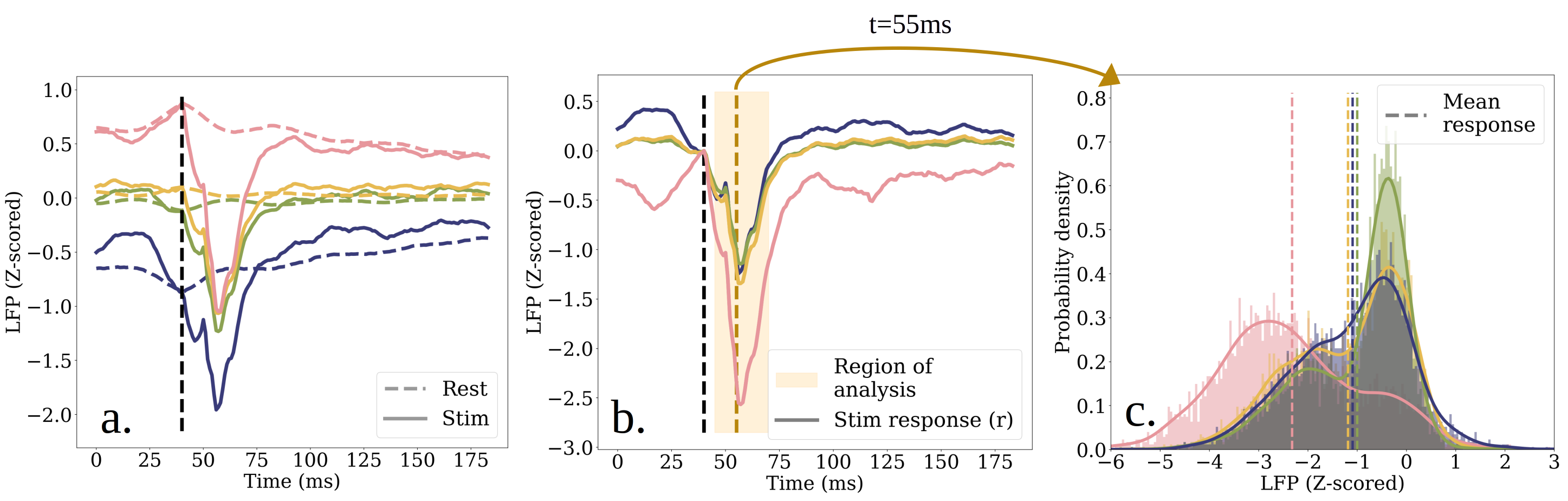}
	\end{subfigure}
	\hfill
\caption{\small \textbf{Example state dependence of stimulation response.}
(a) Trial-averaged resting and stimulation data for N=4 binned initial states, example session and channel. Vertical dashed line at t=$40ms$ denotes binning time. Resting data is trial-matched to stimulation trials as in Section \ref{sec:results.state}. Here we bin both into quartiles and trial-average for the purpose of graphing. (b) Trial-averaged stimulation response ($\boldsymbol{r}_t$) for each bin. The pink ``high'' initial state exhibits a significantly different stimulation response. We perform the statistical test for state dependence over the time range exhibiting the largest magnitude response: $45ms$-$70ms$ (orange region), and it considers the full distribution of initial states rather than binning them. See Section \ref{sec:results.state} and Appendix \ref{apx:statedep} for details of the test. (c) The histogram of stimulation responses at t=$55ms$ for each binned initial state. The long tails are truncated for clarify of presentation. Curves are kernel density estimates (KDE) of the histograms. The means (dashed lines) are $-1.09, -1.00, -1.19, -2.32$ for the example four states. The statistical test identifies differences between these distributions across the entire orange region of analysis. $97.4\%$ of channels show a statistically significant state dependence under that analysis.}
\label{fig:statedep}
\end{figure}

Overall we find that state-dependence is common across the dataset but does not appear on every channel. All sessions have at least one channel with a statistically significant state dependence at $95\%$ confidence. $97.4\%$ of channels exhibited statistically significant state dependence at that level. A visual depiction of one channel's state dependence can be seen in Figure \ref{fig:statedep}.

\subsection{Prediction accuracy}
\label{sec:results.accuracy}

\begin{table}[]
\small
\centering
\caption{Summary of Model Accuracies}
\begin{tabular}{|l|l|l|lll}
\hline
\textbf{Model} & \textbf{Data set}   & \textbf{\begin{tabular}[c]{@{}l@{}}$R^2$ 164ms\\ forward\\ prediction\end{tabular}} & \multicolumn{1}{l|}{\textbf{\begin{tabular}[c]{@{}l@{}}$R^2$ 40ms\\ forward\\ prediction\end{tabular}}} & \multicolumn{1}{l|}{\textbf{\begin{tabular}[c]{@{}l@{}}$R^2$ mean-\\ vs-mean\end{tabular}}} & \multicolumn{1}{l|}{\textbf{\begin{tabular}[c]{@{}l@{}}State-\\ dependent\\ $R^2$\end{tabular}}} \\ \hline
TBFM           & Time domain (train) & 0.533 (0.173)                                                                    & \multicolumn{1}{l|}{0.795 (0.139)}                                                                   & \multicolumn{1}{l|}{0.97 (0.04)}                                                         & \multicolumn{1}{l|}{0.93 (0.07)}                                                              \\ \hline
TBFM           & Time domain (test)  & 0.462 (0.207)                                                                    & \multicolumn{1}{l|}{0.787 (0.143)}                                                                   & \multicolumn{1}{l|}{0.88 (0.11)}                                                         & \multicolumn{1}{l|}{0.88 (0.12)}                                                              \\ \hline
TBFM           & Beta BP (train)     & 0.211 (0.051)                                                                    & \multicolumn{1}{l|}{0.682 (0.060)}                                                                   &                                                                                          &                                                                                               \\ \cline{1-4}
TBFM           & Beta BP (test)      & 0.114 (0.087)                                                                    & \multicolumn{1}{l|}{0.586 (0.153)}                                                                   &                                                                                          &                                                                                               \\ \cline{1-4}
AE-LSTM        & Time domain (test)  & 0.320 (0.166)                                                                    &                                                                                                      &                                                                                          &                                                                                               \\ \cline{1-3}
LSSM           & Time domain (test)  & 0.179 (0.160)                                                                    &                                                                                                      &                                                                                          &                                                                                               \\ \cline{1-3}
\end{tabular}
\label{table:accuracy}
\end{table}

\begin{figure}
	\centering
    \begin{subfigure}[c]{0.99\textwidth}
		\centering
		\includegraphics[width=\textwidth]{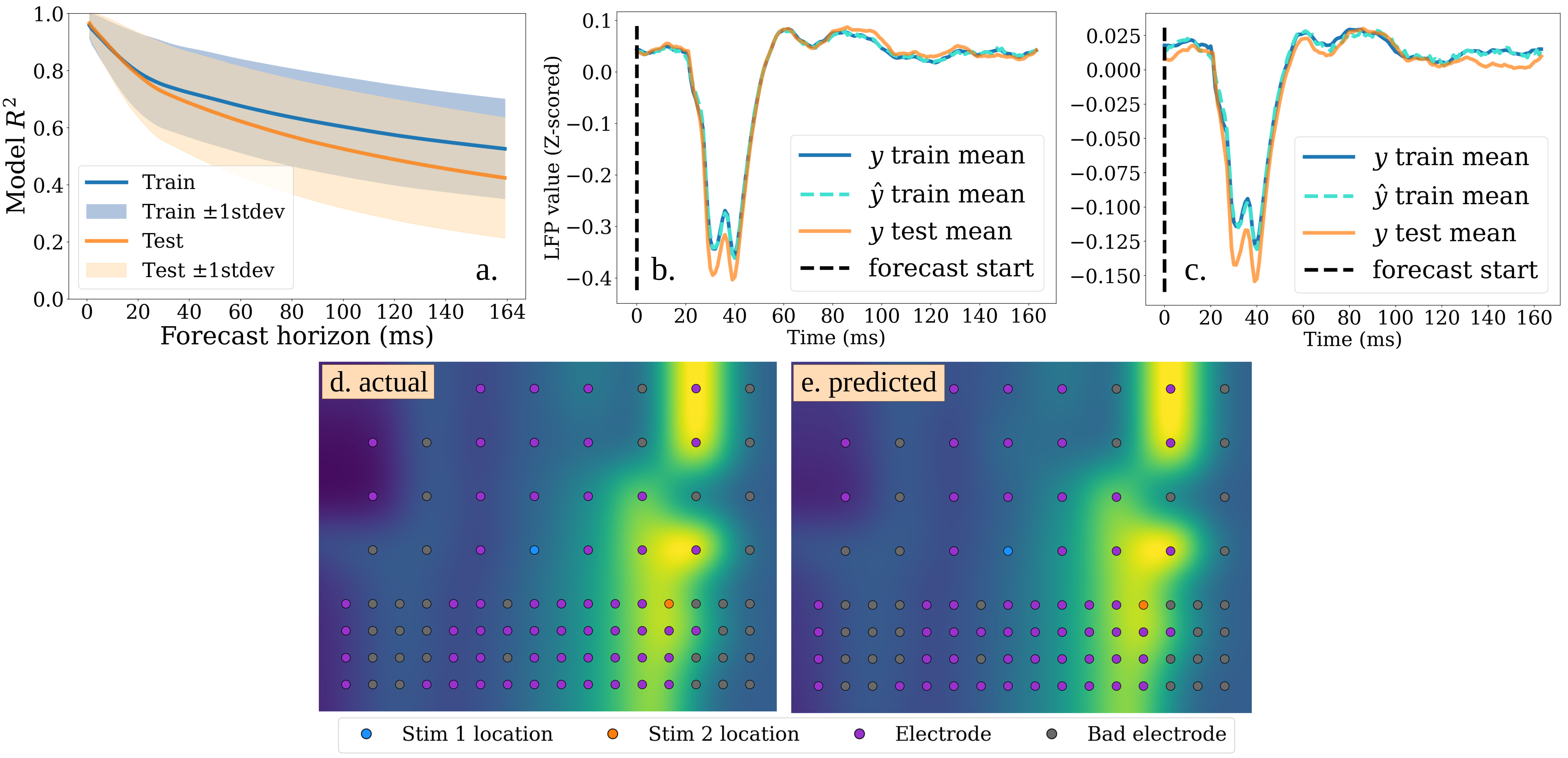}
	\end{subfigure}
	\hfill
\caption{\small \textbf{Performance of the temporal basis function model (TBFM) for time domain data.} \textbf{(a)} Training versus test $R^2$ for time domain data. The plots show $R^2$ computed for
3 runs per session over 40 sessions as a function of prediction horizon. Mean $R^2$ 
for a horizon of $164ms$ is 0.533 (stdev $0.173$) on training data and 0.462 (stdev $0.207$) on test data. \textbf{(b)} Example of mean forecast trajectory ($\boldsymbol{y}_t$) for a single channel near to the stimulation site, and a single session. Our model's mean forecast ($\hat{\boldsymbol{y}}_t$) fits the mean response of the training data well (statistics in main text). \textbf{(c)} a second channel further from the stimulation sites.
\textbf{(d)} Mean spatial LFP response across the electrode array at t=$50ms$ for the same example session as (b,c). \textbf{(e)} Spatial prediction of the same by the model. (d,e) TBFM predicts accurately across space in addition to time (b).}
\label{fig:r2}
\end{figure}

We found that the TBFM, LSSM, and AE-LSTM models all successfully learned to predict the stimulation neural dynamics. The TBF models performed slightly better than the AE-LSTM models overall, and significantly better than LSSMs. The results vary considerably between sessions and filtering methods. Figure \ref{fig:r2}a shows results for time domain data filtered only for line noise. Table \ref{table:accuracy} provides a statistical summary. For those data the overall $R^2$ of the TBFM for a $164ms$-long prediction of LFP response was on average 0.533 (stdev $0.173$) on training data and 0.462 (stdev $0.207$) on test data across sessions. As expected, the $R^2$ is higher for predictions over shorter time horizons.

\begin{figure}
	\centering
	\begin{subfigure}[c]{0.49\textwidth}
		\centering
		\includegraphics[width=\textwidth]{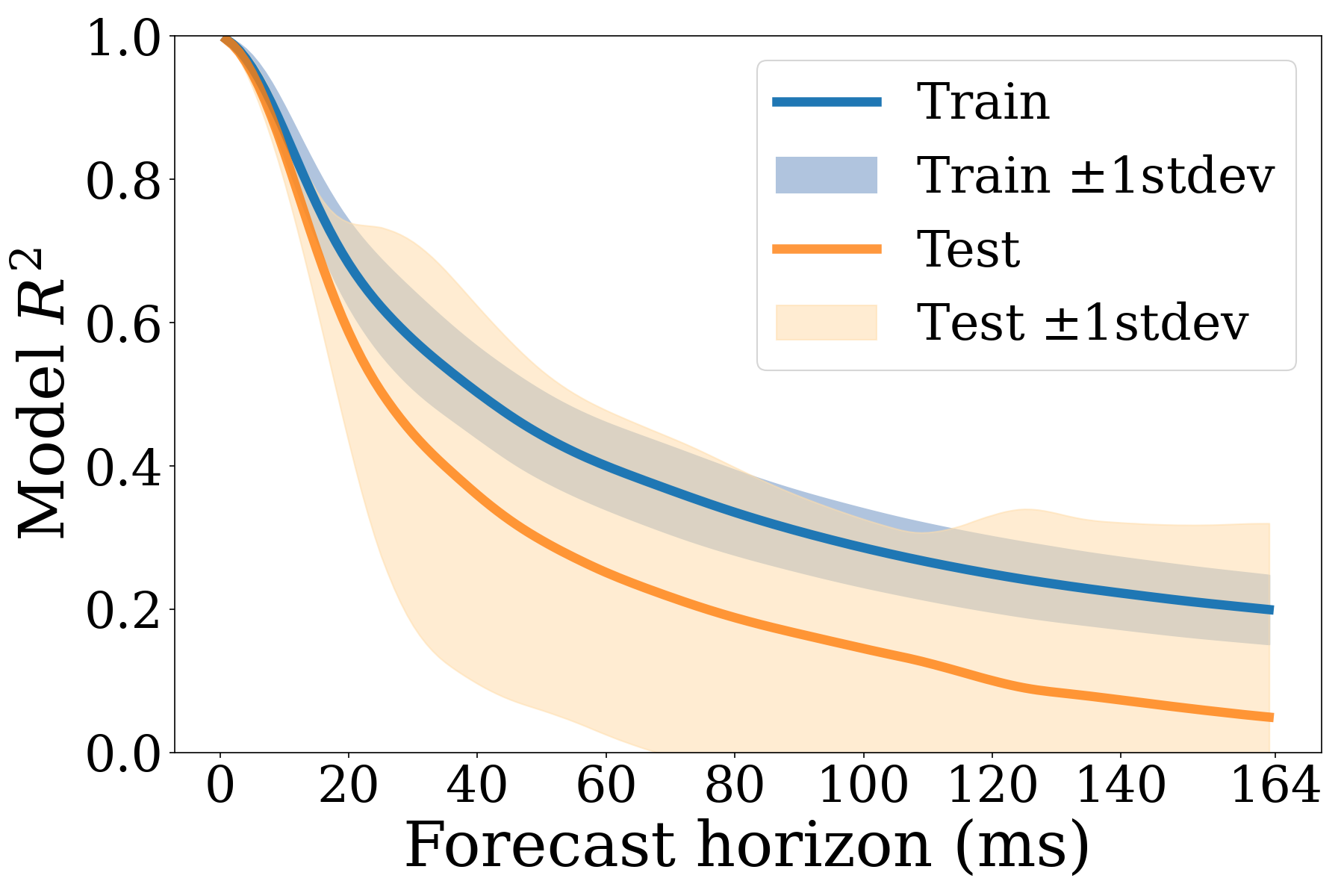}
		\caption{}
	\end{subfigure}
	\hfill
 	\begin{subfigure}[c]{0.49\textwidth}
		\centering
		\includegraphics[width=\textwidth]{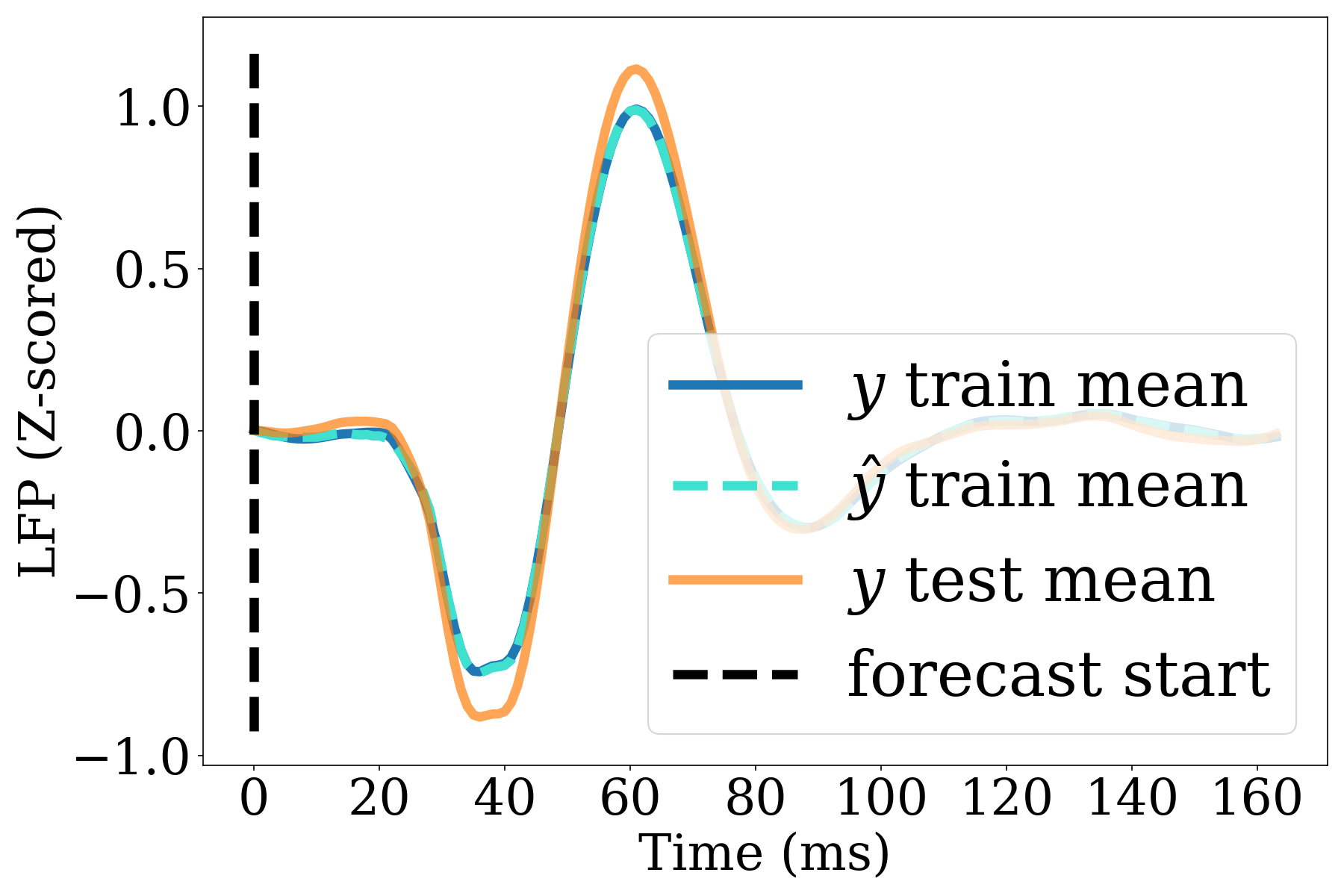}
		\caption{}
	\end{subfigure}
	\hfill
\caption{\small \textbf{Performance of the temporal basis function model (TBFM) for beta-bandpassed data.} (a) Training versus test $R^2$ for beta bandpassed data across all sessions. Training set $R^2$: $0.211$ (stdev $0.051$), test set: $0.114$, (stdev $0.087$).
(b) Example of mean forecast trajectory for a single channel and a single session.}
\label{fig:r2bp}
\end{figure}

As shown in Figure \ref{fig:r2bp}, the $R^2$ values for beta-bandpassed time-domain data are lower than for the less filtered data.  Once again the $40ms$ forecast exhibits much higher test set $R^2$. Despite the lower $R^2$ with this filtering method, we found it was still sufficient to achieve a reasonable level of control in simulation; see Section \ref{sec:results.simulation}.

Importantly, the model's mean predictions track the mean stimulation
response across all trials (see Figure \ref{fig:r2}(b) for a graphical example). This mean response can be viewed as the stimulation-driven dynamics since it averages over all observed initial states and exogenous inputs. Averaged over all sessions, the $R^2$ for the $164ms$ forecast horizon between the model's mean prediction and the mean stimulation response was 0.97 (stdev $0.04$) on the training data and 0.88 (stdev $0.11$) on the test data, suggesting that the model may be an unbiased predictor of the state-agnostic stimulation response. We explore the state-dependence of the model's predictions in Section \ref{sec:results.modelstatedep}.

Figures~\ref{fig:r2}(c,d) show the mean spatial response and the model's prediction for the same session as Figure~\ref{fig:r2}(b). The predicted response in (d) closely tracks the actual response in (c). Together with the $R^2$  measures discussed above, we conclude that the learned TBF model successfully predicts the spatiotemporal stimulation response.

Somewhat surprisingly, our simple TBF model's $R^2$ on the test data exceeded that of the more complex AE-LSTM model on all but 6 of the 40 sessions. Based on observation of training logs, we hypothesize this is due to the inherent difficulty in training deep recurrent models to perform multi-step forward predictions \cite{murphy.ml}, and the limited amount of data relative to the dimensionality of the models. See Appendix \ref{apx:refdyns} for additional analysis. Less surprisingly, the LSSM model performance was lower than AE-LSTM overall. This is less surprising since this model is strictly less expressive than AE-LSTM. TBFM exhibited a higher test set $R^2$ than LSSM on all sessions.

\subsection{Does the TBFM adapt to the state-dependence of the stimulation response?}
\label{sec:results.modelstatedep}
The TBF model seeks to forecast the initial-state-dependent stimulation response. That key property differentiates our approach from approaches based only on the stimulation parameters. We therefore examine 1) whether our model learns to condition its predictions on the initial brain
state; 2) whether that prediction is accurate across initial states.

As seen in Section~\ref{sec:results.state}, the stimulation
response is influenced by initial state, but the strength of that
influence varies between sessions and channels. We analyzed our model's performance by binning data according to the discrete initial states as in Figure \ref{fig:statedep}. Specifically, we bin based on the initial state at t=$20ms$, which is the point where the model's 
prediction begins. For each channel, we create 9 bins such that each bin has an equal number of trials.\footnote{We have not developed a principled way of selecting the bin count, but note
that it may be possible to do so using measures of statistical power.} We calculated the mean response for each bin, and calculated the $R^2$ between it and the mean prediction for the same set of trials. The average across sessions we call the ``state-dependent $R^2$''.

\begin{figure}
	\centering
    \begin{subfigure}[c]{0.99\textwidth}
		\centering
		\includegraphics[width=\textwidth]{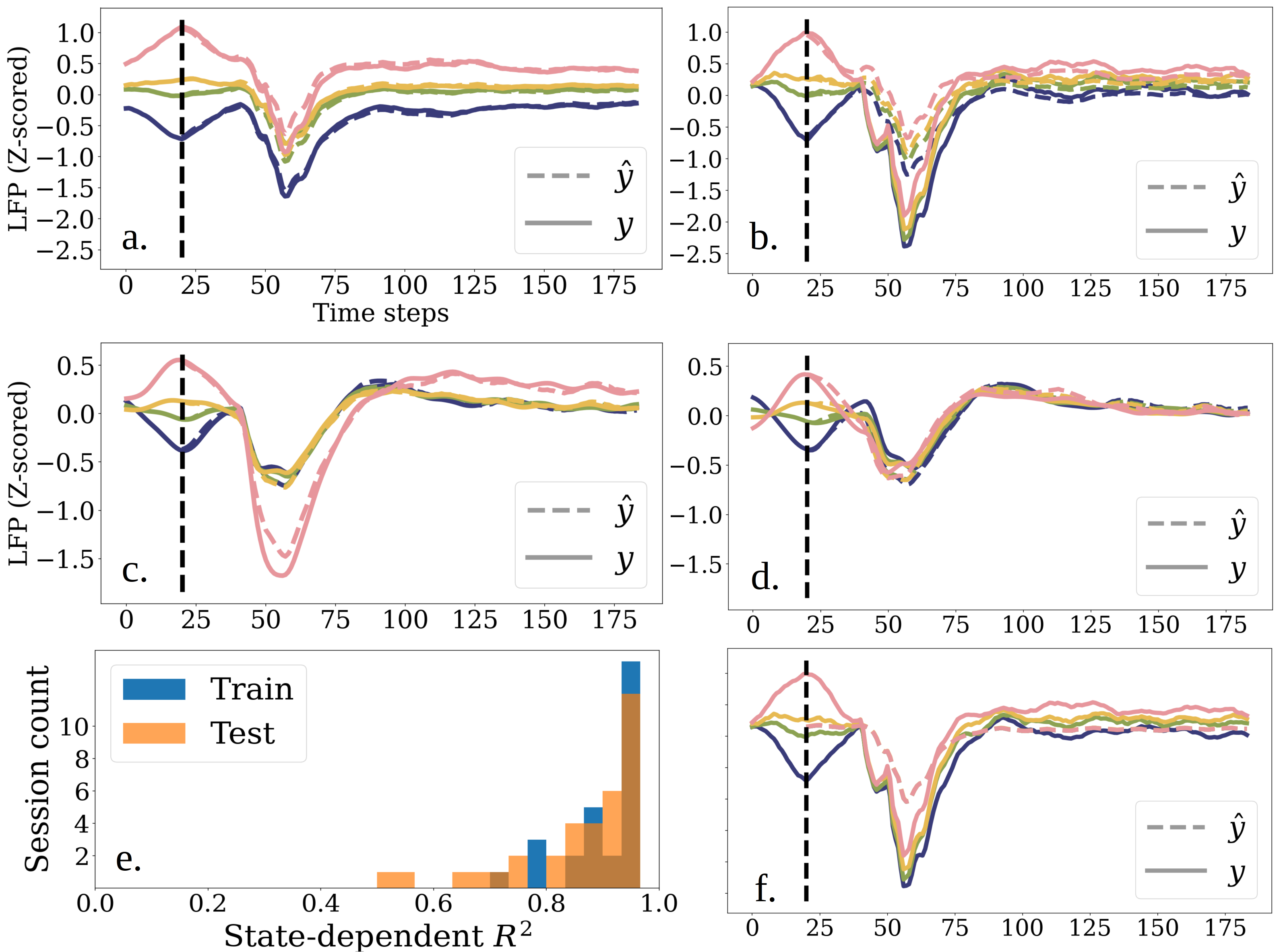}
	\end{subfigure}
    \hfill
\caption{\small \textbf{Mean stimulation response as a function of initial state for time-domain data.}
Mean stimulation response versus prediction across binned initial states (t=$20ms$) \textbf{(a)} training set, example session and channel. State-dependent $R^2$: $0.94$ \textbf{(b)} test set; state-dependent $R^2$: $0.86$. The temporal basis function (TBF) model successfully conditions its prediction on initial state, rather than outputting an unconditioned mean response. \textbf{(c,d)} State dependence in beta bandpassed example. The pink ``high'' state exhibits significantly different response in the training set ($p<0.05$). In the test set its state dependence is somewhat lower. Due to regularization the model performs well on the test set despite the difference.
\textbf{(e)} Histogram of initial-state-dependent $R^2$ measures for training and test datasets. Mean train: 0.930 (stdev $0.0686$), test: 0.878 (stdev $0.1160$).
\textbf{(f)} Result from a state-agnostic TBF model which does not
condition its predictions on initial state. Same channel as (a,b). It learns only a mean stimulation response (overlapping dashed plots). The mean test set $R^2$ for the state-agnostic model across all sessions was
$0.006$ (stdev $0.012$), compared to a mean of $0.462$ (stdev $0.207$) for the state-dependent models.}
\label{fig:results.statedep}
\end{figure}

With a prediction horizon of $164ms$ the state-dependent $R^2$ averaged across sessions is 0.930 (stdev $0.0686$) on training datasets, and 0.878 (stdev $0.1160$) on test datasets. This indicates that the model learns an accurate state-dependent stimulation response rather than a mean (state-agnostic) stimulation response. Figure \ref{fig:results.statedep}(a,b) provides a visual depiction of mean responses and TBFM-based predictions for N=4 states on a session's training and test set, respectively. (c,d) depict the training and test sets for beta bandpassed data. Figure \ref{fig:results.statedep}(e) shows a histogram of the state-dependent $R^2$ across sessions. Appendix \ref{apx:filtering} contains additional results for bandpassed and time-frequency domain data.

To measure whether initial state conditioning improves forecasting,  we compared the overall $R^2$ to a state-agnostic TBF model trained on the same time domain LFP data. The state-agnostic TBF models received sham runway inputs which were identical for all examples, thus eliminating initial state-dependence. The initial-state-dependent models exhibited higher $R^2$ on test sets: the mean test set $R^2$ across all sessions was $0.462$ (stdev $0.207$), compared to a mean of $0.006$ (stdev $0.012$) for state-independent models. Figure \ref{fig:results.statedep}(f) depicts the binned mean predictions (dashed lines) for a state-agnostic model, and we see that they are identical for all states (dashed lines overlap; compare with (a)).

\subsection{Modeling multiple stimulation parameters}
\label{results:msts}
We test the TBFM's ability to work with multiple stimulation parameterizations by building one which generates bases for multiple sessions, using different stimulation descriptors as input. Our dataset does not vary stimulation parameters within a given session, but we can demonstrate that a rich enough basis generator can generalize across sessions, and therefore across a limited set of stimulation parameters. Specifically we learn a TBFM which can generalize across three sessions where the interval between pulses was different in each: $10ms$, $30ms$, and $100ms$. Because the three sessions contain different numbers of usable electrodes and spatial activity patterns, the basis weight estimators are not shared between them.

For this case we performed a simple hyperparameter search and found best performance with 15 bases, four hidden layers, each with a width of four, and with a regularizer weight $\lambda = 0.03$.

The test set $R^2$ across the sessions increased for two of the three sessions relative to the single session model. This surprising fact is due to our one-size-fits-all approach to hyperparameter tuning: the hyperparameters used for the single session models were all the same and worked better for some sessions than others. Overall we can conclude that our MLP basis generator is sufficiently rich to generate good results in this case. Appendix \ref{apx:multisession} contains additional detail.

\subsection{Sample efficiency}
\label{sec:results.sampleefficiency}
We analyzed our models' sample efficiency by varying the training set size and measuring $R^2$ on the test set. To measure the ability of a model to generalize between the early part and late part of a session, we source our training sets from the first N trials of each session, and the test set from the end (see Appendix \ref{apx:dataset} for additional details). Since test set $R^2$ improves nearly monotonically with training set size, we consider the optimal training set size to be the smallest which has a test set $R^2$ within $1\%$ of the maximum.

We found that the TBFM requires $N$$\approx$6k training examples on average across sessions to maximize test set performance, with a mode of $5k$. A $5k$ dataset requires $<$$17min$ to collect in an experimental setting. Given this relatively small amount of training data and the short training time required (see Section \ref{sec:results.training.time}), the model can easily be trained
and deployed within the span of a typical experimental session. The 5k training set size was used to obtain the rest of the results in this paper.

\begin{figure}
\centering
\begin{subfigure}[c]{0.50\textwidth}
    \includegraphics[width=\textwidth]{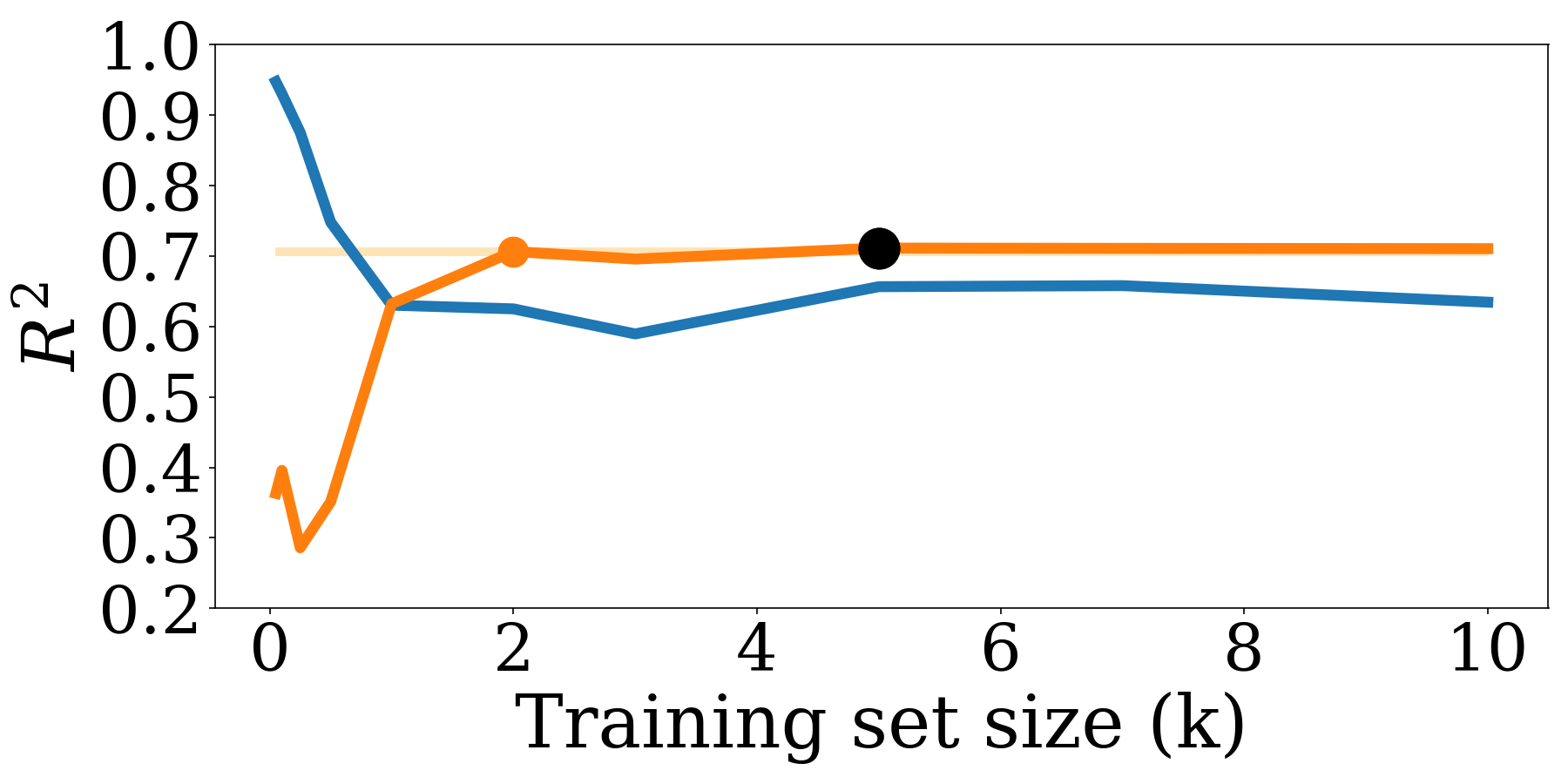}
    \caption{Temporal basis function model (TBFM)}
\end{subfigure}
\hfill
\begin{subfigure}[c]{0.49\textwidth}
    \includegraphics[width=\textwidth]{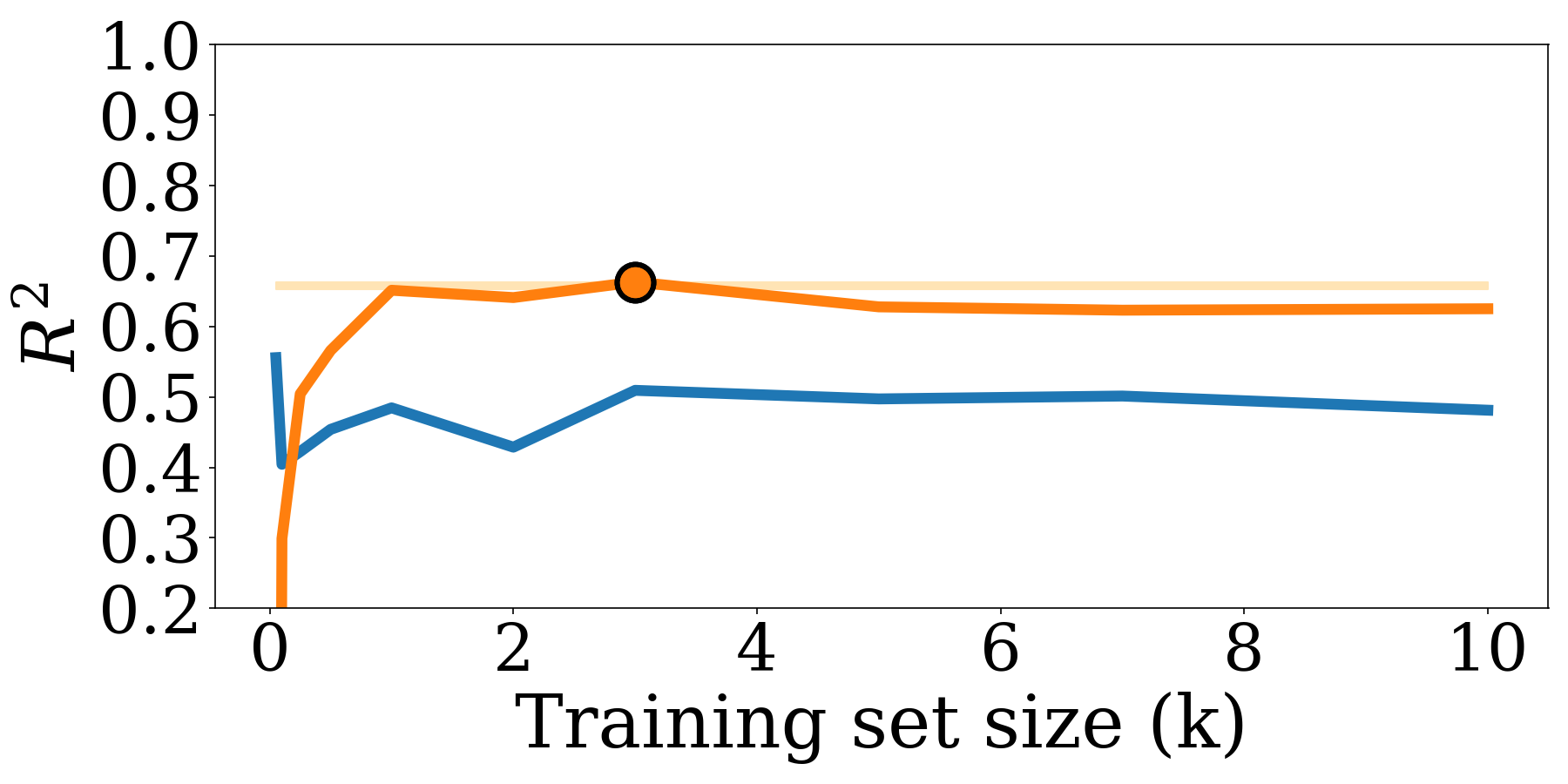}
    \caption{Linear state space model (LSSM)}
\end{subfigure}
\hfill
\begin{subfigure}[c]{0.49\textwidth}
    \includegraphics[width=\textwidth]{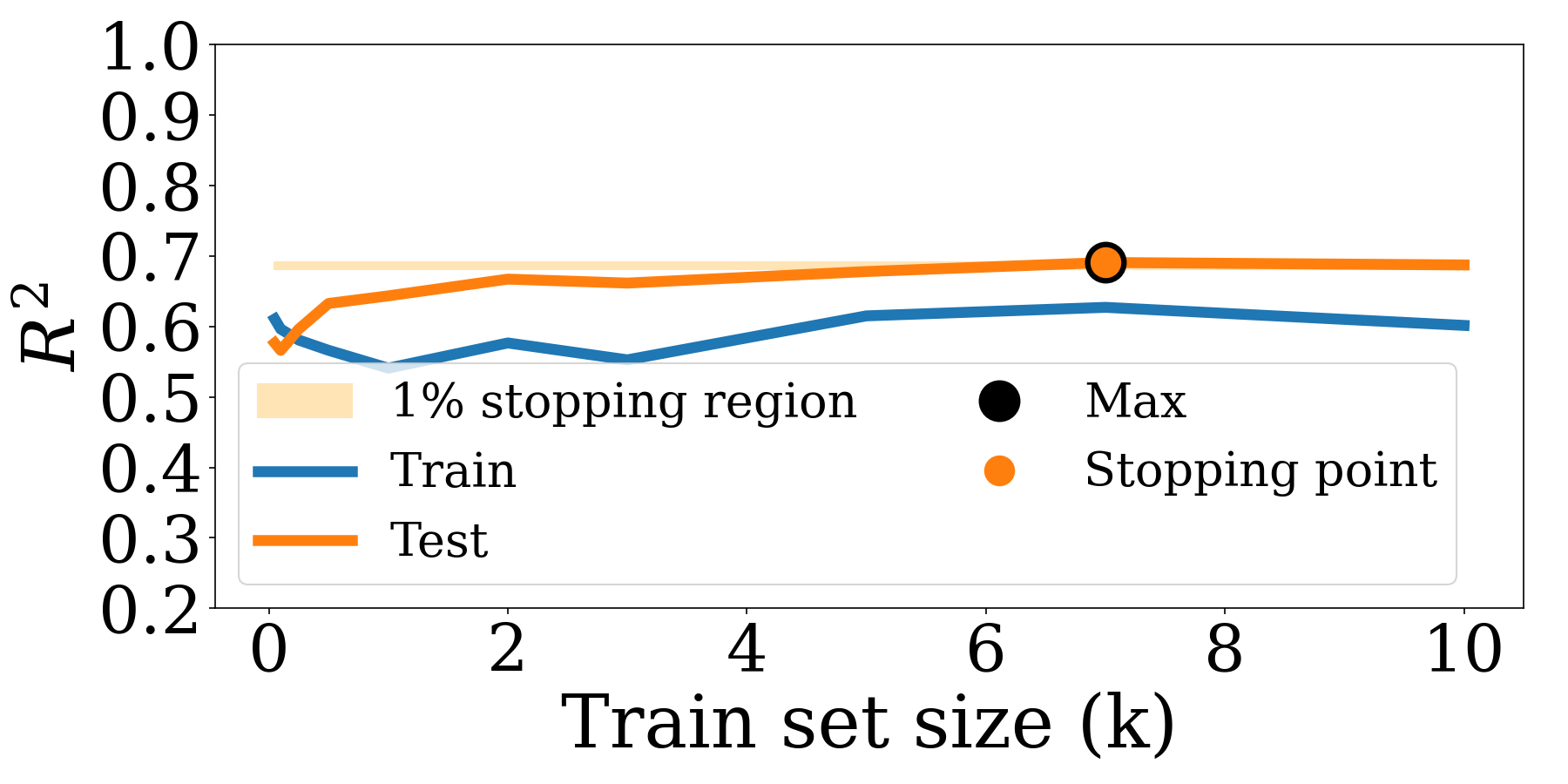}
    \caption{Autoencoder with LSTM dynamics (AE-LSTM)}
\end{subfigure}
\hfill
\caption{\small \textbf{Sample efficiencies for an example session.} We vary the training set size and measure the $R^2$ of the temporal basis function (TBF) model on both the training set and test set. The training set is sourced from the first N trials of the session. The test set is sourced from the end and is fixed size (2500 trials).
We consider a good training set size (orange marker) to be one within $1\%$ of the maximum test set $R^2$ on that session. The resulting stopping points on this session were 2k, 3k, and 7k for TBFM, LSSM, AE-LSTM respectively.}
\label{fig:sample.efficiency}
\end{figure}

The AE-LSTM model exhibited a similar trend on a single session, with a similar inflexion point at $N$$\approx$7k training examples. See Figure \ref{fig:sample.efficiency}. On the same session the TBFM achieved an optimal stopping point at $N$$\approx$2k. However, it is computationally intractable to repeat this analysis on all sessions with the AE-LSTM since that would require an estimated 9 months of compute time, due to slow training. As a result, we do not perform a statistical test of the models' differences in sample efficiency. Likewise the LSSM had an inflexion point at $3k$. Once again we trained it on only the one session due to the slow training speed, so we cannot statistically test the difference. 

\subsection{Training time}
\label{sec:results.training.time}

Overall a TBF model on our time series dataset  requires $\approx$2-5 minutes ($\approx$120-300s)  of training time, depending on the number of usable electrodes
in the given session. One session had as few as 40 usable electrodes, requiring $133.1$ (stdev $0.02$) seconds of training time. A different session had 94 usable electrodes, requiring $280.2$ (stdev $0.07$) seconds. Such training times are short enough to allow the use of the TBF model in experiments without requiring significant time allocation for the training process.

The LSSM model by comparison required an average of $90.6min$ (stdev $13.4min$). The AE-LSTM model required on average $11.53hr$ (stdev $1.11hr$) to train. This long training time reflects the slow execution time of recurrent networks compared to TBFMs and the lower learning rates needed to ensure stability of training deep recurrent networks. The order of magnitude difference in training time between the models highlights the significant advantage of using a simpler and non-recurrent model for predicting stimulation response.

\subsection{Inference latency}
\label{sec:results.inference}

We benchmark the latency of our trained models on test data by providing different runways of time series data (N=$10000$) to our models as input and recording the mean execution time needed for forecasting. The execution time includes any movement of data to/from our graphics processing unit (GPU, where applicable), as well as Z-scoring and the model execution time. We assume the data already resides in the main memory on the computer, as it would if our process had just received it from a device driver. We further assume that any latency introduced by filtering, drivers, the stimulator, etc.\ would be equal across all models, and thus not useful to include in the latency measure when comparing models. The devices we benchmark on are a CPU (AMD Ryzen Threadripper PRO 5955WX 16-Cores) and a GPU (NVIDIA GeForce RTX 4090).

In our first simulated demonstration (Section \ref{sec:results.simulation}) we use a forecast horizon of $40ms$, and in the second a horizon $164ms$. Recurrent models like LSSM and AE-LSTM have latencies more directly proportional to the length of the forecast horizon. As a result, we compare results for these two forecast horizons.

Table \ref{table:latency} shows the results. The lowest latency model is a compiled TBF model\footnote{Appendix \ref{apxsubsec:compilation}} running on the CPU, requiring $0.115ms$ (stdev $0.006$) on average for a $164ms$ forecast. This represents a $34\%$ latency reduction relative to the uncompiled model, and a $5\%$ reduction relative to a compiled model running on the GPU. Results are similar for the $40ms$ forecast.

Using a GPU requires the movement of data to the GPU and moving the result back from the GPU, both of which contribute to latency. With the AE-LSTM model, this movement is worth the payoff in terms of reduced execution time, but with the TBF models, it isn't: execution is faster on CPU overall. With GPU use, however, the AE-LSTM model exhibits latency two orders of magnitude higher that with the CPU, reflecting the LSTM model's complexity and the recurrent nature of its computational graph. Our LSSM implementation runs faster on CPU, but once again exhibits 1-2 orders of magnitude slower execution. Shortening the horizon forecast has a stronger effect on the recurrent models as expected, but they remain 1 order of magnitude slower than the TBFM model in this case.

\begin{table}
\small
\centering
\caption{Model execution time (in ms), Average of N=$10000$ Runs}
\begin{tabular}{l|rr|rr|}
\cline{2-5}
                                      & \multicolumn{2}{c|}{\textbf{164ms forecast}}   & \multicolumn{2}{c|}{\textbf{40ms forecast}}                        \\ \hline
\multicolumn{1}{|l|}{\textbf{Model}}  & \multicolumn{1}{l|}{CPU}            & \multicolumn{1}{l|}{GPU} & \multicolumn{1}{l|}{CPU}            & \multicolumn{1}{l|}{GPU} \\ \hline
\multicolumn{1}{|l|}{TBFM Uncompiled} & \multicolumn{1}{r|}{0.174 ($\pm$0.013)}  & 0.226 ($\pm$0.012)    & \multicolumn{1}{r|}{0.165 ($\pm$0.036)} &  0.226 ($\pm$0.004)      \\ \hline
\multicolumn{1}{|l|}{TBFM Compiled\footnote{Appendix \ref{apxsubsec:compilation}}.}   & \multicolumn{1}{r|}{\cellcolor[HTML]{FFFFC7}0.115 ($\pm$0.006)}  & 0.121 ($\pm$0.003)     & \multicolumn{1}{r|}{\cellcolor[HTML]{FFFFC7}0.108 ($\pm$0.007)} & 0.121 ($\pm$0.004)     \\ \hline
\multicolumn{1}{|l|}{LSSM}        & \multicolumn{1}{r|}{10.111 ($\pm$1.928)} & 16.939 ($\pm$0.110)    & \multicolumn{1}{r|}{1.524 ($\pm$0.086)} &  4.693 ($\pm$0.619)     \\ \hline
\multicolumn{1}{|l|}{AE-LSTM}        & \multicolumn{1}{r|}{41.628 ($\pm$6.480)} & 34.156 ($\pm$1.262)     & \multicolumn{1}{r|}{9.548 ($\pm$0.414)} &  8.494 ($\pm$0.069)    \\ \hline
\end{tabular}
\label{table:latency}
\end{table}

\subsection{Forward stagewise additive modeling}
\label{sec:results.additive}
\begin{figure}
	\centering
	\begin{subfigure}[c]{0.49\textwidth}
		\centering
		\includegraphics[width=\textwidth]{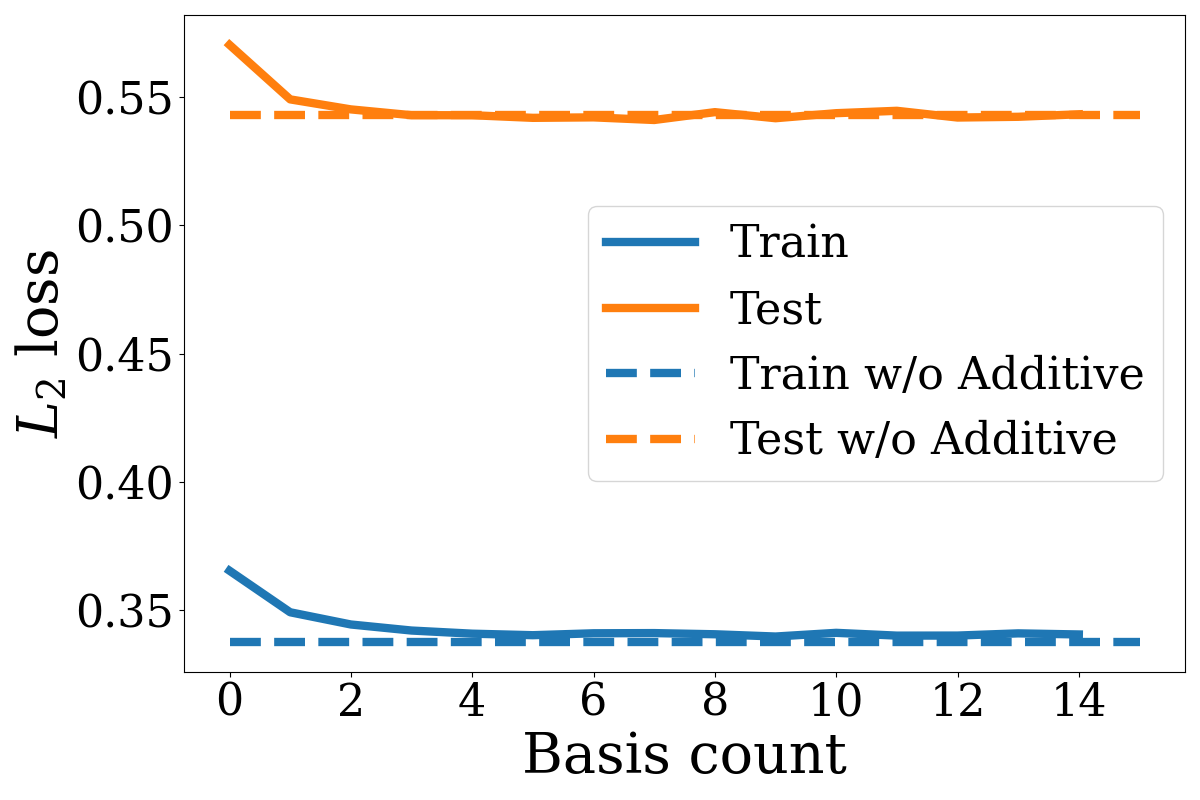}
		\caption{$L_2$ loss decreases with basis count}
	\end{subfigure}
	\hfill
	\begin{subfigure}[c]{0.50\textwidth}
		\centering
		\includegraphics[width=\textwidth]{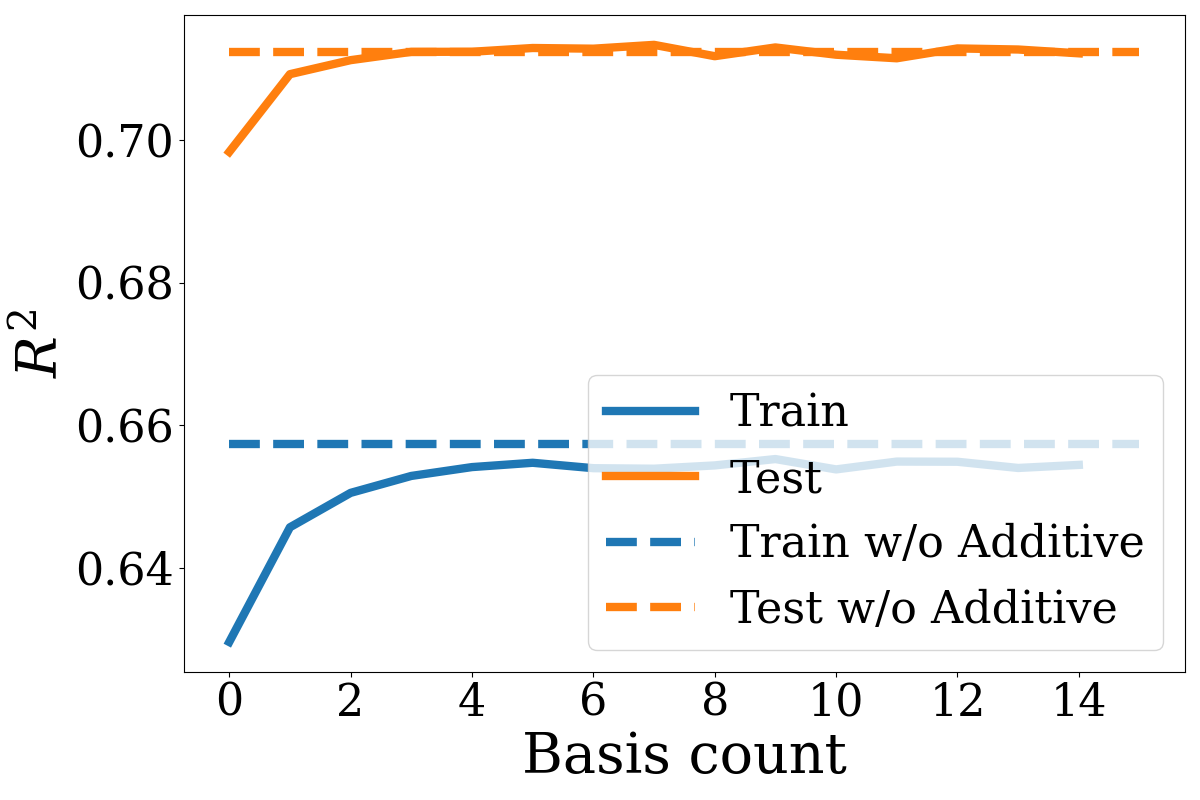}
		\caption{$R^2$ increases with basis count}
	\end{subfigure}
	\hfill
\caption{\small \textbf{Forward Stagewise Additive Modeling (FSAM) results on an example session.} As with principal component analysis, FSAM tends to reduce $L_2$ loss \textbf{(a)} and increase $R^2$ \textbf{(b)} as additional basis functions are added, though with diminishing returns. The dashed plots show results with our original non-additive method (train $R^2: 0.65$, test $R^2: 0.71$) using 12 basis functions on the same session.}
\label{fig:additiveresult}
\end{figure}

We also tested forward stagewise additive modeling (FSAM; see Section~\ref{sec:additive}) as a method for incrementally adding new basis functions to the model. We found that FSAM yields similar prediction accuracy to the method we used above on the two sessions on which we tested it (we did not attempt it on all sessions due to limits on computational time). In general, the loss decreases and $R^2$ increases on the training set as we add more basis functions (Figure \ref{fig:additiveresult}). The results for the test set are similar: test set $R^2$ does not improve after we add $\approx$$8$ basis functions (Figure~\ref{fig:additiveresult}). This suggests a stopping rule for choosing an optimal set of basis functions, namely, stopping the additive process when a convergence point is reached on a validation set. Note that because this type of training requires time proportional to the number of bases ($\approx$$52min$ in our example), it may be most appropriate in some cases to apply this to a single session and to use that result to pick a basis count for all other sessions.

\begin{figure}
	\centering
  	\begin{subfigure}[c]{\textwidth}
		\centering
		\includegraphics[width=\textwidth]{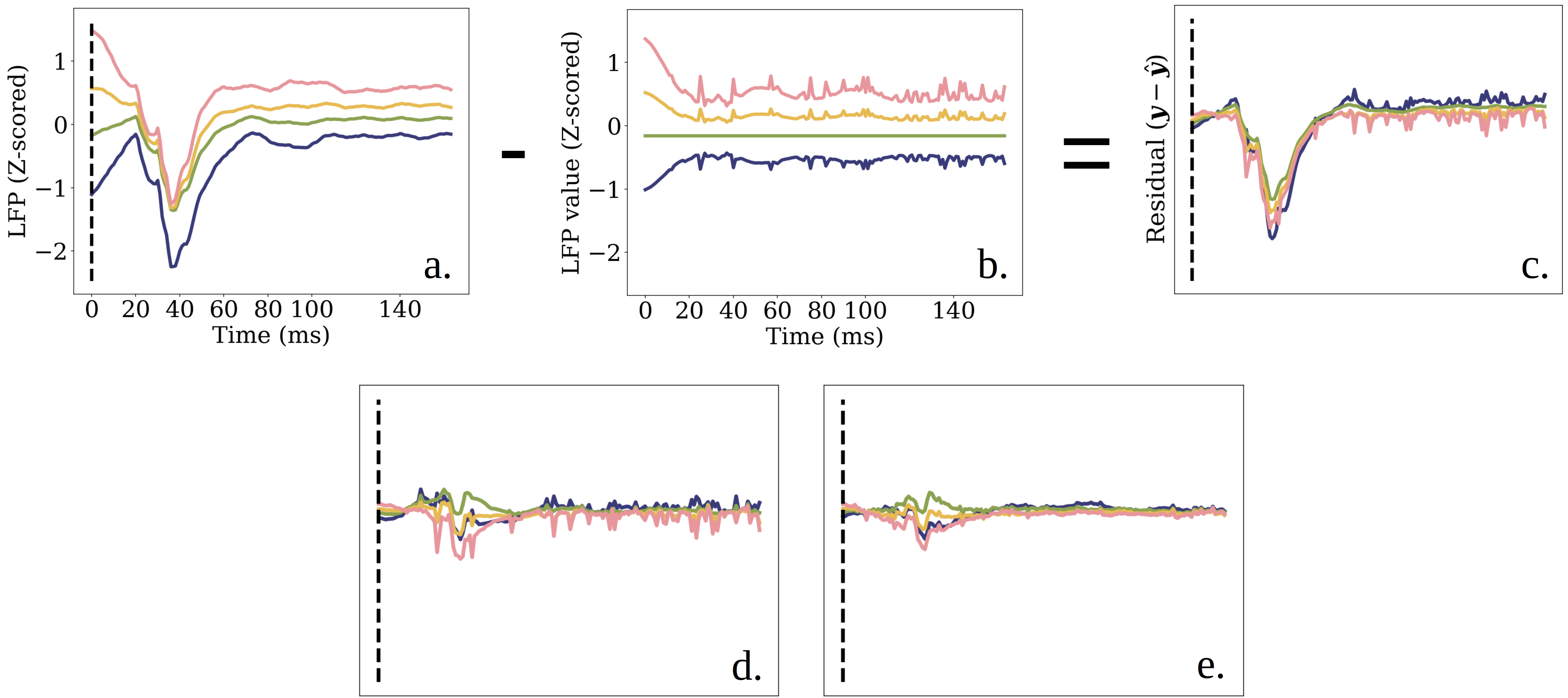}
	\end{subfigure}
	\hfill
\caption{\small \textbf{Residuals in Forward Stagewise Additive Modeling (FSAM).} \textbf{(a)} Mean responses computed from three sets of initial states: low (blue), average (orange), and high (green). \textbf{(b)} Mean prediction $\hat{\boldsymbol{y}}$ for each of the states made with a single basis function. All predictions are a weight multiplied by that single basis function, plus an intercept. \textbf{(c)} The mean residuals ($\boldsymbol{y}_c - \hat{\boldsymbol{y}_c}$) for these three initial states are similar after adding the first basis function, implying that this basis function can be interpreted as capturing much of the overall effect of initial state on the stimulation response. \textbf{(d)} Adding the second basis function removes most of the effect of stimulation that is common across initial states, namely, the two downward deflections in response due to the two pulses. \textbf{(e)} At 12 basis functions, the remaining variance is smaller, indicating that this number is sufficient for capturing most of the variance of the stimulation response across all those states, as measured by state-dependent $R^2$.}
\label{fig:additiveresiduals}
\end{figure}

The FSAM approach additionally provides some amount of interpretability: as seen  in Figure \ref{fig:additiveresiduals}, the first two basis functions appear to play distinct roles. The first basis function can be interpreted as the initial state's contribution to the stimulation response. We see the model residuals are largely similar across states after adding a single basis function, indicating that function is accounting for the effect of initial state. The second basis function can be interpreted as the channel- and initial-state agnostic stimulation response $\boldsymbol{a}_t$ as described in Appendix \ref{apx:statedep}: it removes across all states the majority of the double-trough response due to the two pulses. The remaining basis functions play roles less easy to identify (see Appendix \ref{apx:additivebasis} for examples and compare to the basis functions of a non-additive model in Appendix \ref{apx:basis}).

\subsection{Simulation}
\label{sec:results.simulation}

\begin{figure}
  	\begin{subfigure}[c]{\textwidth}
		\centering
		\includegraphics[width=\textwidth]{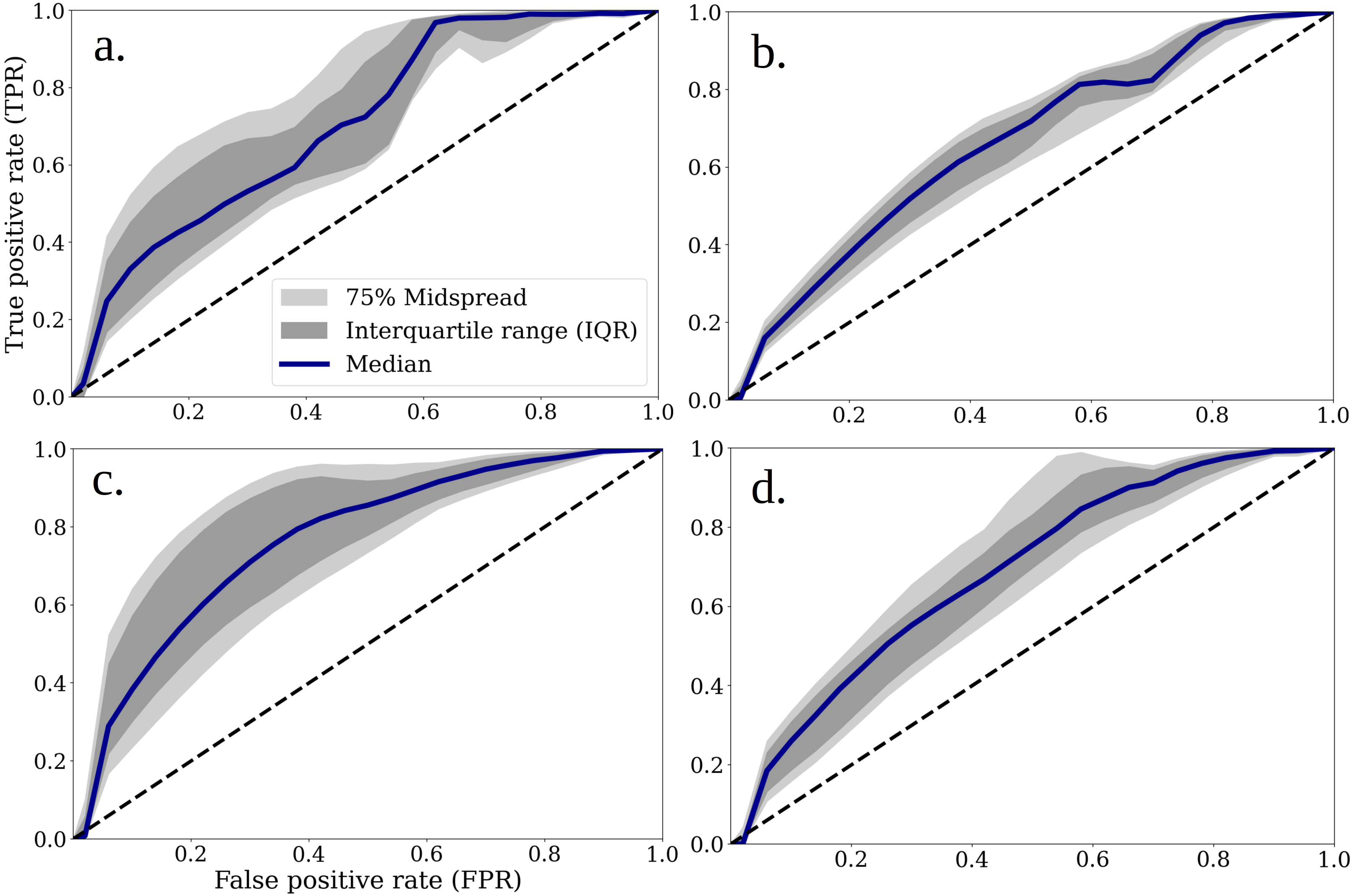}
	\end{subfigure}
	\hfill
\caption{\small \textbf{Results of simulated closed-loop brain stimulation}
\textbf{(a)} Demo 1, unfiltered time domain, mean AUC 0.704 (stdev 0.090); \textbf{(b)} Demo 1, $\beta$ bandpassed, mean AUC 0.652 (stdev 0.045); \textbf{(c)} Demo 2, unfiltered time domain, mean AUC 0.694 (stdev 0.086); \textbf{(d)} Demo 2, $\beta$-bandpassed time domain, mean AUC 0.615 (stdev 0.076).}
\label{fig:results.demo}
\end{figure}

\subsubsection{Demonstration 1: stimulating during target brain states}
The goal of the controller in this demonstration is to time stimulation to the occurrence of specific neural states. Since there is an assumed delay in the stimulation, the controller must forecast neural activity to decide whether to choose to stimulate. While the TBFM's test set $R^2$ values drop off quickly as a function of forecast horizon, this task requires an accurate forecast only until $t$=$40ms$. Here we continue to use a training set size of $5k$ but increase the test set size to $9k$. Since our sessions contain $15k$ trials, this simulates a realistic experimental setup where we gather data, train the TBFM, then attempt to use it for closed-loop control a short time later.

The target states are defined as LFP ranges across two channels, specifically the two channels where stimulation was applied during the experiment. We varied the margin around the targets (Section \ref{sec:methods-demo1}) $\delta$ to measure its effect on the Type 1, 2 error tradeoff. Specifically, we varied $\delta \in (-200, 200)$. The resulting ROC curves are in Figure \ref{fig:results.demo}(a,b).

The variation in $\delta$ caused the stimulator to drastically vary the proportion of trials where it stimulates. ``False positive'' here refers to the controller applying stimulation but the trajectory does not in-fact cross into the target zone, and ``false negative'' refers to cases where the controller does not apply stimulation but the trajectory does cross into the target zone. The mean area under curve (AUC) across sessions was $0.704$ (stdev $0.090$) for raw time domain data, and $0.652$ (stdev $0.045$) for beta-bandpassed time domain data. These results indicate that the TBFM achieved overall controllability. Example individual trials can be found in Appendix \ref{apx:demoexamples}.

\subsubsection{Demonstration 2: stimulating to drive brain activity towards target trajectories}
In our second demonstration, the goal of the controller is to drive the LFP trajectory towards a target trajectory. As above, the controller assumes the stimulation will be applied after a delay of $20ms$ and must decide whether to apply stimulation or not at the current time point based on how close a predicted trajectory will be to the reference trajectory. The bias towards over- or under-stimulation in this demonstration can be varied through the use of $\epsilon_p \in \mathbb{R}$, which alters the threshold distance between the predicted and reference paths at which stimulation will be applied. Specifically we varied $\epsilon_p \in (0, 30)$.

Figure \ref{fig:results.demo}(c,d) depicts the resulting ROC curves. The mean AUC was $0.721$ (stdev $0.090$) for raw time domain data and $0.637$ (stdev $0.061$) for beta-bandpassed data. Unsurprisingly, we see lower AUCs on this demonstration because of the longer forecast horizon which it requires. The model's $R^2$ on the test set was predictive of the AUC on this demo, as measured by the area under curve (AUC): the two exhibit a correlation with p=$1.01e^{-3}$. In general $R^2$ increases as we reduce the prediction horizon and loop latency, suggesting latency reductions may allow us to improve closed loop stimulation performance. Example individual trials can be found in Appendix \ref{apx:demoexamples}.

\section{Discussion and Future Work}
\label{sec:discussion}

We have presented a new approach to designing neural co-processors \cite{bryan.coproc,rao.braincoproc,rao.coproc} for model-based closed-loop neural stimulation. Our approach, based on the temporal basis function model (TBFM), is designed to efficiently learn the forward dynamics of a neural network while reducing some of the challenges associated with the application of machine learning (ML) methods to this domain. Specifically, our approach addresses sample efficiency, training time, and inference latency. We presented results showing that the TBF model can learn a single-trial forward model of the state-dependent neural dynamics in a cortical area after stimulation using existing optogenetics data from nonhuman primates. As we showed, this improved on models which ignored the state-dependence of stimulation effects. We then used this learned model to demonstrate controllers for closed-loop stimulation on two simulated applications.

\subsection{Clinical application of neural stimulation may be significantly improved by models of closed-loop, state-dependent stimulation responses}

Neural and behavioral responses to stimulation are often state-dependent \cite{kabir.statedep, bradley.statedep}. This has led to increased focus on precise characterization of that dependency in order to improve clinical outcomes of neural stimulation. For example, it may be necessary to account for state dependence to scale up visual prostheses which target the visual cortex \cite{dugue.statedep, walker.statedep}. Likewise, evidence suggests that plasticity induced by stimulation is significantly enhanced if it targets specific neural states \cite{bloch.opto}, which may enable e.g. improved recovery after stroke \cite{zanos.closedloop}. Despite this state-dependence suggesting a crucial path forward to improve clinical outcomes, it remains under-explored. New experiments and mathematical models like TBFMs are necessary to better characterize it.

Significant work has been done to model neural stimulation \cite{vertex.stim, nejatbakhsh.stimmodel} but crucially little has focused on single-trial state-dependent stimulation effects. Even less work has focused on multi-step spatiotemporal forecasting, with most previous work focusing predominantly on single-step forecasting \cite{shanechi.stimmodel}. Yet stimulation has multi-step effects which may be better predicted by models explicitly trained to perform multi-step predictions (Section \ref{sec:results.accuracy}). Unsurprisingly, models trained to perform single-step forecasts often perform worse on multi-step forecasting due to e.g., compounding errors \cite{venkatraman.multistep}, which we found to be true also with our task (results available upon request).

As we show (Section \ref{sec:results.accuracy}), common linear state space models (LSSMs) - even ones trained explicitly to perform single-trial, state-dependent, multi-step forecasts - do not perform well on our data. This suggests that nonlinear models like TBFMs and AE-LSTMs may be necessary for multi-step forecasts in some settings. Furthermore, we confirm clear state-dependence in the stimulation responses (Section \ref{sec:results.state}), and demonstrate that models which explicitly consider initial state perform significantly better on these data (Section \ref{sec:results.modelstatedep}). As a result, nonlinear models like TBFMs may be necessary to capture the state-dependence phenomenon of stimulation for clinical therapies.

\subsection{Efficient model-based controllers may enable the precise targeting of state-dependent stimulation responses, leading to new clinical approaches}

Once a multi-step forecast model for stimulation based on a TBFM has been learned from data, we can attempt to leverage it for model-based control. Despite the significant work done on stimulation modeling, crucially little work has been done to evaluate the feasibility of the models for real-time control of stimulation. A controller and model's inference latency must be evaluated and compared against the forecast horizon to determine if the model could plausibly be useful in a clinical setting. As we show (Section \ref{sec:results.inference}) even efficient LSSMs may not operate quickly enough for some settings, though some further optimizations may be possible.

Likewise, sample and training efficiency must be evaluated. Any adaptive, learning based system we attempt to deploy in a clinical setting must be practical and economic. Hours-long data collection procedures with each subject would significantly constrain the feasibility of such systems (likewise with training time).

As a result, future modeling efforts targeted at closed-loop stimulation ought to be evaluated against the metrics of latency, sample efficiency, and training time to better ensure their feasibility. By concentrating on these metrics, TBFMs seek to mitigate these problems while exhibiting sufficient accuracy to achieve control.

Once an efficient model has been identified, at least two types of control may be leveraged to improve clinical outcomes. The first targets stimulation to particular neural states, which may for example improve the quality of visual perception in a visual prosthesis \cite{dugue.statedep}, or plasticity during stroke recovery \cite{zanos.closedloop}. The second shapes the expected neural trajectory, shaping it to resemble, for example, healthy visual cortical activity for the purpose of a visual prosthesis \cite{tafazoli.acls}. As we show in Section \ref{sec:results.simulation}, a model may not need extreme accuracy to achieve such control, but rather may need only \textit{sufficient} accuracy while achieving the efficiencies enumerated above.

\subsection{Leveraging TBFMs and machine learning to identify and target novel forms of state dependence}
Future work identifying hidden forms of state-dependence may open new doors for novel stimulation therapies. Such identification may be plausible in a reinforcement learning (RL)\cite{barto.rl} setting where a therapeutic or behavioral effect is ``rewarded'', and the AI attempts to proactively identify the state-dependent stimulation effect which maximizes that reward (e.g., \cite{pan.coproc}). In this case the TBFM would act as a world model which is learned jointly with the control policy and rewards. With this approach, we would not decompose the identification of useful state-dependent effects into a separate subproblem which we would solve prior to attempting control, but rather would attempt to solve that problem implicitly during the experiments through exploration of the joint state and stimulation parameter space. However, a major concern with RL-based efforts is sample efficiency. Further investigation is needed to determine if an RL-based approach can be made sample efficient enough to perform well in experimental or clinical practice.

It also remains unclear how well an ML approach will adapt to the increase in dimensionality that comes with having a diverse space of stimulation parameters, which may include both the pulse parameters (amplitude, frequency, etc.) at a single electrode location as well the choice of locations. This will require varying stimulation parameters across a wider range than in the simple paired-pulse stimulation data we leveraged in this paper. Future experiments will also be needed to address practical issues associated with ML-based closed-loop stimulation systems, such as measuring and adapting to loop latency and jitter.


\section*{Code and data availability}
\label{sec:cad}
A $PyTorch$ implementation of the temporal basis function model can be found
at: \url{https://github.com/mmattb/py-tbfm}. It includes demo code based on
synthetic data. Optogenetic stimulation data used in this paper is available upon reasonable request to the authors.

\section*{Conflict of interest}
The authors of this work are not aware of any conflicts of interest
related to it.

\section*{Acknowledgements}
This work was supported by National Science Foundation (NSF) EFRI
grant no.\ 2223495, a Weill Neurohub Investigator grant, a CJ and Elizabeth Hwang endowed professorship (RPNR), National Institutes of Health grant nos.\ R01NS119593, R01MH125429 (FS, AYS), and the National Defense Science and Engineering Graduate (NDSEG) Fellowship Program NDSEG10578COMPSCI (MJB). Any opinions, findings, and conclusions or recommendations expressed in this material are those of the authors and do not necessarily reflect the views of the funders.

\section*{References}
\bibliographystyle{iopart-num}
\bibliography{refs}

\providecommand{\newblock}{}
\begin{thebibliography}{10}
\expandafter\ifx\csname url\endcsname\relax
  \def\url#1{{\tt #1}}\fi
\expandafter\ifx\csname urlprefix\endcsname\relax\def\urlprefix{URL }\fi
\providecommand{\eprint}[2][]{\url{#2}}

\bibitem{niparko.cochlear}
Niparko J~K 2009 {\em Lippincott Williams and Wilkins\/} (Oxford University Press)

\bibitem{weiland.retinal}
Weiland J~D and Humayun M~S 2014 {\em IEEE transactions on bio-medical engineering\/} {\bf 61}(5) 1412--1424 Retinal prosthesis

\bibitem{tomlinson.propr}
Tomlinson T and Miller L~E 2016 {\em Advances in experimental medicine and biology\/} {\bf 957} 367--388 Toward a proprioceptive neural interface that mimics natural cortical activity

\bibitem{tabot.tact}
Tabot G~A, Dammann J~F, Berg J~A, Tenore F~V, Boback J~L, Vogelstein R~J and Bensmaia S~J 2013 {\em Proceedings of the National Academy of Sciences\/} {\bf 110} 18279--18284 Restoring the sense of touch with a prosthetic hand through a brain interface ISSN 0027-8424 \urlprefix\url{https://www.pnas.org/content/110/45/18279}

\bibitem{tyler.tact}
Tyler D~J 2015 {\em Current opinion in neurology\/} {\bf 28}(6) 574--581 Neural interfaces for somatosensory feedback: bringing life to a prosthesis

\bibitem{dadarlat.tact}
Dadarlat M~C, O'Doherty J~E and Sabes P~N 2015 {\em Nature neuroscience\/} {\bf 18}(1) 138--144 A learning-based approach to artificial sensory feedback leads to optimal integration

\bibitem{sharlene.tact}
Flesher S~N, Collinger J~L, Foldes S~T, Weiss J~M, Downey J~E, Tyler-Kabara E~C, Bensmaia S~J, Schwartz A~B, Boninger M~L and Gaunt R~A 2016 {\em Science Translational Medicine\/} {\bf 8} 361ra141--361ra141 Intracortical microstimulation of human somatosensory cortex \urlprefix\url{https://www.science.org/doi/abs/10.1126/scitranslmed.aaf8083}

\bibitem{cronin.tact}
Cronin J~A, Wu J, Collins K~L, Sarma D, Rao R~P~N, Ojemann J~G and Olson J~D 2016 {\em IEEE transactions on haptics\/} {\bf 9}(4) 515--522 Task-specific somatosensory feedback via cortical stimulation in humans

\bibitem{nicolelis.bmbi}
O’Doherty J, Lebedev M, Ifft P, Zhuang K, Shokur S, Bleuler H and Nicolelis M 2011 {\em Nature\/} {\bf 479}(7372) 228--231 Active tactile exploration using a brain–machine–brain interface \urlprefix\url{https://doi.org/10.1038/nature10489}

\bibitem{bryan.coproc}
Bryan M~J, Jiang L~P and Rao R~P~N 2023 {\em Journal of Neural Engineering\/} {\bf 20} 036004 Neural co-processors for restoring brain function: results from a cortical model of grasping \urlprefix\url{https://dx.doi.org/10.1088/1741-2552/accaa9}

\bibitem{bolus.opto}
Bolus M, Willats A, Rozell C and Stanley G 2021 {\em Journal of neural engineering\/} {\bf 18}(3) State-space optimal feedback control of optogenetically driven neural activity

\bibitem{kahana.biomarker}
Kahana M~J~e~a 2021 {\em medRxiv\/} Biomarker-guided neuromodulation aids memory in traumatic brain injury \urlprefix\url{https://www.medrxiv.org/content/early/2021/05/22/2021.05.18.21256980}

\bibitem{berger.closedloop}
Berger T~W, Song D, Chan R~H~M, Marmarelis V~Z, LaCoss J, Wills J, Hampson R~E, Deadwyler S~A and Granacki J~J 2012 {\em IEEE transactions on neural systems and rehabilitation engineering : a publication of the IEEE Engineering in Medicine and Biology Society\/} {\bf 20}(2) 198--211 A hippocampal cognitive prosthesis: multi-input, multi-output nonlinear modeling and vlsi implementation

\bibitem{tafazoli.acls}
Tafazoli S, MacDowell C, Che Z, Letai K, Steinhardt C and Buschman T 2020 {\em Journal of Neural Engineering\/} {\bf 17} 056007 Learning to control the brain through adaptive closed-loop patterned stimulation \urlprefix\url{https://doi.org/10.1088/1741-2552/abb860}

\bibitem{castano.pd}
Castaño-Candamil S, Ferleger B~I, Haddock A, Cooper S~S, Herron J, Ko A, Chizeck H~J and Tangermann M 2020 {\em Frontiers in Human Neuroscience\/} {\bf 14} 421 A pilot study on data-driven adaptive deep brain stimulation in chronically implanted essential tremor patients \urlprefix\url{https://www.frontiersin.org/article/10.3389/fnhum.2020.541625}

\bibitem{little.park}
Little S~e~a 2016 {\em Journal of neurology, neurosurgery, and psychiatry\/} {\bf 87}(7) 717--21 Bilateral adaptive deep brain stimulation is effective in parkinson's disease.

\bibitem{bradley.statedep}
Bradley C, Nydam A~S, Dux P~E and Mattingley J~B 2022 {\em Nature Reviews Neuroscience\/} {\bf 23}(8) 459--475 State-dependent effects of neural stimulation on brain function and cognition

\bibitem{zanos.closedloop}
Zanos S 2019 {\em Cold Spring Harb Perspect Med.\/} {\bf 9}(11) Closed-loop neuromodulation in physiological and translational research.

\bibitem{kabir.statedep}
Kabir A, Dhami P, Dussault~Gomez M~A, Blumberger D~M, Daskalakis Z~J, Moreno S and Farzan F 2024 {\em Journal of Neuroscience\/} {\bf 44} Influence of large-scale brain state dynamics on the evoked response to brain stimulation ISSN 0270-6474 \urlprefix\url{https://www.jneurosci.org/content/44/39/e0782242024}

\bibitem{bloch.statedep}
Bloch J, Khateeb K, Silversmith D, O'Doherty J, Sabes P and Yazdan-Shahmorad A 2019 {\em Conference proceedings: Annual International Conference of the IEEE Engineering in Medicine and Biology Society. IEEE Engineering in Medicine and Biology Society. Conference\/} {\bf 2019} 6446--6449 Cortical stimulation induces network-wide coherence change in non-human primate somatosensory cortex

\bibitem{zanos.beta}
Zanos S, Rembado I, Chen D and Fetz E~E 2018 {\em Current Biology\/} Phase-locked stimulation during cortical beta oscillations produces bidirectional synaptic plasticity in awake monkeys

\bibitem{rao.braincoproc}
Rao R~P~N 2020 Brain co-processors: Using {AI} to restore and augment brain function {\em Handbook of Neuroengineering\/} ed Thakor N (Springer)

\bibitem{rao.coproc}
Rao R~P~N 2019 {\em Current Opinion in Neurobiology\/} {\bf 55} 142--151 Towards neural co-processors for the brain: Combining decoding and encoding in brain-computer interfaces

\bibitem{gallego.manifold}
Gallego J~A, Perich M~G, Miller L~E and Solla S~A 2017 {\em Neuron\/} {\bf 94} 978--984 Neural manifolds for the control of movement

\bibitem{furht.realtime}
Furht B, Grostick D, Gluch D, Rabbat G, Parker J and McRoberts M 1991 {\em Introduction to Real-Time Computing\/} (Boston, MA: Springer US) pp 1--35 ISBN 978-1-4615-3978-0 \urlprefix\url{https://doi.org/10.1007/978-1-4615-3978-0_1}

\bibitem{huang.realtime}
Huang J~Q and Lewis F 2003 {\em IEEE Transactions on Neural Networks\/} {\bf 14} 377--389 Neural-network predictive control for nonlinear dynamic systems with time-delay

\bibitem{azadeh.data}
Yazdan-Shahmorad A, Silversmith D~B, Kharazia V and Sabes P~N 2018 {\em eLife\/} {\bf 7} e31034 Targeted cortical reorganization using optogenetics in non-human primates ISSN 2050-084X \urlprefix\url{https://doi.org/10.7554/eLife.31034}

\bibitem{pan.coproc}
Pan M, Schrum M, Myers V, Biyik E and Dragan A~D 2024 {\em CoRR\/} {\bf abs/2406.06714} Coprocessor actor critic: A model-based reinforcement learning approach for adaptive brain stimulation. \urlprefix\url{http://dblp.uni-trier.de/db/journals/corr/corr2406.html#abs-2406-06714}

\bibitem{xu.control}
Chen X and Tomizuka M 2023 {\em Introduction to Modern Controls: with Illustrations in MATLAB and Python\/} (Ind.)

\bibitem{barto.rl}
Sutton R~S and Barto A~G 2019 {\em Reinforcement learning: An introduction, 2nd ed.\/} (The MIT Press)

\bibitem{moerland.mbrl}
Moerland T~M, Broekens J and Jonker C~M 2020 {\em CoRR\/} {\bf abs/2006.16712} Model-based reinforcement learning: {A} survey (\textit{Preprint} \eprint{2006.16712}) \urlprefix\url{https://arxiv.org/abs/2006.16712}

\bibitem{vertex.stim}
Thornton~C H~F and M K 2019 {\em Wellcome Open Research\/} The virtual electrode recording tool for extracellular potentials (vertex) version 2.0: Modelling in vitro electrical stimulation of brain tissue \urlprefix\url{https://doi.org/10.12688/wellcomeopenres.15058.1}

\bibitem{shanechi.stimmodel}
Yang Y, Qiao S, Sani O, Sedillo J, Ferrentino B, Pesaran B and Shanechi M 2021 {\em Nature Biomedical Engineering\/} {\bf 5}(4) 324--345 Modelling and prediction of the dynamic responses of large-scale brain networks during direct electrical stimulation \urlprefix\url{https://doi.org/10.1038/s41551-020-00666-w}

\bibitem{azadeh.data2}
Yazdan-Shahmorad A, Diaz-Botia C, Hanson T~L, Kharazia V, Ledochowitsch P, Maharbiz M~M and Sabes P~N 2016 {\em Neuron\/} {\bf 89}(5) 927--39 A large-scale interface for optogenetic stimulation and recording in nonhuman primates

\bibitem{azadeh.data3}
Ledochowitsch P and Yazdan-Shahmorad e~a 2015 {\em Journal of neuroscience methods\/} {\bf 256} 220--31 Strategies for optical control and simultaneous electrical readout of extended cortical circuits

\bibitem{azadeh.data4}
Yazdan-Shahmorad A, Silversmith D~B and Sabes P~N 2018 {\em Annual International Conference of the IEEE Engineering in Medicine and Biology Society. IEEE Engineering in Medicine and Biology Society. Annual International Conference\/} {\bf 2018} 5479--5482 Novel techniques for large-scale manipulations of cortical networks in non-human primates

\bibitem{azadeh.data5}
Yazdan-Shahmorad A, Diaz-Botia C, Hanson T, Ledochowitsch P, Maharabiz M and Sabes P 2015 {\em Progress in Biomedical Optics and Imaging - Proceedings of SPIE\/} {\bf 9305} Demonstration of a setup for chronic optogenetic stimulation and recording across cortical areas in non-human primates

\bibitem{bloch.opto}
Bloch J, Greaves-Tunnell A, Shea-Brown E, Harchaoui Z, Shojaie A and Yazdan-Shahmorad A 2022 {\em iScience\/} {\bf 25} 104285 Network structure mediates functional reorganization induced by optogenetic stimulation of non-human primate sensorimotor cortex ISSN 2589-0042 \urlprefix\url{https://www.sciencedirect.com/science/article/pii/S2589004222005557}

\bibitem{flint.lmps}
Flint R~D, Ethier C, Oby E~R, Miller L~E and Slutzky M~W 2012 {\em Journal of neurophysiology\/} {\bf 108} 18--24 Local field potentials allow accurate decoding of muscle activity.

\bibitem{busch.erpstatedep}
Busch N~A, Dubois J and VanRullen R 2009 {\em Journal of Neuroscience\/} {\bf 29} 7869--7876 The phase of ongoing eeg oscillations predicts visual perception ISSN 0270-6474 (\textit{Preprint} \eprint{https://www.jneurosci.org/content/29/24/7869.full.pdf}) \urlprefix\url{https://www.jneurosci.org/content/29/24/7869}

\bibitem{bedard.powerlaw}
Bédard C, Kröger H and Destexhe A 2006 {\em Physical review letters\/} {\bf 97}(11) Does the 1/f frequency scaling of brain signals reflect self-organized critical states?

\bibitem{papadopoulos.betaburst}
Papadopoulos S, Darmet L, Szul M~J, Congedo M, Bonaiuto J~J and Mattout J 2024 {\em Imaging Neuroscience\/} {\bf 2} 1--15 Surfing beta burst waveforms to improve motor imagery-based bci

\bibitem{guggenmos.latency}
Guggenmos D~J, Azin M, Barbay S, Mahnken J~D, Dunham C, Mohseni P and Nudo R~J 2013 {\em Proceedings of the National Academy of Sciences\/} {\bf 110} 21177--21182 Restoration of function after brain damage using a neural prosthesis \urlprefix\url{https://www.pnas.org/doi/abs/10.1073/pnas.1316885110}

\bibitem{murphy.ml}
Murphy K~P 2022 {\em Probabilistic Machine Learning: An introduction\/} (MIT Press) \urlprefix\url{probml.ai}

\bibitem{kim.dyncorrespond}
Kim N~H, Xie Z and van~de Panne M 2020 Learning to correspond dynamical systems (\textit{Preprint} \eprint{1912.03015}) \urlprefix\url{https://arxiv.org/abs/1912.03015}

\bibitem{shi.deepkoopman}
Shi H and Meng M~Q~H 2022 Deep koopman operator with control for nonlinear systems (\textit{Preprint} \eprint{2202.08004}) \urlprefix\url{https://arxiv.org/abs/2202.08004}

\bibitem{jung.lstmmpc}
Jung M, {da Costa Mendes} P~R, Önnheim M and Gustavsson E 2023 {\em Engineering Applications of Artificial Intelligence\/} {\bf 123} 106226 Model predictive control when utilizing lstm as dynamic models ISSN 0952-1976 \urlprefix\url{https://www.sciencedirect.com/science/article/pii/S0952197623004104}

\bibitem{dugue.statedep}
Dugu{\'e} L, Marque P and VanRullen R 2011 {\em Journal of Neuroscience\/} {\bf 31} 11889--11893 The phase of ongoing oscillations mediates the causal relation between brain excitation and visual perception ISSN 0270-6474 \urlprefix\url{https://www.jneurosci.org/content/31/33/11889}

\bibitem{walker.statedep}
Allison-Walker T~J, Ann~Hagan M, Chiang~Price N~S and Tat~Wong Y 2020 Local field potential phase modulates neural responses to intracortical electrical stimulation. {\em Annual International Conference of the IEEE Engineering in Medicine and Biology Society. IEEE Engineering in Medicine and Biology Society. Annual International Conference\/} vol 2020 pp 3521--3524

\bibitem{nejatbakhsh.stimmodel}
Nejatbakhsh A, Fumarola F, Esteki S, Toyoizumi T, Kiani R and Mazzucato L 2024 {\em Physical review research\/} {\bf 5} Predicting the effect of micro-stimulation on macaque prefrontal activity based on spontaneous circuit dynamics.

\bibitem{venkatraman.multistep}
Venkatraman A, Hebert M and Bagnell J 2015 {\em Proceedings of the AAAI Conference on Artificial Intelligence\/} {\bf 29} Improving multi-step prediction of learned time series models \urlprefix\url{https://ojs.aaai.org/index.php/AAAI/article/view/9590}

\bibitem{karamanakos.fcsmpc}
Karamanakos P and Geyer T 2020 {\em IEEE Transactions on Power Electronics\/} {\bf 35} 7434--7450 Guidelines for the design of finite control set model predictive controllers

\bibitem{kraskov.mutualinformation}
Kraskov A, St\"ogbauer H and Grassberger P 2004 {\em Phys. Rev. E\/} {\bf 69}(6) 066138 Estimating mutual information \urlprefix\url{https://link.aps.org/doi/10.1103/PhysRevE.69.066138}

\bibitem{efron.permutationtest}
Efron B and Hastie T 2021 {\em Computer Age Statistical Inference, Student Edition: Algorithms, Evidence, and Data Science\/} Institute of Mathematical Statistics Monographs (Cambridge University Press)

\bibitem{gretton.hsic}
Gretton A, Fukumizu K, Teo C, Song L, Sch\"{o}lkopf B and Smola A 2007 A kernel statistical test of independence {\em Advances in Neural Information Processing Systems\/} vol~20 ed Platt J, Koller D, Singer Y and Roweis S (Curran Associates, Inc.) \urlprefix\url{https://proceedings.neurips.cc/paper_files/paper/2007/file/d5cfead94f5350c12c322b5b664544c1-Paper.pdf}

\end{thebibliography}

\FloatBarrier
\newpage
\section{Appendix}
\subsection{Dataset and Processing}
\label{apx:dataset}

Our data comes from a previously published excitatory optogenetic
stimulation study \cite{azadeh.data}. That study involved the induction
of plasticity using pairs of stimulation pulses which were spaced in
time, with spacing ranging from 10-100ms. The pulses had a width of 5ms,
and the pulse pairs were typically spaced 200ms apart.

The subjects were two adult male rhesus macaques, one 8 years old and the other 7. Their somatosensory (S1) and primary motor (M1) cortices
were optogenetically photosensitized using a viral-mediated expression of the C1V1 opsin. Neural data were recorded using a $\mu$ECoG array,
sampled at $24kHz$ using a Tucker-Davis Technologies system (Florida, USA). Photosensitivity was
verified using fluorescent imaging of the eYFP marker, and by optogenetic stimulation tests.

Optical stimulation was applied by way of fiber optic (Fiber Systems,
Texas, USA). The light source was a $488nm$ laser (PhoxX 488–60, Omicron-Laserage, Germany). The optics were co-located with $\mu$ECoG
electrodes which were chosen according to the
results of epifluorescent imaging and measured physiological response to stimulation (additional details in \cite{azadeh.data,azadeh.data2}). During stimulation, the subjects watched cartoons while sitting in a primate chair with their head position fixed. 

A total of 70 experimental sessions were performed, each
containing a combination of resting state recordings, single stimulation
pulse recordings to validate the equipment and photosensitivity, and paired pulse recordings.
A typical session included 1 or 2 hours of paired pulse recordings, and
that data forms the primary portion of our work. We present results for 40
of these sessions. We rejected the remaining 30 sessions based on a
number of criteria; some sessions satisfied multiple of these criteria:
\begin{enumerate}
    \item 17 sessions used a 143ms timing between pulse
pairs, and to keep a consistent format in our dataset we opted to reject those sessions
    \item 15 sessions contained no useful paired pulse stimulation data
\end{enumerate}

We rejected electrodes on the basis of
high impedance or low signal to noise ratio, measured from 
experimental blocks during which stimulation was not applied. The number of usable electrodes in a session varied between 42 and 94, with an average of 78.6 (stdev $13.2$). Additional details in \cite{bloch.opto}. Data was recorded at a sample rate of $24kHz$, was loss-pass filtered at 500Hz with Chebychev filtering for anti-aliasing, and downsampled to $1kHz$. We filter each channel using an IIR notch filter at 60, 180, and 300Hz to remove line noise (filter quality Q of 20, 60, 100 respectively). Finally, we downcast the data from 64bit to 32bit, finding that saves significant memory without a loss in model accuracy.

To compute time-varying power spectral densities (PSDs) we use short time FFT (STFFT).
For beta band we use $100ms$ Gaussian windows, with a standard deviation of $10ms$, and a stride of $5ms$. This results in trial lengths of 36 time steps. We keep the bins centered on $10Hz$, $20Hz$, and $30Hz$. For high gamma band we use a window of $20ms$ and keep the bands centered at $50$, $100$, and $150Hz$

We present additional results for bandpassed time domain data below. For both beta (13-30Hz) and high gamma (67-200Hz) we apply a first order Butterworth filter.

Train and test dataset splits are typically performed using random
sampling, but we chose to split instead based on the time position
within the session. Specifically the training set comes from the first 5k pulse pairs, and the test set comes from the last. During an experiment our model will need to generalize from early to late session stimulations, and plasticity changes the stimulation response over
the course of the session \cite{bloch.opto}. Hence, our splitting
method ensures we are testing our ability to generalize across the
session. This splitting method does not appear to change results to
a statistically significant degree overall (p=$0.57$; see
\ref{apx:crossval} for full analysis).

\subsection{TBFM technical detail}
\label{apx:arch}

This section provides additional details on the specific hyperparameters and architectural details used for TBFM results throughout the paper. 

\subsubsection{Stimulation descriptor}
The stimulation descriptor consists of 3 vectors of length $T_h$. The first is a ``clock vector'' with value 0.0 at t=$0ms$, and 1.0 at t=$184ms$, with linear interpolation between the two. This clock vector allows the model to have an explicit time dependency by keeping track of elapsed time since the start of the current trial. The other two vectors are one-hot encoded and express the onset of the two stimulation pulses at different time points in the experiment. In our optical stimulation dataset, there were only 3 types of stimulation descriptors corresponding to the inter-pulse intervals $10ms$, $30ms$, $100ms$ as described above. Note that in future experiments stimulation parameters such as pulse width, location, and timing may exhibit a greater variety. For those cases this descriptor would be adapted to encode those parameters as well.

\subsubsection{Basis function generator}
Our basis function generator is an MLP which receives as input the stimulation descriptors, and outputs $B \in \mathbb{R}^{b,T_h}$: a matrix of temporal basis functions. Each row $B_{i,*} \in \mathbb{R}^{T_h}$ contains one temporal basis function. We found that an MLP with four layers, a width of four, and hyperbolic tangent (Tanh) nonlinearities worked well across all applications.

\subsubsection{Training}
The hyperparameter $\lambda$ controls regularization. Through a small amount of hyperparameter optimization we found $\lambda = 0.05$ to work well across sessions.

Our model is implemented in $PyTorch$ and leverages $PyTorch$'s error backpropagation and $AdamW$ optimizer. We found that a learning rate of $2e^{-4}$ worked sufficiently across experiments.

\subsubsection{Compilation}
\label{apxsubsec:compilation}
Since the trial-specific half of the network (the basis weight estimator) is linear we can compile the model down to a linear model after training, provided that our stimulation parameter set is a small discrete set. This leads to more memory and compute efficient inference and further decreases latency.

This allows for faster execution by discarding the nonlinear basis generator. The bases are a function of stimulation parameters and not of any particular trial's initial state. That allows us to save the matrix of basis functions $B$ for each unique stimulation parameterization, and then to discard the nonlinear MLP portion of the model. In future experiments, the space of stimulation parameters will be larger, but will often still be a small discrete set and therefore compilation will still be feasible.

We use these pre-trained basis functions along with the linear weight estimator (see Figure \ref{fig:tfm_arch}), resulting in pre-compiled models, each model with its own set of basis functions. This leads to faster execution (see Section  \ref{sec:results.inference}) and facilitating the model's performant implementation on a range of general purpose processors and microcontrollers supporting fast linear operations. 

\subsection{Technical detail of simulated applications}
\label{apx:demos}

For both simulations we leverage our TBFM and a simple form of finite control set model predictive control (FCS-MPC; \cite{karamanakos.fcsmpc}). A control set is the set of possible controls of a system from which the controller can choose. In our case these are stimulation parameters, and specifically the binary choice to stimulate at t=$40ms$ or not on a given trial. The controller's cost function can be used to specify our relative tolerance for false positives versus false negatives, for representing the trade-offs between power efficiency and stimulation efficacy, and to capture the risk of side effects from over-stimulation.

\subsubsection{Demonstration 1: stimulating during target brain states}
As described in Section \ref{sec:methods-demo1}, this controller targets stimulation to the predicted occurrence of specific features of neural activity. We defined the target in terms of the LFP values at the two channels nearest to the stimulation sites. For each channel we bin the LFPs of resting data from the channel into quartiles, thus providing four possible target ranges to choose from per channel. That results in 16 possible targets in total. Our trials are sourced from the test set used to evaluate our stimulation model. We randomly choose $50\%$ of these trials to be ``should stimulate'' trials. For those we choose the target range to be the quartile through which the data from the trial passes. For the rest of the trials (``do not stimulate'' trials), the target range is some other randomly chosen target. See Figure \ref{fig:demos}(a) for a visual depiction of examples of these two types of trials.

For this demonstration, the cost function the controller minimizes is as follows: stimulation has a cost of 0 if the forecast enters the target range at t=$40ms$, and 1 otherwise. Not stimulating has a cost of 0 if the forecast does not enter the target, and 1 otherwise. The optimal controller minimizes this cost function and stimulates only if the forecast enters the target range at t=$40ms$.

\subsubsection{Demonstration 2: stimulating to drive brain activity towards target trajectories}
For this demonstration we generate reference trajectories by calculating a weighted average of test set trajectories, beginning at the stimulation onset t=$40ms$. Such a procedure ensures that all reference trajectories will be within the distribution of test set trajectories, and will therefore be approximately achievable through stimulation. The weights for the average are drawn randomly for each trial, resulting in unique runways and trajectories for each. Figure \ref{fig:demos}(b) depicts an example visually.

The controller stimulates only if the $L_2$ distance between the reference and the TBFM's prediction is within a threshold value $\epsilon_p \in \mathbb{R}$. Similar to Demonstration 1, varying this threshold allows for a bias in favor of under- or over-stimulation.

The cost function is defined as:

$$
C = \begin{cases}
  \begin{tabular}{rl}
    $0$, &
    $\left [ L_2(T^*_{r,0}, \hat{T}_{0}) + L_2(T^*_{r,1}, 
    \hat{T}_{1}) \right ] < \epsilon_p$ \\
    $\infty$, & \text{otherwise}
  \end{tabular}
\end{cases}
$$

\noindent where $L_2$ is the mean squared error loss function, $T^*_{r,0}$ and $T^*_{r,1}$ are the reference trajectories for channels 0 and 1, $\hat{T}_0$ and $\hat{T}_1$ are the forecasted trajectories for the two channels assuming stimulation is applied.

\subsection{Reference models}
\label{apx:referencemodels}

\subsubsection{Linear state space model}
While there are a wide variety of LSSMs, some simple versions were previously demonstrated for modeling neural stimulation \cite{bolus.opto, shanechi.stimmodel}. These are commonly learned using the Kalman Filter, though they can also be trained using other methods such as backpropagation. In discrete time these simple LSSMs are specified as:

\begin{equation}
\begin{split}
x_{k+1} & = Ax_k + Bu_k \\
y_k & = Cx_k
\end{split}
\end{equation}

Here $x$ is a latent state, $u$ is the control input (e.g. stimulation parameters), and $y$ is the prediction. Forward prediction can be performed by specifying an initial latent state $x_0$ and autoregressing forward in time.

We leverage this simple formulation by providing our stimulation descriptor as input. We estimate the initial state $x_0$ using the Moore-Penrose pseudoinverse of $C$ and the last value of the runway. We train the model explicitly using backpropagation to perform multisstep forecasting and therefore do not use a Kalman Filter. The $L_2$ prediction loss is used to train the three matrices of parameters $A, B, C$.

\subsubsection{LSTM-based nonlinear dynamical systems model}
\label{apx:refdyns}
We base this AE-LSTM model on the more complex long short-term memory (LSTM) model \cite{murphy.ml}, a nonlinear neural network for representing neural dynamics as a dynamical system with external inputs (in our case, stimulation pulses as inputs). The LSTM-based model uses an autoencoder architecture to lift the LFP data into a latent space, and predicts the effect of stimulation using an estimated dynamical system defined in the latent space. The model predicts the change in neural activity between time steps, and performs forward prediction through a simple first-order integration.

In this demo case the latent space is 96 dimensions - equal to the number of electrodes in the ECog array. We chose this dimensionality to ensure we were not losing critical information when transforming into the latent space, but note that a lower dimensionality may provide similar results without penalty due to the inherently low dimensional nature of neural data.

\begin{figure}
\centering
\begin{subfigure}[c]{0.99\textwidth}
    \includegraphics[width=\textwidth]{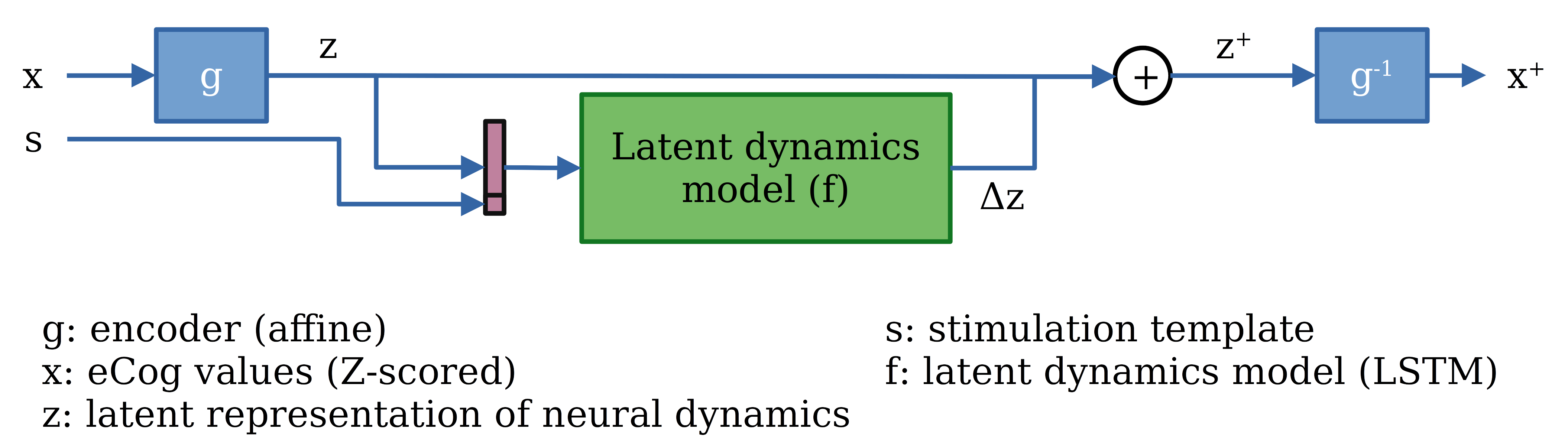}
\end{subfigure}
\caption{\textbf{(a) LSTM-based Dynamical Systems model (AE-LSTM)}
Here the $+$ superscript refers to a single step prediction. Details in text body.}
\label{fig:odelstm}
\end{figure}

As in TBFM, we leverage the stimulation descriptor, which we concatenate to the estimated latent state of the system $z$. The LSTM estimates a single step change in the system, which is summed into the latent state to make a single step prediction $z^+$. To make multi-step predictions the single step prediction is passed back into the dynamics model repeatedly. Thus: multi-step predictions are made using the first-order Euler integration method. See Figure \ref{fig:odelstm}.

We train the model on the same data sets as the TBFM. We crop random
sub-windows of size 60ms which we split into a 20ms runway and a
40ms prediction horizon. Like TBFM we leverage a multi-step MSE loss
function. Finally, we validate the model on the test set by performing
the full 164ms multistep prediction.

Training uses a tripartite loss function which can be compared to \cite{kim.dyncorrespond}:

\begin{itemize}
    \item an autoencoder
loss $\text{MSELoss}(x, g^{-1}(g(x)))$
    \item a dynamics prediction loss $\text{MSELoss}(z_t + \Delta z, z_{t+1})$ for all time steps in our window. Note however that we
    unroll predictions to make multi-step predictions, feeding our
    prediction back to the dynamics model at each step. The dynamics
    loss is a multi-step loss.
    \item a nearest-neighbor loss, which attempts to force all LFP values to align near each other in latent space so the model can take advantage of similar dynamics across channels and space. The
    simplest way to do that is to force the latent states to be centered at 0; hence $\text{MSELoss}(\overline{z}, 0)$.
\end{itemize}

\begin{figure}
	\centering
    \includegraphics[width=0.6\textwidth]{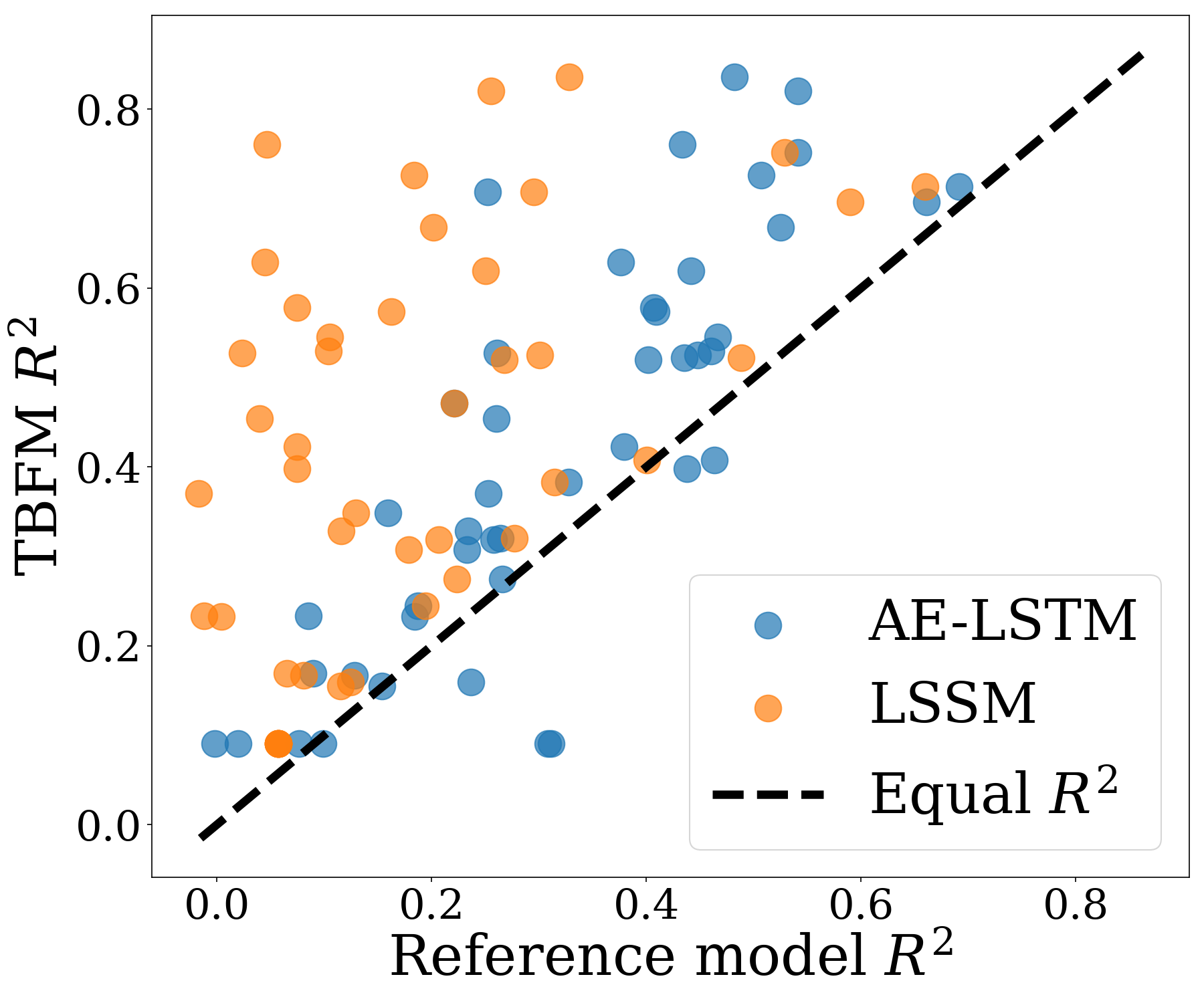}
    \hfill
\caption{\small \textbf{Comparison of $R^2$ on test set, reference models versus temporal basis function models (TBFMs)} The TBFM model's $R^2$ exceeds the autoencoder with LSTM dynamics (AE-LSTM) model on 32 of 40 sessions. It exceeds the linear state space model (LSSM) on 39 of 40.}
\label{fig:odelstmresults}
\end{figure}

The performance results compared to the TBFM can be found in Figure \ref{fig:odelstmresults}.

\subsubsection{Spectral decomposition reference model}
\label{apx:refpca}
As mentioned in Section \ref{sec:additive}, an alternative method for
building a TBFM is through the use of PCA. Using time domain data as an example: with our window size of
184 time steps we can consider every channel's LFP values as being
a single point in $\mathbb{R}^{184}$. Applying PCA we derive a set of
principle components (PCs) which minimize the residuals for the given
PC count. These are temporal basis functions. Once we have done this, we need to predict the coordinates
of unseen examples in the reduced PC space using the runway, precisely
as we do with our proposed TBFM approach. We attempted this with both a linear and MLP weight estimator.

While this solution is elegant and familiar it results in reduced
performance, particularly on unseen examples - i.e. it overfits. Generally the reduction on the test set is $\approx$$10\%$ of the $R^2$ value.

\subsection{Hypothesis testing for state dependence}
\label{apx:statedep}
As explained in Section \ref{sec:results.state}, we identify within-channel state dependency by statistically testing for the influence of initial state $x_{40}$ on the subsequent stimulation response $\boldsymbol{r}_t, t \in (45,70)$. We first reject $x_{40}$ values greater than 5 stdevs from the mean to ensure our test is not dominated by outliers. For each stimulation trial $\boldsymbol{x}_t$ we identify the baseline trial $\boldsymbol{b}_t$ which minimizes $|x_{40} - b_{40}|$; that is: we identify the baseline trial whose initial state is closest to the initial state of the stimulation trial. $\boldsymbol{r}_t$ is then defined as $\boldsymbol{x}_t - \boldsymbol{b}_t$. We restrict subsequent analysis to $t \in (45,70)$ under the presumption that statistical tests will be more able to identify state dependence in the time span of highest magnitude stimulation response.

Under these assumptions state dependence would imply that $\boldsymbol{r}_t$ is statistically dependent on initial state $x_{40}$. More formally, we specify:

$$
\boldsymbol{r}_t = \boldsymbol{a}_t + \boldsymbol{f}(t, x_{40}), t \in (45,70)
$$

where $\boldsymbol{a}_t$ is the state-agnostic portion of stimulation response and $\boldsymbol{f}$ represents any state dependency. We estimate $\boldsymbol{a}_t$ as the trial-averaged stimulation response ($\boldsymbol{r}_t$). And thus we can analyze:

$$
\boldsymbol{f}(t, x_{40}) = \boldsymbol{r}_t - \mathbb{E}[\boldsymbol{r}_t] , t \in (45,70)
$$

\begin{figure}
	\centering
     \begin{subfigure}[c]{0.99\textwidth}
		\centering
		\includegraphics[width=\textwidth]{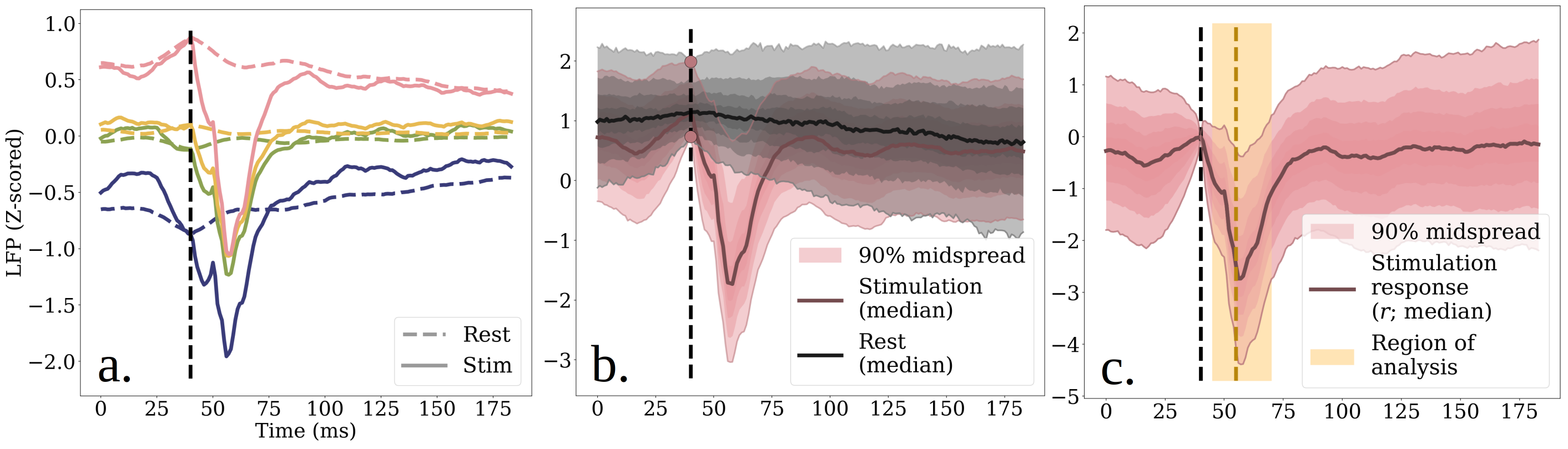}
	\end{subfigure}
	\hfill
\caption{\small \textbf{Distributions of stimulation and resting trials, single session and channel} (a) Trial-averaged resting (dashed lines) and stimulation (solid lines) trials. Each are binned into quartiles for graphing. Note that we do not perform binning for our statistical analysis. (b) $90\%$ midspread and median of stimulation and resting trials for the high (pink) initial state as depicted in (a). (c) Midspread and median of stimulation responses ($\boldsymbol{r_t}$) for the same state.}
\label{fig:statedepapx}
\end{figure}

Figure \ref{fig:statedepapx} gives a visual reference of the distributions involved for a single session and channel.

\begin{figure}
	\centering
     \begin{subfigure}[c]{0.72\textwidth}
		\centering
		\includegraphics[width=\textwidth]{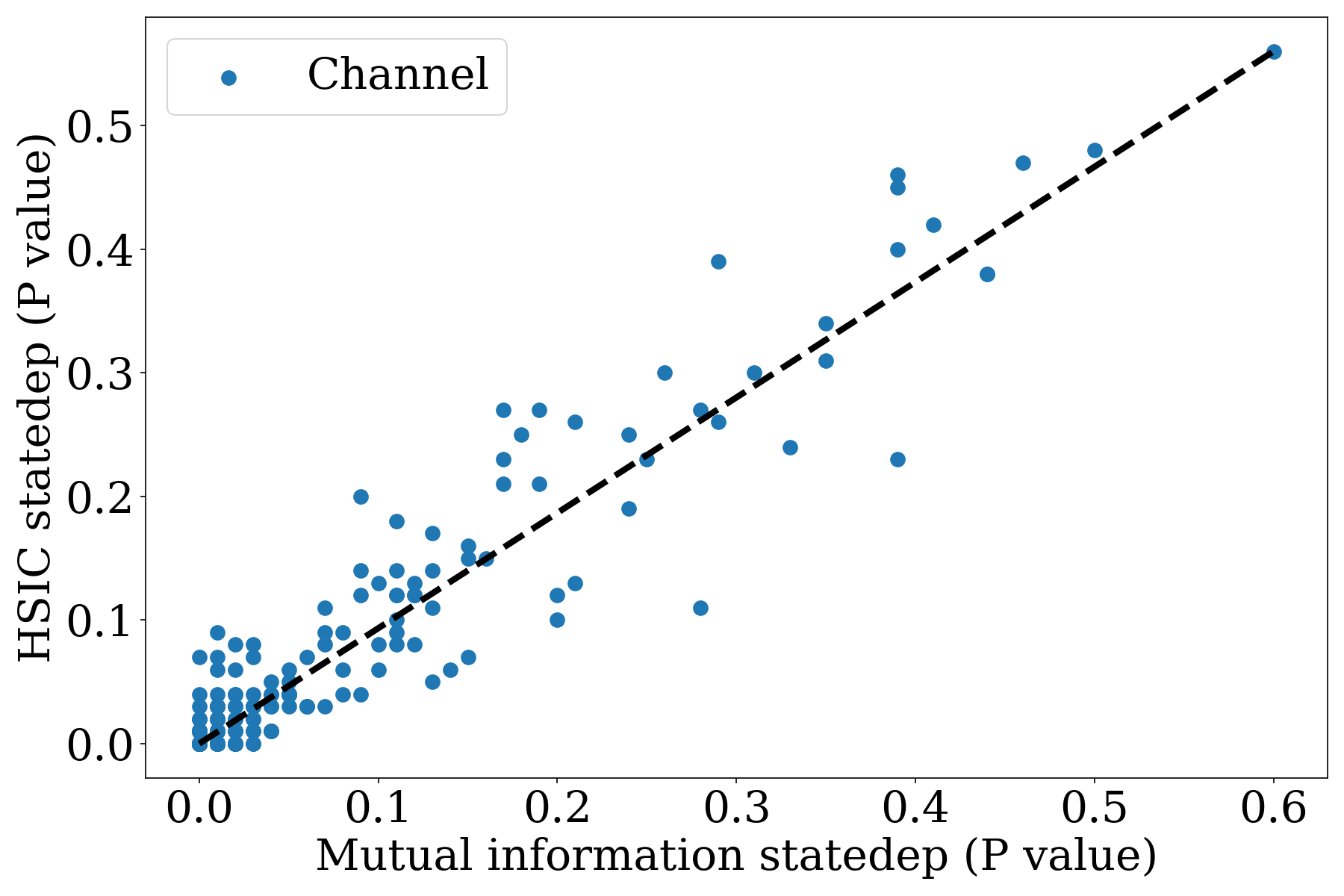}
	\end{subfigure}
	\hfill
\caption{\small \textbf{Mutual information-based state dependence measure versus Hilbert-Schmidt Independence Critereon (HSIC) state dependence measure}
The two measures are highly correlated: $0.96$ $(P \approx0.0)$, and thus appear to capture, largely, the same thing.}
\label{fig:hsicvmi}
\end{figure}

Here, we hypothesis test whether the right hand side is statistically independent of $x_{40}$. If we fail to reject the null hypothesis we cannot detect a dependence of $\boldsymbol{f}$ on $x_{40}$ and therefore cannot detect state dependence.

We use two different statistical tests: 1.) the Kraskov–Stögbauer–Grassberger (KSG) estimator for mutual information (KSG-MI) \cite{kraskov.mutualinformation} together with a permutation test \cite{efron.permutationtest}; and the Hilbert-Schmidt Independence Criteron (HSIC) \cite{gretton.hsic}.

For the KSG-MI test we permute $\boldsymbol{f}$ values to assign them randomly to $x_{40}$ values, then subsequently apply the KSG estimator. We perform 1000 such permutations and our P-value is calculated as the percentage of the permutations where the resulting KSG-MI estimate meets or exceeds the KSG-MI estimate on unshuffled data. We report results for a P threshold of $0.05$. The HSIC-based test provides a P-value as well, and we likewise report results for a $0.05$ threshold.

The two methods give largely the same results: that $\approx97.5\%$ of channels exhibit statistically significant state dependence. The P values of the two methods have a Pearson correlation of $0.96$ $(P \approx0.0)$. See Figure \ref{fig:hsicvmi}.

\subsection{All results across filtering methods}
\label{apx:filtering}
Here we present full results across three filtering methods: 
high gamma bandpass time domain (50-200Hz), beta band time-frequency domain, and high gamma band time-frequency domain. Bandpasses were applied using a first order Butterworth filter. These filtering methods target biologically relevant bands which are widely studied in literature. Presenting such filtered data may more easily allow a future co-processor to specify target dynamics in terms of those signal features.

Time-varying power spectral density (PSD) was calculated using short time FFT (STFFT). We use Gaussian windows, stride of $5ms$. For beta band we use a window size of $100ms$ and a standard deviation of 10, and for high gamma $20ms$ and a standard deviation of 5. This results in trial lengths of 36 time steps, a runway of 4 steps, and stimulation onset at 4 time steps later. We use the 10, 20, and 30hz bands for beta band, and the 100-200hz bands for high gamma.

Table \ref{table:results} contains a summary of results for both
bandpassed and time-frequency domain data. In general the $R^2$ on the beta
bandpassed (presented earlier) data exceeds that of the high gamma bandpassed data, and both
have $R^2$ lower than the not-bandpassed time domain data. Nevertheless, the trained models largely track the mean responses
and state-dependent mean responses in both cases, indicating that the model
successfully identifies those responses despite the decreased
predictability of the series in general. Models trained on time-frequency domain data perform better overall than those based on bandpassed data, though not in terms of the state-dependent $R^2$.

\begin{figure}
	\centering
	\begin{subfigure}[c]{0.50\textwidth}
		\centering
		\includegraphics[width=\textwidth]{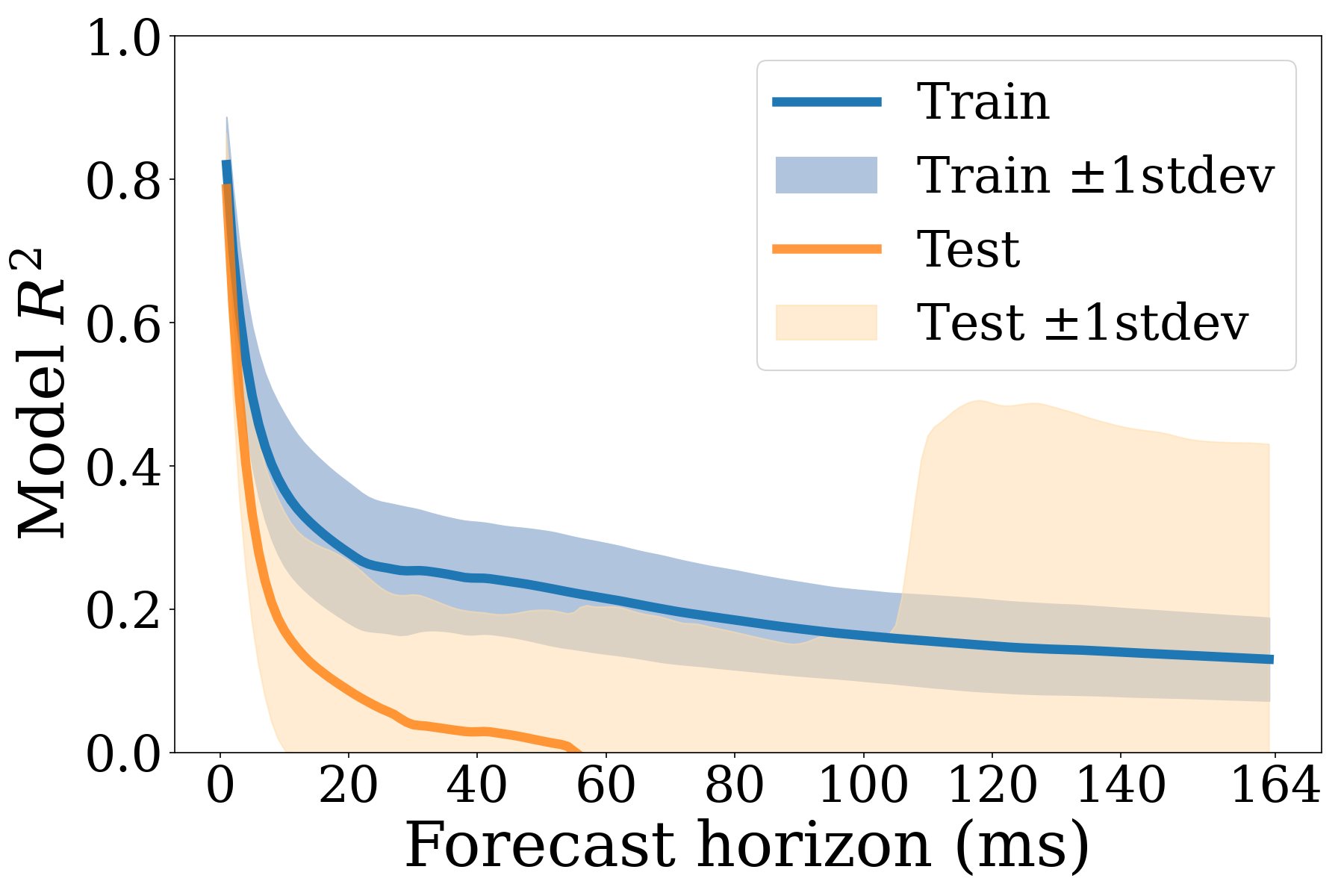}
		\caption{High gamma bandpassed}
	\end{subfigure}
	\hfill
 	\begin{subfigure}[c]{0.49\textwidth}
		\centering
		\includegraphics[width=\textwidth]{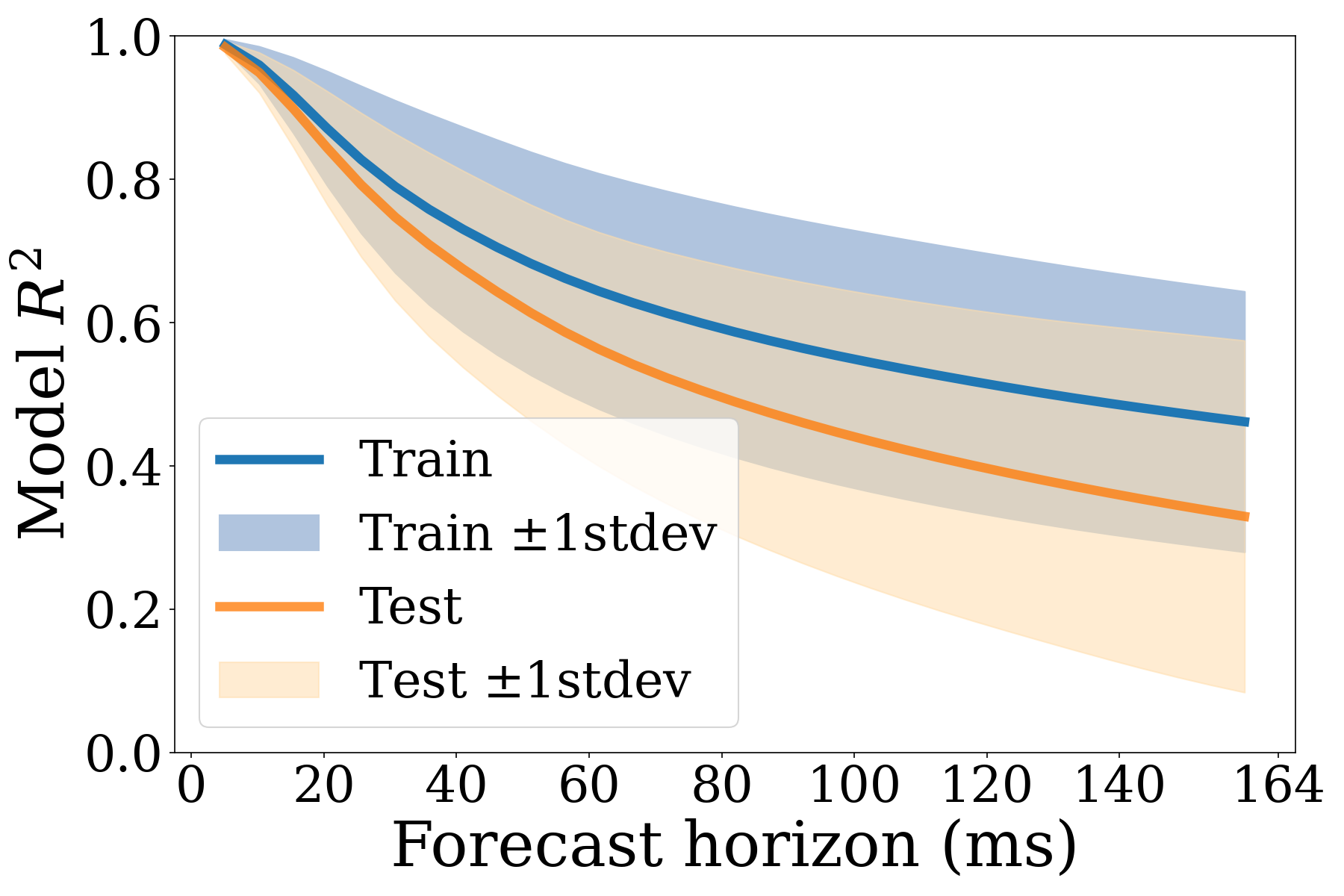}
		\caption{Beta band time-frequency domain}
	\end{subfigure}
	\hfill
	\begin{subfigure}[c]{0.50\textwidth}
		\centering
		\includegraphics[width=\textwidth]{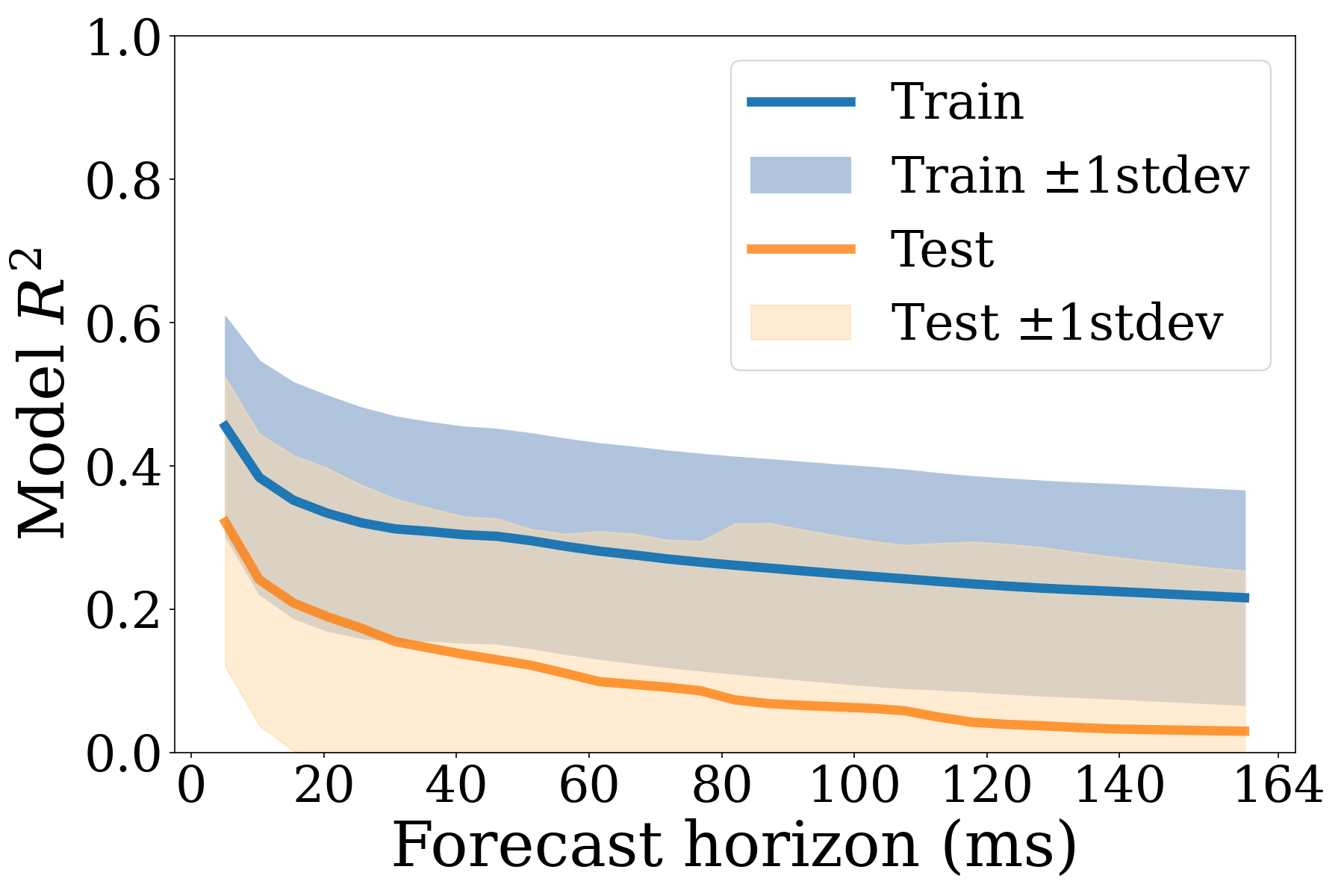}
		\caption{ High gamma band time-frequency domain}
	\end{subfigure}
	\hfill
\caption{\small \textbf{$R^2$ versus prediction horizon, mean of 40 sessions} See Table \ref{table:results} for statistics.}
\label{fig:bptrend}
\end{figure}

\begin{figure}
	\centering
	\begin{subfigure}[c]{0.49\textwidth}
		\centering
		\includegraphics[width=\textwidth]{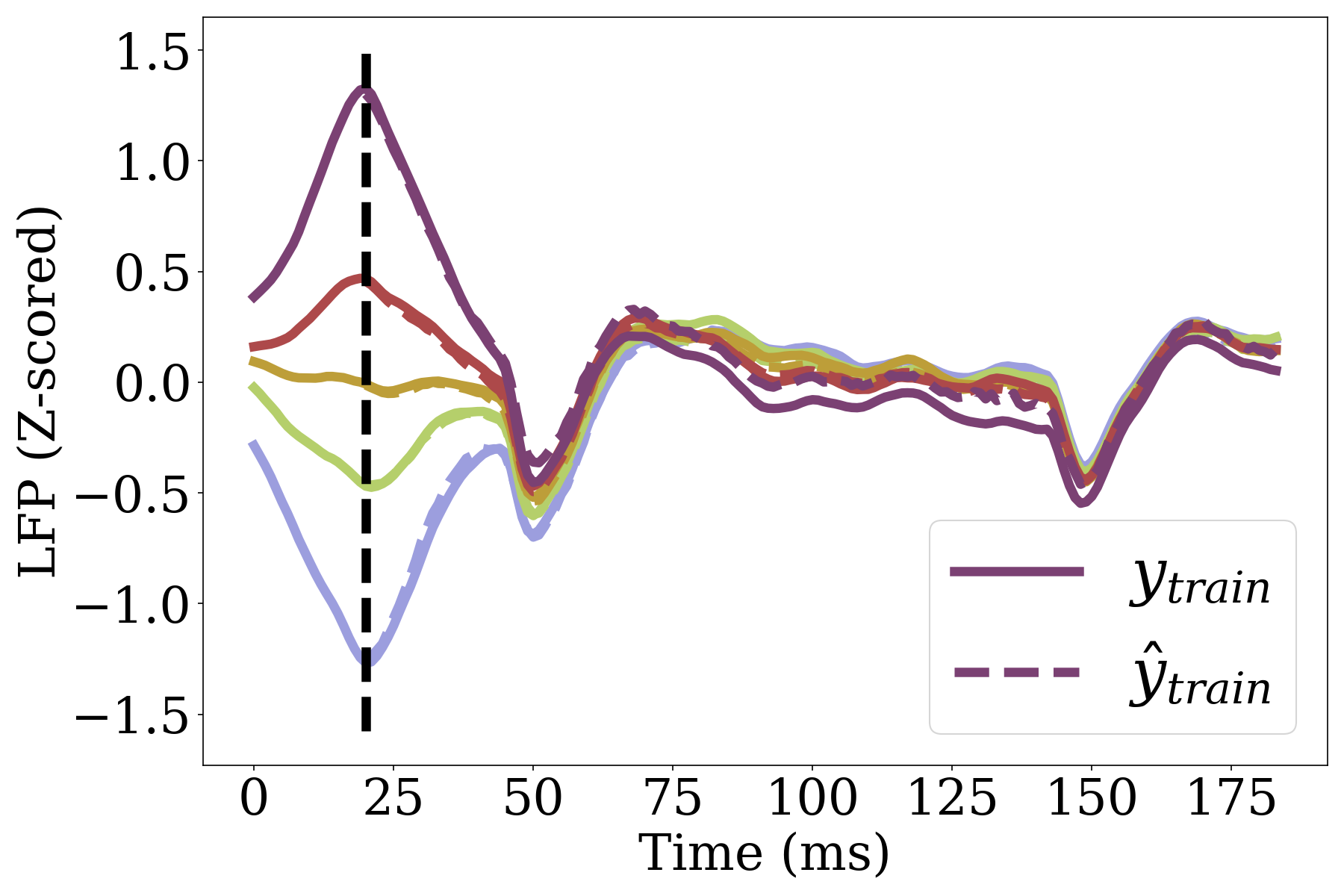}
		\caption{}
	\end{subfigure}
	\hfill
	\begin{subfigure}[c]{0.50\textwidth}
		\centering
		\includegraphics[width=\textwidth]{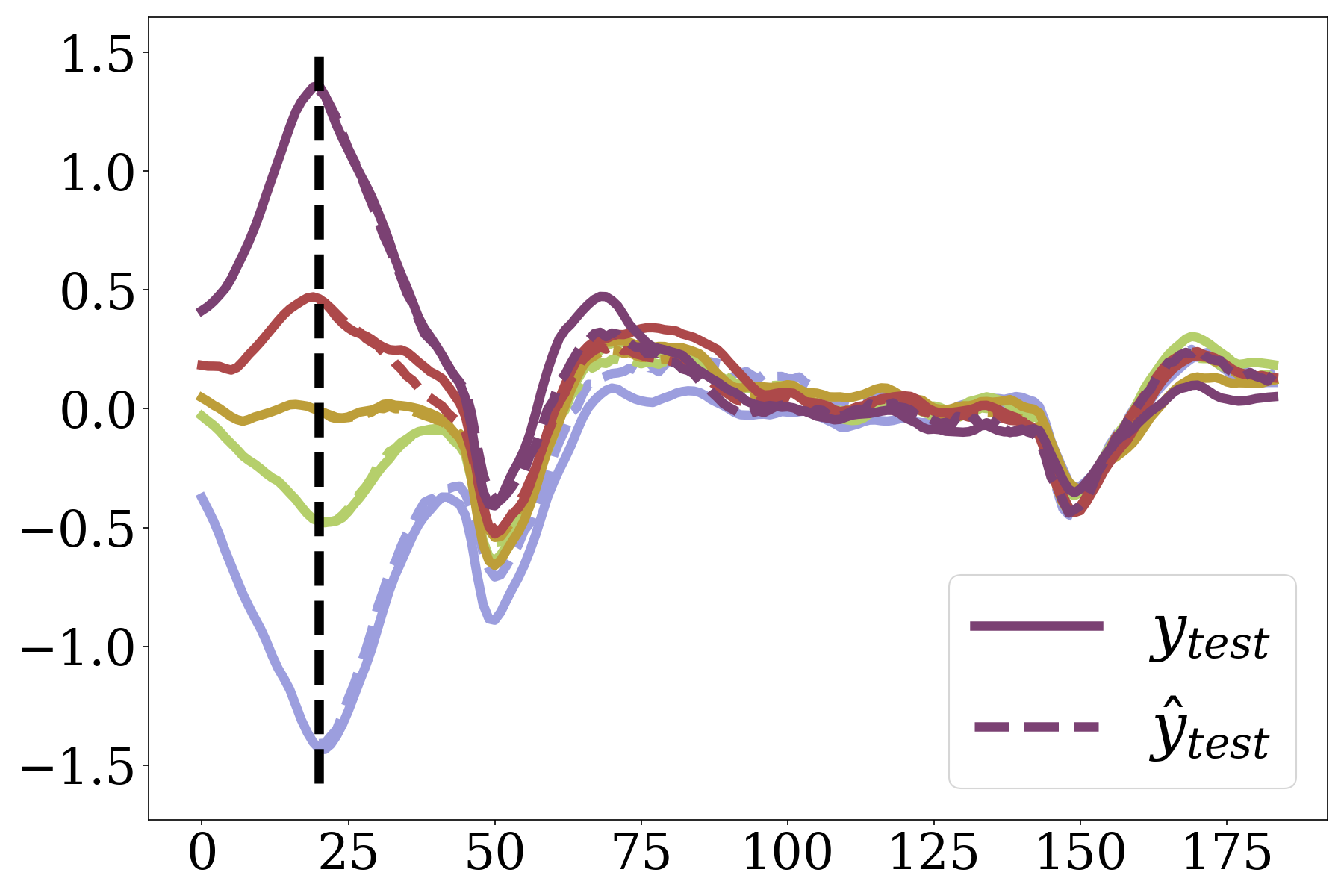}
		\caption{}
	\end{subfigure}
	\hfill
	\begin{subfigure}[c]{0.49\textwidth}
		\centering
		\includegraphics[width=\textwidth]{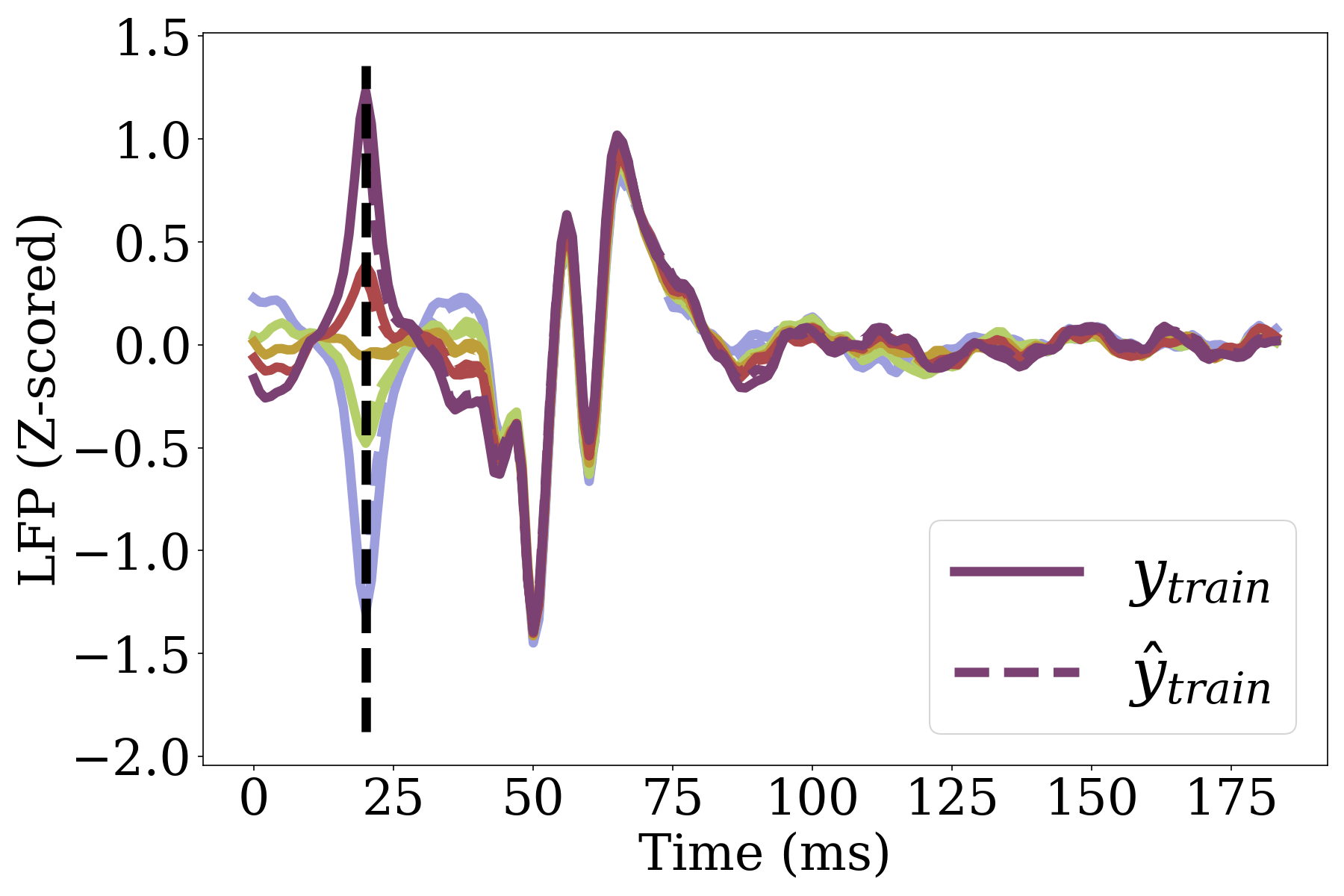}
		\caption{}
	\end{subfigure}
	\hfill
	\begin{subfigure}[c]{0.50\textwidth}
		\centering
		\includegraphics[width=\textwidth]{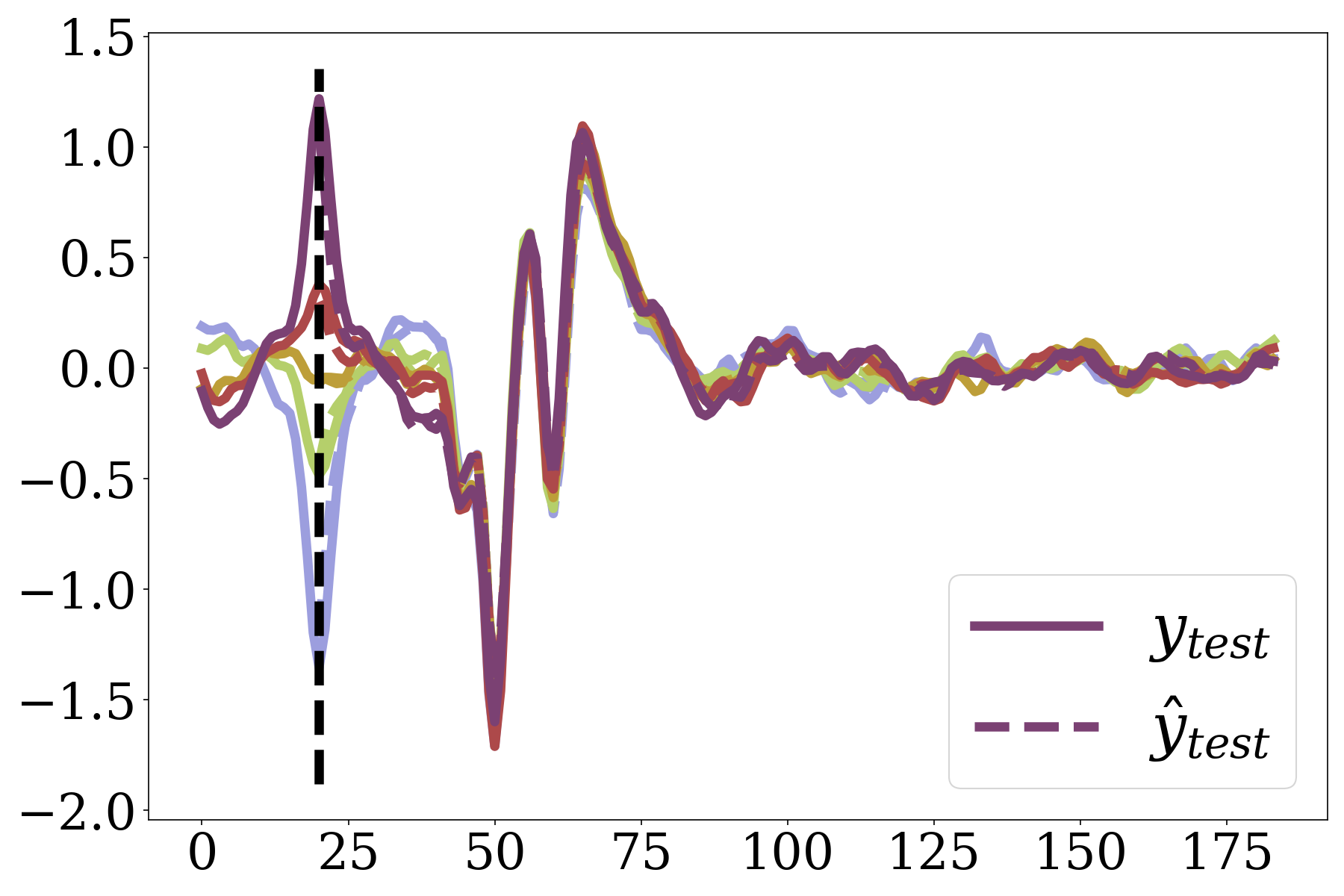}
		\caption{}
	\end{subfigure}
	\hfill
\caption{\small \textbf{Example state-dependent predictions for bandpassed data} Initial state binned at t=$20ms$ into five states and averaged within each bin; example single channel and single session.
\emph{Top row:} beta bandpassed \textbf{(a)} Training set; \textbf{(b)} test set \emph{Bottom row:} high gamma bandpassed
\textbf{(c)} Training set; \textbf{(d)} test set.}
\label{fig:bpstatedeppreds}
\end{figure}

\begin{figure}
	\centering
	\begin{subfigure}[c]{0.49\textwidth}
		\centering
		\includegraphics[width=\textwidth]{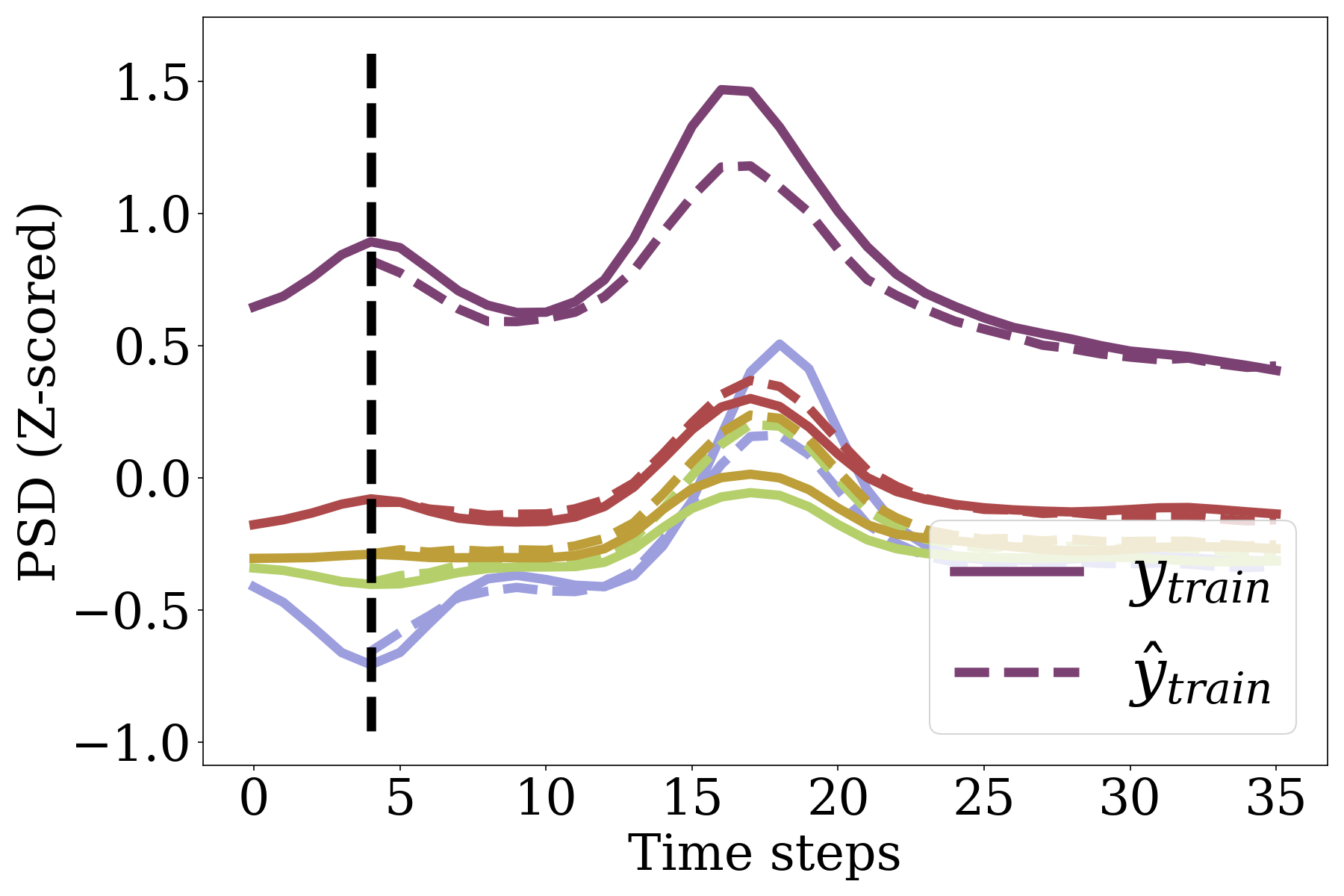}
		\caption{}
	\end{subfigure}
	\hfill
	\begin{subfigure}[c]{0.50\textwidth}
		\centering
		\includegraphics[width=\textwidth]{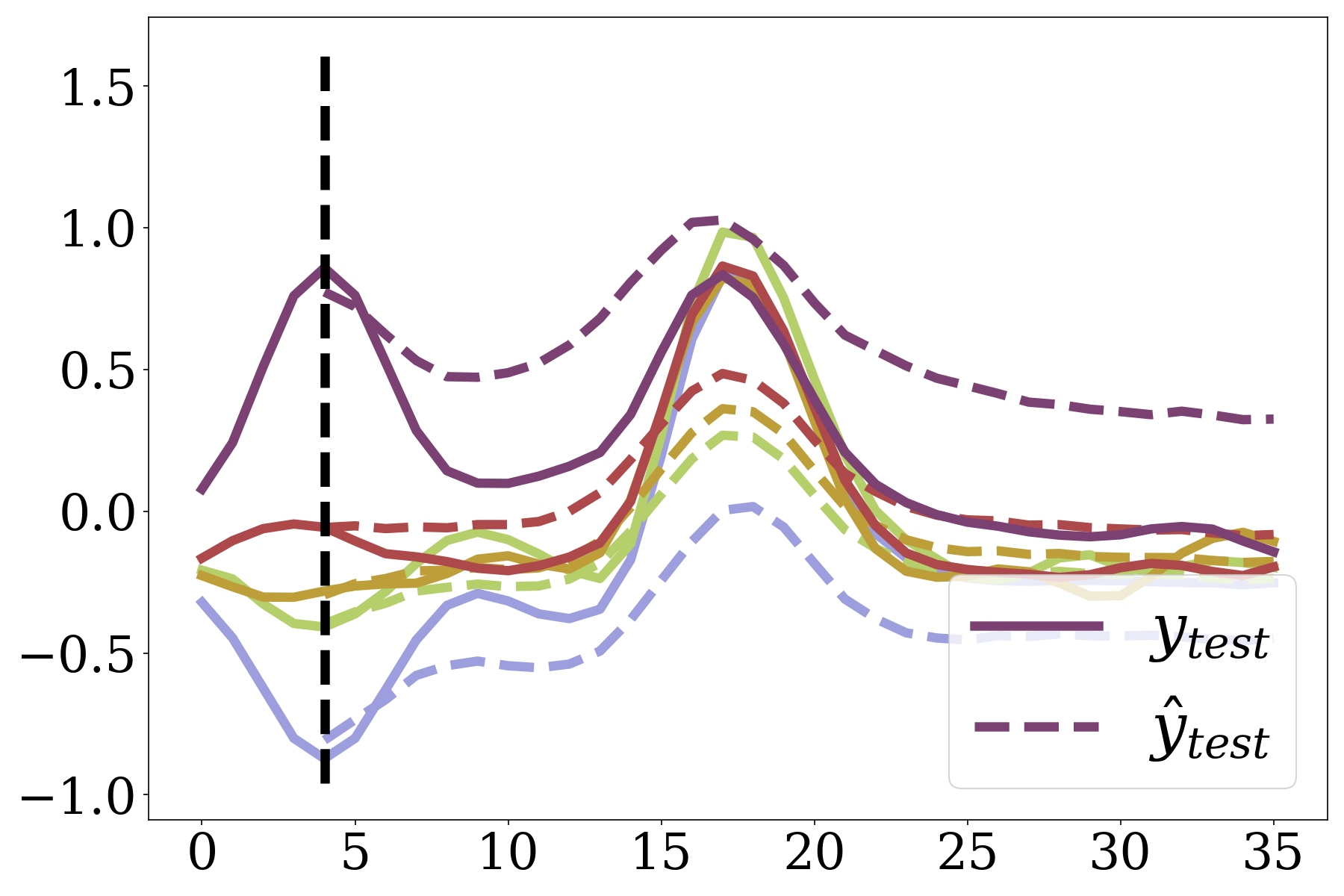}
		\caption{}
	\end{subfigure}
	\hfill
	\begin{subfigure}[c]{0.49\textwidth}
		\centering
		\includegraphics[width=\textwidth]{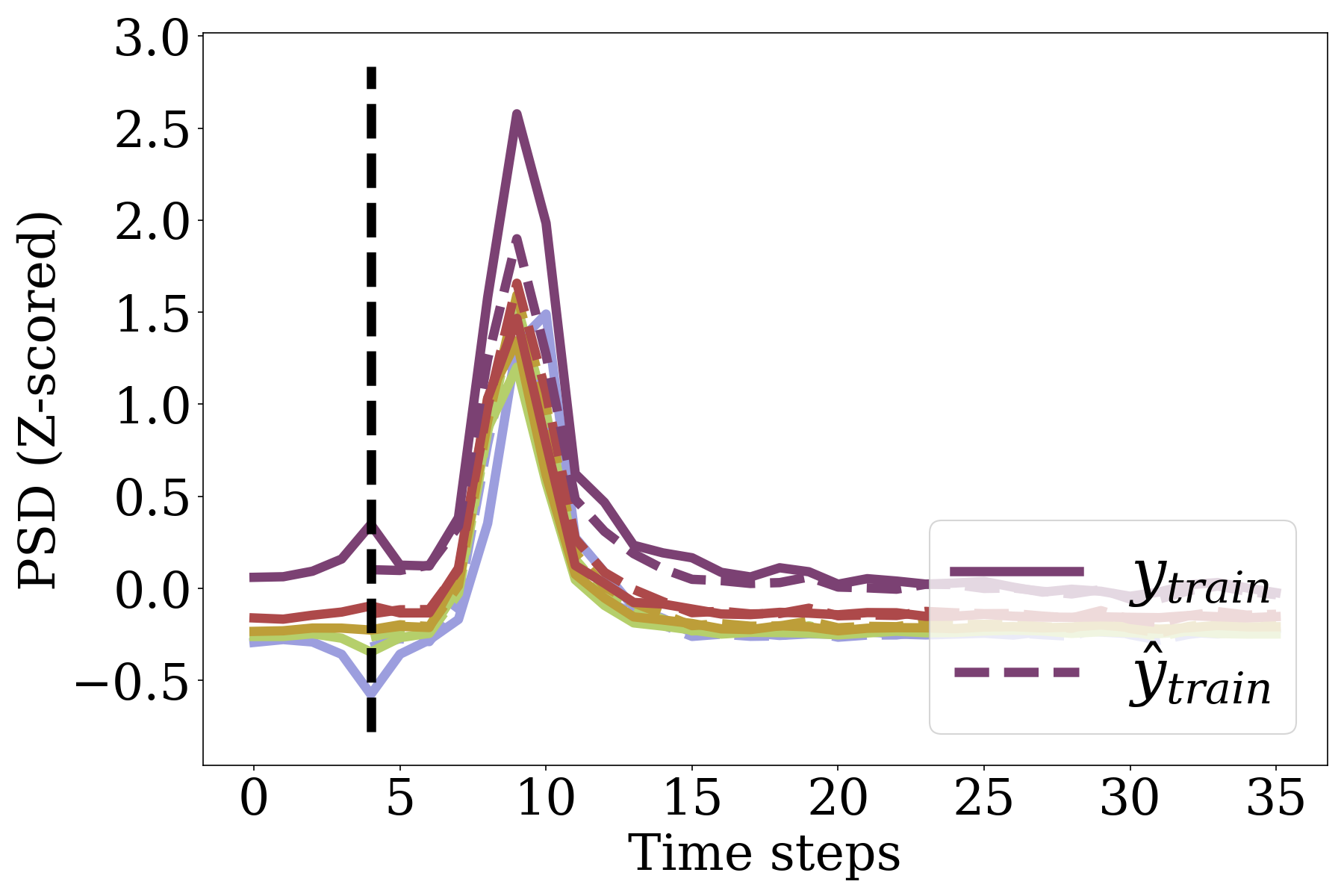}
		\caption{}
	\end{subfigure}
	\hfill
	\begin{subfigure}[c]{0.50\textwidth}
		\centering
		\includegraphics[width=\textwidth]{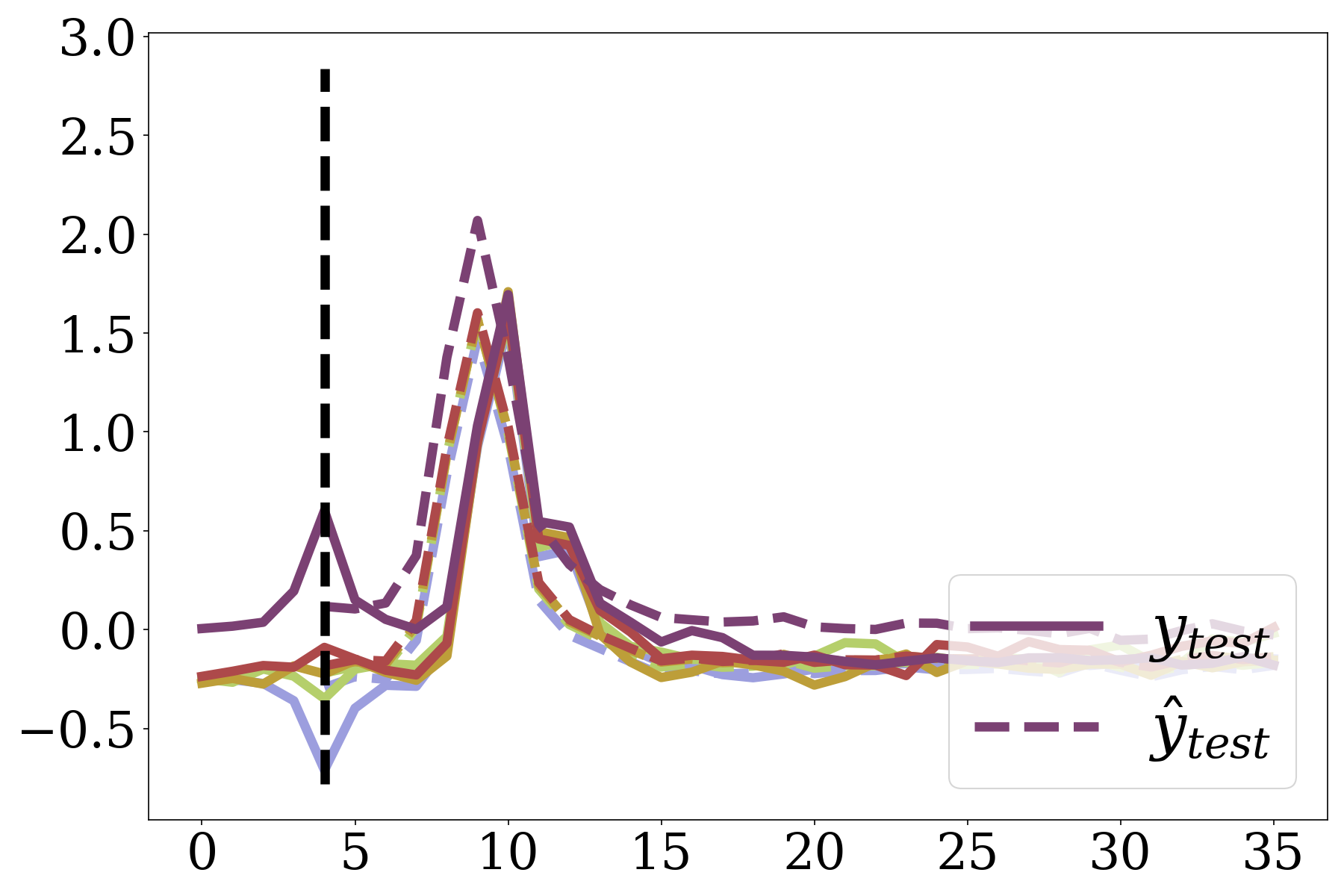}
		\caption{}
	\end{subfigure}
	\hfill
\caption{\small \textbf{Example state-dependent predictions, time-frequency domain data} Initial state binned at the 4th time step (approx t=$20ms$) into five states and averaged within each bin; example single channel and single session.
\emph{Top row:} beta band, time-frequency domain \textbf{(a)} Training set; \textbf{(b)} test set \emph{Bottom row:} high gamma band, time-frequency domain
\textbf{(c)} Training set; \textbf{(d)} test set.}
\label{fig:fdstatedeppreds}
\end{figure}

In Figure \ref{fig:bptrend} we see that shorter-horizon predictions exhibit higher $R^2$ across all filtering methods, as expected. Co-processors built on
data in these bands may need to restrict themselves to shorter horizon
predictions, and perhaps to lower loop latencies as a result. In Figures \ref{fig:bpstatedeppreds}, \ref{fig:fdstatedeppreds} we see example state dependent predictions for example sessions and channels.

\begin{table}
\centering
\caption{Summary of results on bandpassed time domain (BP) and PSD data; averages across 40 sessions}
\begin{tabular}{l|rr|}
\cline{2-3}
                                                  & \multicolumn{2}{c|}{\textbf{Metric}}                                                                                     \\ \hline
\multicolumn{1}{|l|}{\textbf{Data set}}           & \multicolumn{1}{l|}{$R^2$ 164ms forward prediction} & \multicolumn{1}{l|}{State-dependent $R^2$} \\ \hline
\multicolumn{1}{|l|}{High gamma BP (train)} & \multicolumn{1}{r|}{0.135 ($\pm$0.059)}         & 0.768 ($\pm$0.134)                           \\ \hline
\multicolumn{1}{|l|}{High gamma BP (test)}  & \multicolumn{1}{r|}{0.000 ($\pm$0.082)}        & 0.590 ($\pm$0.236)                        \\ \hline
\multicolumn{1}{|l|}{Beta PSD (train)} & \multicolumn{1}{r|}{0.459 ($\pm$0.182)}         & 0.541 ($\pm$0.204)                           \\ \hline
\multicolumn{1}{|l|}{Beta PSD (test)}  & \multicolumn{1}{r|}{0.373 ($\pm$0.176)}         & 0.488 ($\pm$0.197)                           \\ \hline
\multicolumn{1}{|l|}{High gamma PSD (train)} & \multicolumn{1}{r|}{0.396 ($\pm$0.195)}         & 0.685 ($\pm$0.177)                           \\ \hline
\multicolumn{1}{|l|}{High gamma PSD (test)}  & \multicolumn{1}{r|}{0.282 ($\pm$0.183)}        & 0.612 ($\pm$0.177)                           \\ \hline
\end{tabular}
\label{table:results}
\end{table}

\subsection{Low frequencies dominate dataset, model captures mostly those}
\label{apx:lowfreq}

\begin{figure}
	\centering
     \begin{subfigure}[c]{0.49\textwidth}
		\centering
		\includegraphics[width=\textwidth]{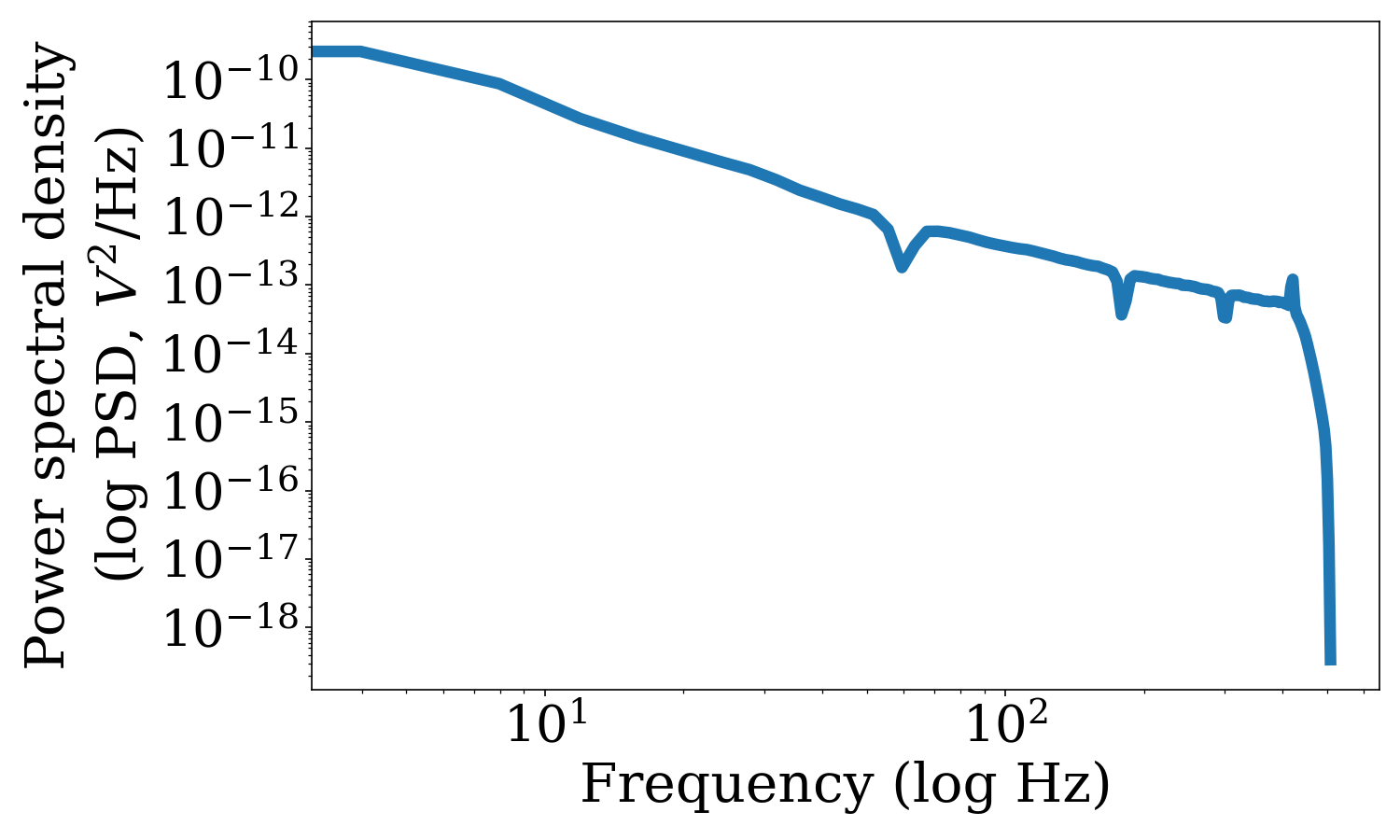}
		\caption{}
	\end{subfigure}
	\hfill
   	\begin{subfigure}[c]{0.49\textwidth}
		\centering
		\includegraphics[width=\textwidth]{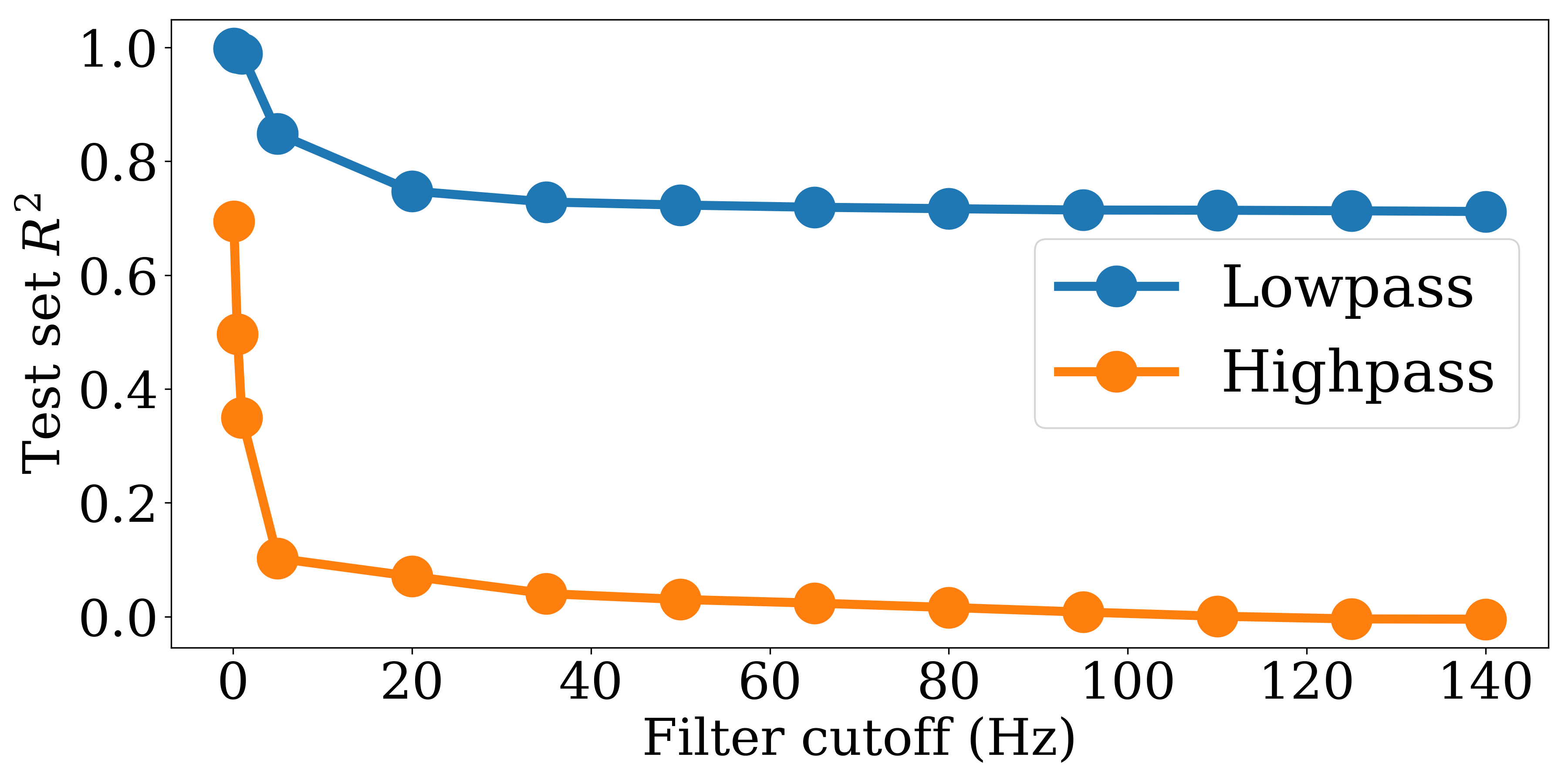}
		\caption{Test set $R^2$ of TBFM model v filtering, time domain data, average all sessions}
	\end{subfigure}
	\hfill
\caption{\small \textbf{1/f frequency power scaling law leads to higher model performance on lower frequencies.} (a) Power spectral density (PSD) of resting state data, session, channel, and trial averaged, log-log scale. Linear trend indicates the characteristic 1/f power scaling law. Notches are due to noise removal filters at 60, 180, and 300Hz. (b) Application of high- and low-pass filters indicate that removing the lowest frequencies from time domain data has the largest effect on model performance. The TBFM models' $R^2$s are driven mostly by lower frequencies.}
\label{fig:psd}
\end{figure}

Resting state data across all sessions and channels exhibit the characteristic 1/f power scaling law. This scaling law appears commonly in neural data and indicates that lower frequencies exhibit significantly higher power than higher frequencies \cite{bedard.powerlaw}. Averaging the spectra cross all sessions and channels and graphing on a log-log scale we see the characteristic straight line associated with this scaling law (Figure \ref{fig:psd}a).

Unsurprisingly then it appears that the $R^2$ of TBFMs are driven more by lower frequency components than higher frequencies when they are trained on time domain data (Figure \ref{fig:psd}b). We see this manifest in Appendix \ref{apx:filtering} where TBFMs exhibit highest $R^2$ on less filtered data, moderate $R^2$ on beta bandpassed data, and comparatively low $R^2$ on gamma bandpassed data.

\FloatBarrier
\newpage
\subsection{Modeling multiple stimulation parameters}
\label{apx:multisession}

\begin{table}[h]
\begin{tabular}{r|rr|rr|}
\cline{2-5}
\multicolumn{1}{l|}{}                                & \multicolumn{2}{l|}{\textbf{Test set $R^2$}}                                          & \multicolumn{2}{l|}{\textbf{Test set mean $R^2$}}                                     \\ \hline
\multicolumn{1}{|r|}{\textbf{Session stim interval}} & \multicolumn{1}{l|}{Single}                             & \multicolumn{1}{l|}{Multi}        & \multicolumn{1}{l|}{Single}                             & \multicolumn{1}{l|}{Multi}        \\ \hline
\multicolumn{1}{|r|}{10ms}                           & \multicolumn{1}{r|}{\cellcolor[HTML]{FFFFC7}0.713} & 0.671                         & \multicolumn{1}{r|}{0.969}                         & \cellcolor[HTML]{FFFFC7}0.984 \\ \hline
\multicolumn{1}{|r|}{30ms}                           & \multicolumn{1}{r|}{0.422}                         & \cellcolor[HTML]{FFFFC7}0.938 & \multicolumn{1}{r|}{\cellcolor[HTML]{FFFFC7}0.696} & 0.521                         \\ \hline
\multicolumn{1}{|r|}{100ms}                          & \multicolumn{1}{r|}{0.159}                         & \cellcolor[HTML]{FFFFC7}0.259 & \multicolumn{1}{r|}{0.946}                         & \cellcolor[HTML]{FFFFC7}0.984 \\ \hline
\end{tabular}
\caption{Summary of multisession TBFM results versus corresponding single session models}
\label{table:multis}
\end{table}

Figure \ref{apx:multisession} provides mean predictions for train and test sets across the three sessions. Table \ref{table:multis} summarizes $R^2$ values.

\begin{figure}[h]
	\centering
    \begin{subfigure}[c]{0.33\textwidth}
		\centering
		\includegraphics[width=\textwidth]{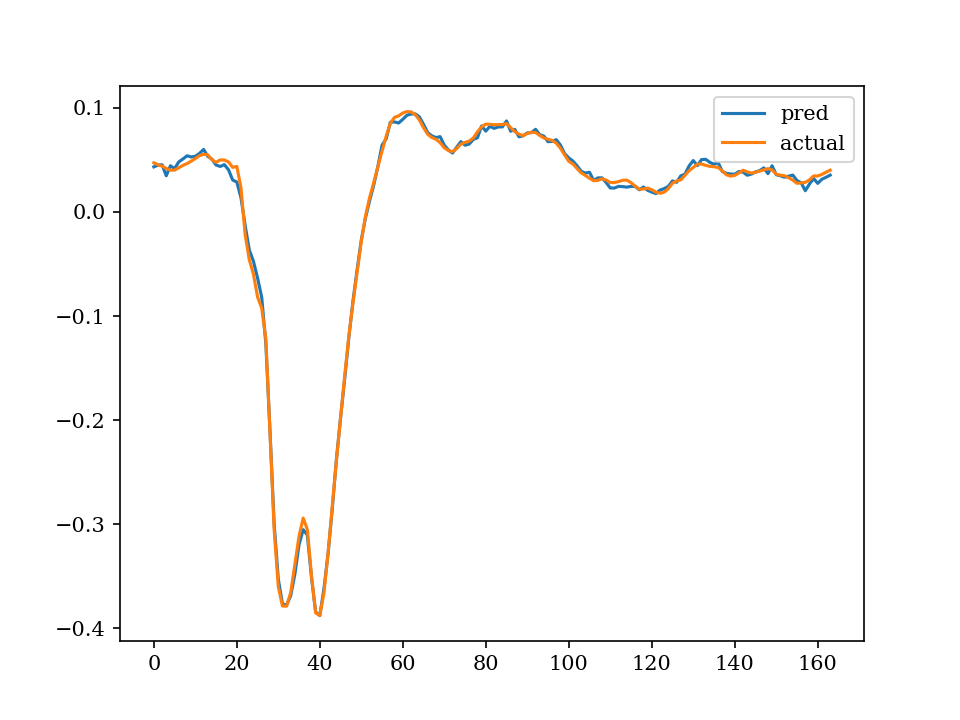}
		\caption{Train, 10ms interval}
	\end{subfigure}
	\hfill
	\begin{subfigure}[c]{0.32\textwidth}
		\centering
		\includegraphics[width=\textwidth]{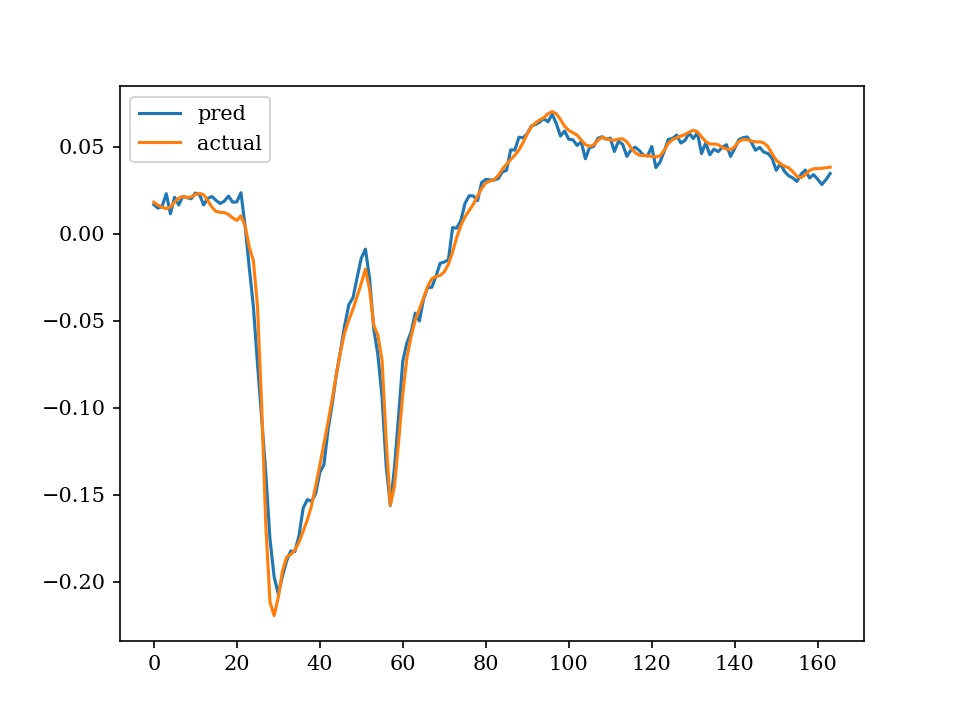}
		\caption{Train, 30ms interval}
	\end{subfigure}
	\hfill
	\begin{subfigure}[c]{0.33\textwidth}
		\centering
		\includegraphics[width=\textwidth]{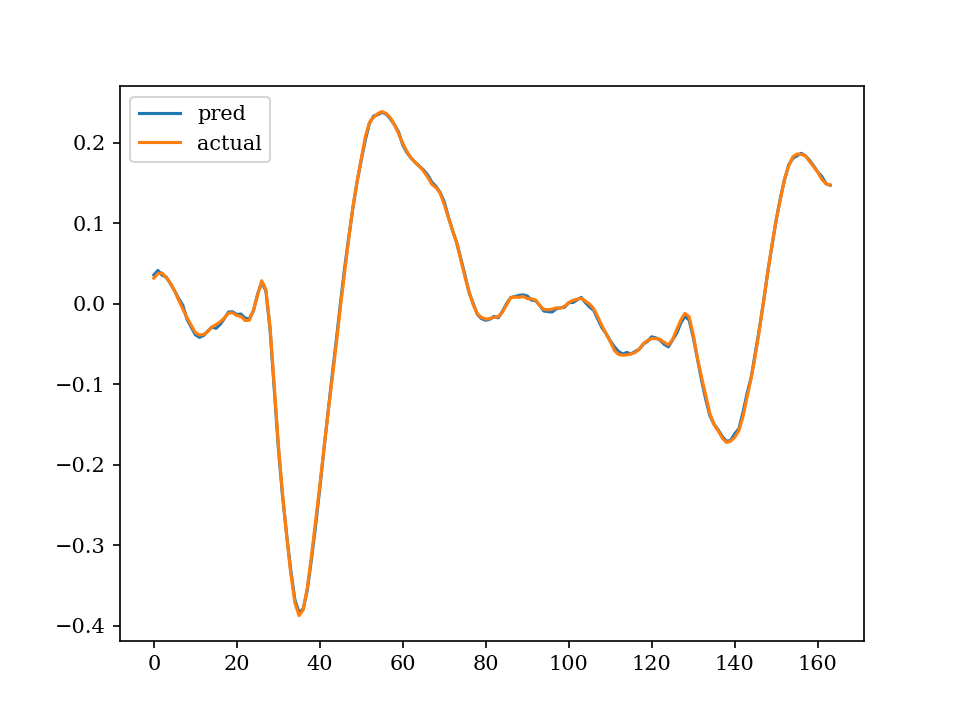}
		\caption{Train, 100ms interval}
	\end{subfigure}
	\hfill
    \begin{subfigure}[c]{0.33\textwidth}
		\centering
		\includegraphics[width=\textwidth]{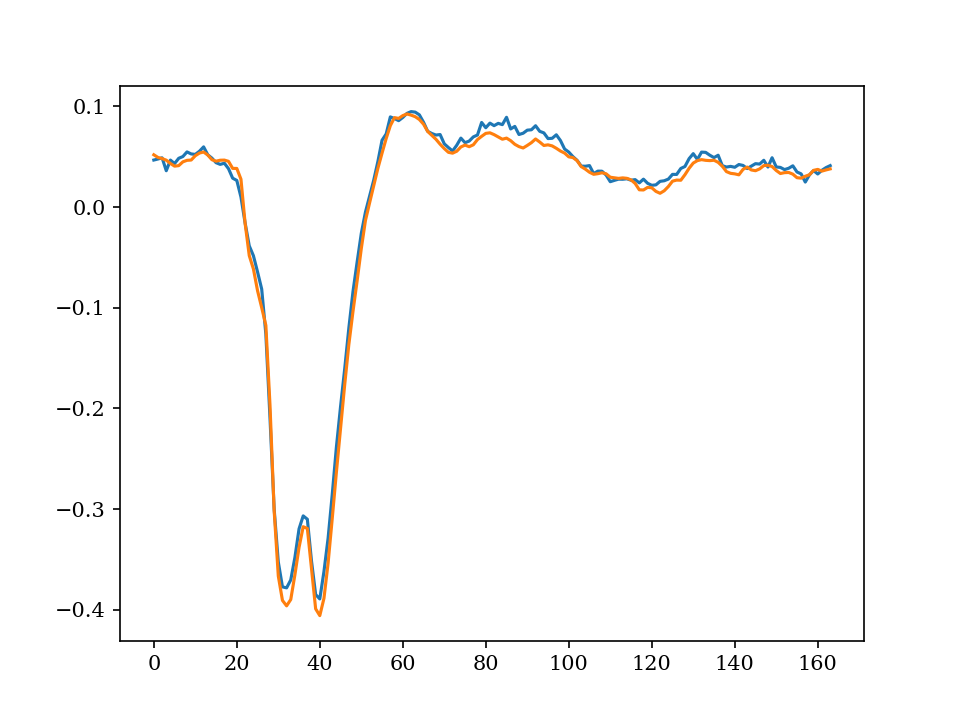}
		\caption{Test, 10ms interval}
	\end{subfigure}
	\hfill
    \begin{subfigure}[c]{0.32\textwidth}
		\centering
		\includegraphics[width=\textwidth]{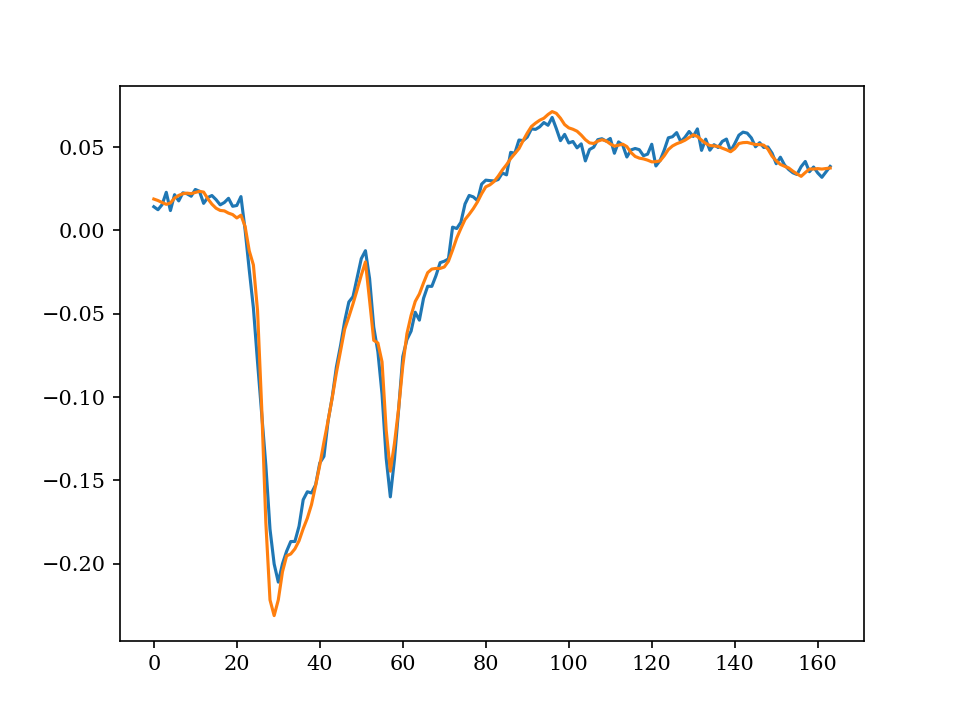}
		\caption{Test, 30ms interval}
	\end{subfigure}
	\hfill
	\begin{subfigure}[c]{0.33\textwidth}
		\centering
		\includegraphics[width=\textwidth]{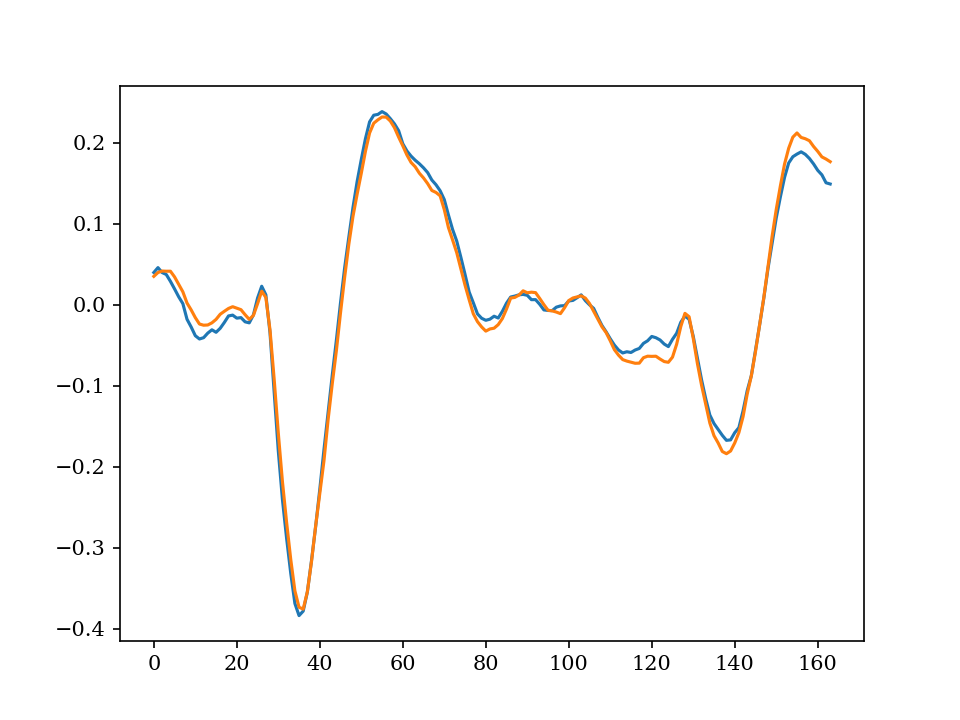}
		\caption{Test, 100ms interval}
	\end{subfigure}
	\hfill
\caption{\small \textbf{Mean predicted versus actual, single temporal basis function model (TBFM) with multiple stimulation parameters} Example sessions and channels, trial averaged. }
\label{fig:multisession_preds}
\end{figure}

\FloatBarrier
\newpage

\subsection{Effect of non-stationarity on model performance, possibly due to plasticity}
\label{apx:crossval}
The dataset was originally collected to study neural plasticity resulting from paired pulse stimulation and indeed, such plasticity can be detected in the dataset \cite{bloch.opto}. To evaluate the effect of plasticity on our model's performance, we performed random shuffling and five-fold cross validation of trials within each 2-hour session. With high probability, this results in the training and test datasets both having trials drawn from throughout the session, thereby controlling for the effect of plasticity that may have occurred between the beginning and the end of the session. We hypothesize the effect of plasticity and other sources of non-stationarity to be two-fold: 1) training set performance should be lower on the cross-validation experiment compared to our original experiment since a single model will need to generalize across a wider variety of trials; and 2) the difference between train and test set performances should be smaller on the cross validation experiment due to the training set being more similar to the test set.

We see precisely those effects, but the differences are not always statistically significant under the two-sided two-sample t-test on the time domain data. Cross validation yielded an average $R^2$ of 0.515 (stdev $0.180$) on the training set, compared to $0.533$ (stdev $0.173$) without (P-value: $0.566$). On the test set we had 0.482 (stdev $0.179$) compared to $0.462$ (stdev $0.207$; P-value: $0.578$). Model performance on bandpassed and time-frequency domain data showed the same relationships. However, beta bandpassed and high gamma bandpassed data do show statistically significant differences (Table \ref{table:cross}, \ref{apx:filtering} outlines the filtering methods). As a result, any effects of plasticity or other sources of non-stationarity may exist but are small enough that they are only sporadically detectable statistically.

We hypothesize that this difficulty detecting the differences is due to two effects. First, the changes due to plasticity were somewhat small, and most detectable in the coherence of specific frequency bands \cite{bloch.opto}, and in no case do we attempt to predict coherence changes. As shown in Table \ref{table:cross} we do see lower P values on time-frequency domain data, but the difference is once again not statistically significant. This leads to the second hypothesized effect: with N=$40$ our statistical test is likely underpowered to detect the relatively small effect of plasticity.

\begin{figure}
	\centering
	\begin{subfigure}[c]{0.49\textwidth}
		\centering
		\includegraphics[width=\textwidth]{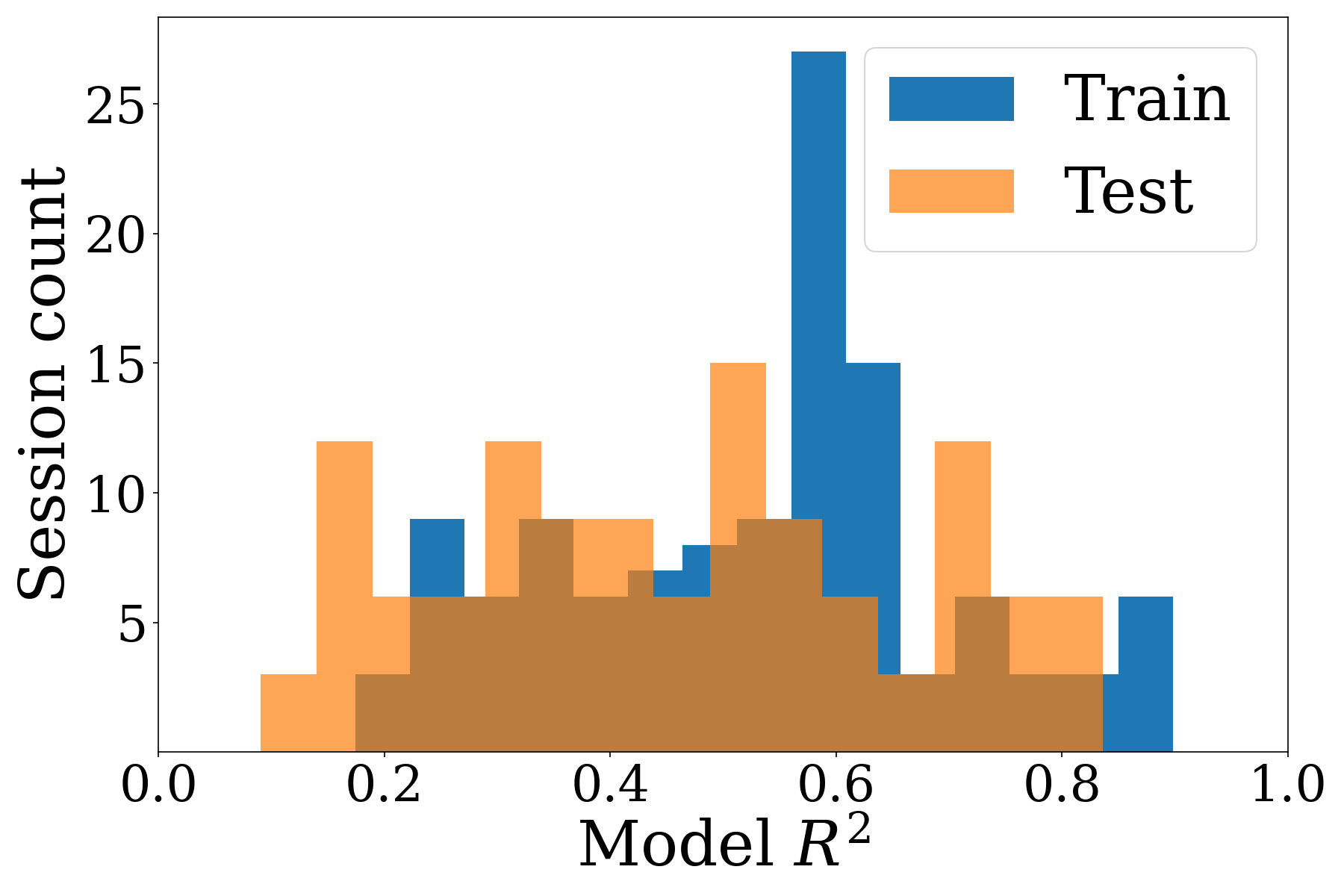}
		\caption{}
	\end{subfigure}
	\hfill
	\begin{subfigure}[c]{0.50\textwidth}
		\centering
		\includegraphics[width=\textwidth]{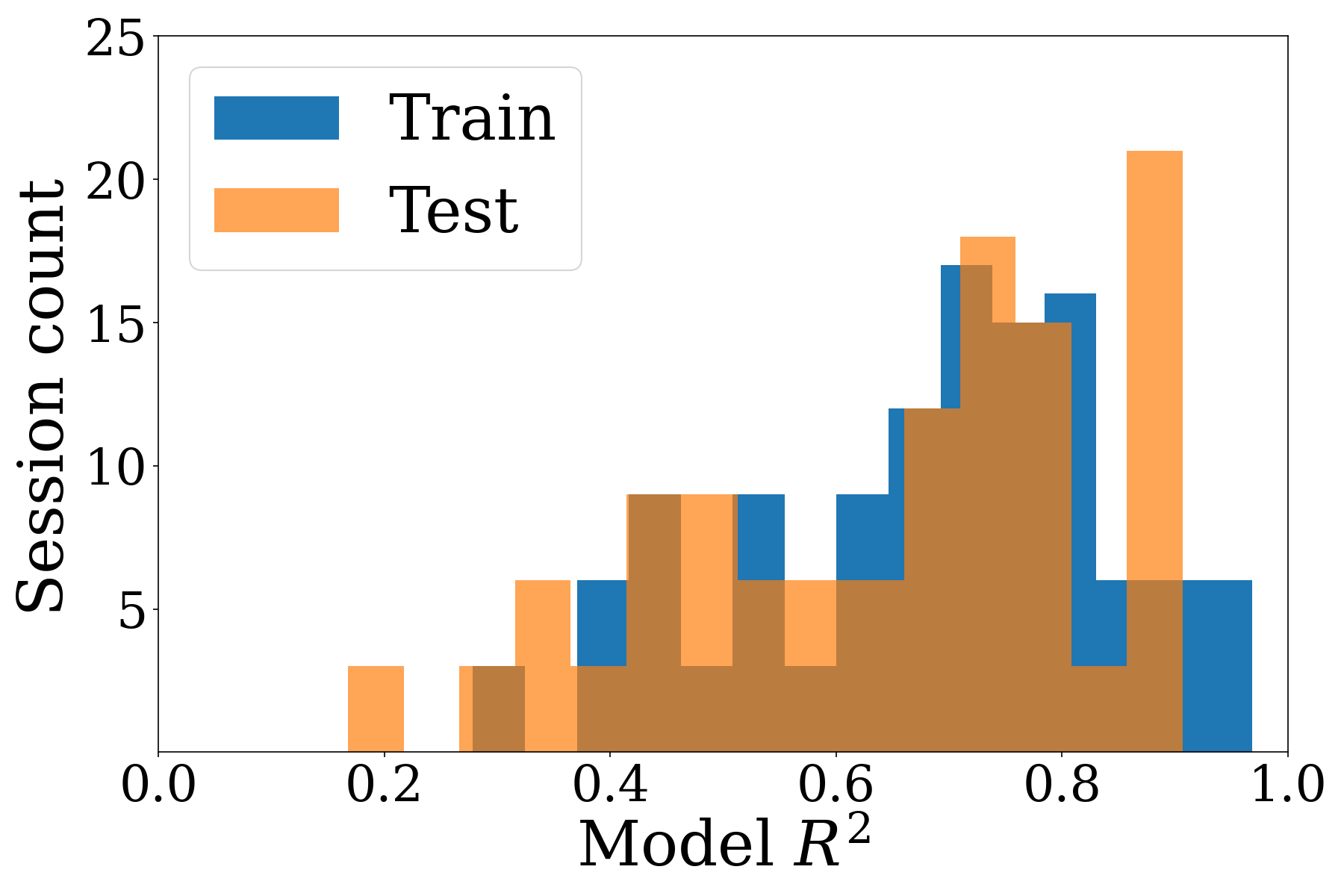}
		\caption{}
	\end{subfigure}
	\hfill
  	\begin{subfigure}[c]{0.50\textwidth}
		\centering
		\includegraphics[width=\textwidth]{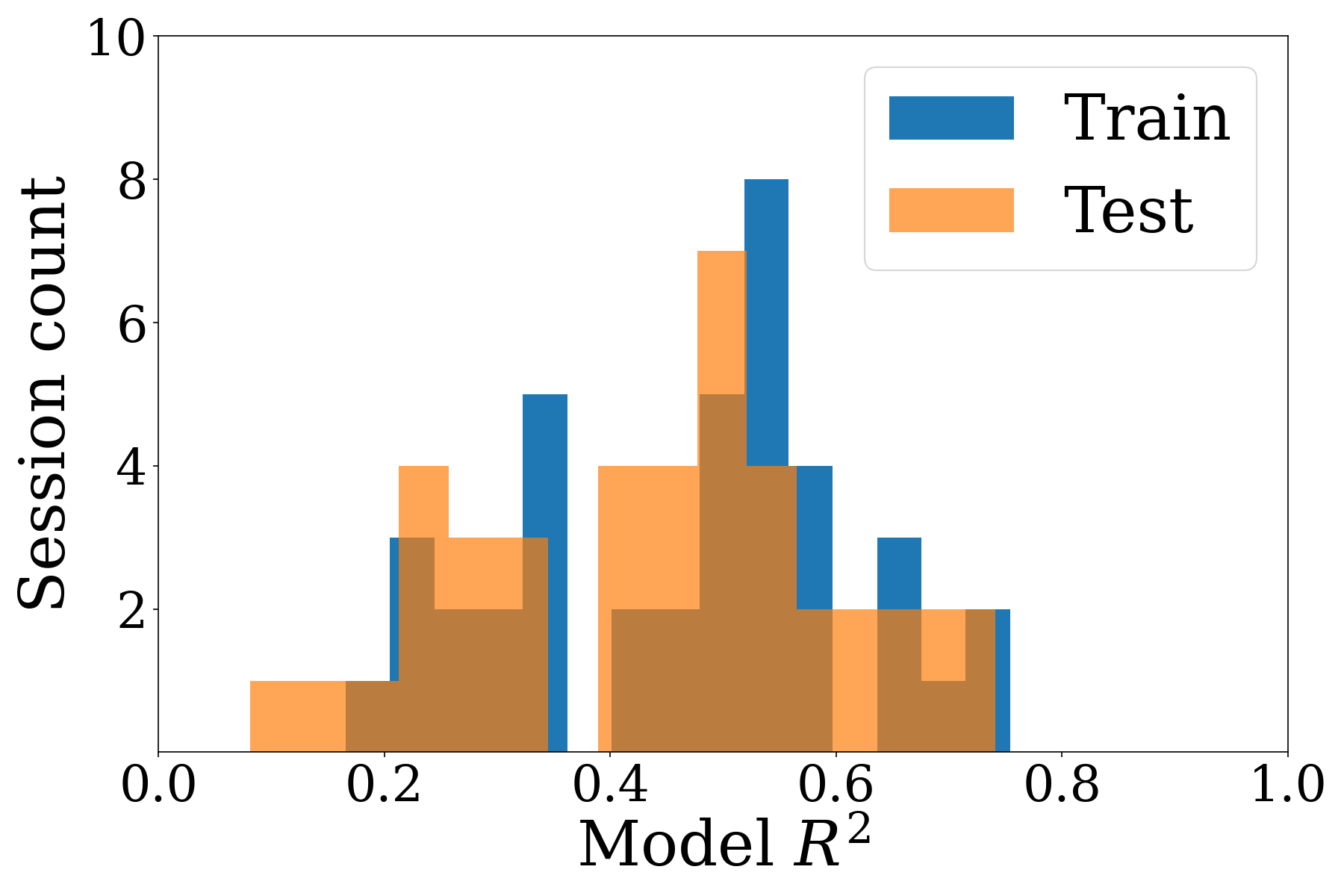}
		\caption{}
	\end{subfigure}
	\hfill
\caption{\small \textbf{Cross validation results on time series data (a) $R^2$ train versus test sets} Results for
3 runs per session, 40 sessions. Mean $R^2$ on train set 0.533 ($\pm0.173$). Mean $R^2$ on test set 0.462 ($\pm0.207$).
\textbf{(b) $R^2$ 60ms forward prediction} Results rise marginally if we
forward predict over a shorter time horizon - just beyond the typical effect of the first stimulation pulse: 0.682 ($\pm$0.162) and 0.648 ($\pm$0.190) for the train and test sets, respectively.
\textbf{(c) $R^2$ 5 fold cross validation} To control for the
effects of plasticity we can perform random shuffling and cross
validation. Here we have 0.515 ($\pm$0.180) and 0.482 ($\pm$0.179) on
the train and test sets respectively. This difference is not
statistically significant under the two-sided two-sample t-test:
p=$0.574$, p=$0.557$ on the train and test sets, respectively.}
\label{fig:crossr2}
\end{figure}

See Figure \ref{fig:crossr2} for time domain results. See Table \ref{table:cross} for a summary of results across filters.

\begin{table}[]
\begin{tabular}{|l|l|l|l|}
\hline
\textbf{Data set}     & \textbf{$R^2$ (stdev)} & \textbf{Cross validation $R^2$} & \textbf{P-value} \\ \hline
Time domain (train)   & 0.533 (0.173)       & 0.515 (0.180)                & 0.57             \\ \hline
Time domain (test)    & 0.462 (0.207)       & 0.482 (0.179)                & 0.56             \\ \hline
High gamma BP (train) & 0.135 (0.059)       & 0.072 (0.031)                & \textless{}0.01 \\ \hline
High gamma BP (test)  & 0.000 (0.082)       & 0.036 (0.028)                & 0.01            \\ \hline
Beta BP (train)       & 0.211 (0.051)       & 0.164 (0.029)                & \textless{}0.01 \\ \hline
Beta BP (test)        & 0.114 (0.087)       & 0.135 (0.030)                & 0.15           \\ \hline
Beta PSD (train)      & 0.459 (0.182)       & 0.423 (0.155)                & 0.26             \\ \hline
Beta PSD (test)       & 0.373 (0.176)       & 0.405 (0.154)                & 0.31             \\ \hline
Gamma PSD (train)     & 0.396 (0.195)       & 0.355 (0.161)                & 0.23             \\ \hline
Gamma PSD (test)      & 0.282 (0.183)       & 0.332 (0.159)                & 0.12             \\ \hline
\end{tabular}
\caption{Summary of cross validation results}
\label{table:cross}
\end{table}

\FloatBarrier
\newpage
\subsection{Example basis functions}
\label{apx:basis}

\begin{figure}[h!]
	\centering
  	\begin{subfigure}[c]{0.89\textwidth}
		\centering
		\includegraphics[width=\textwidth]{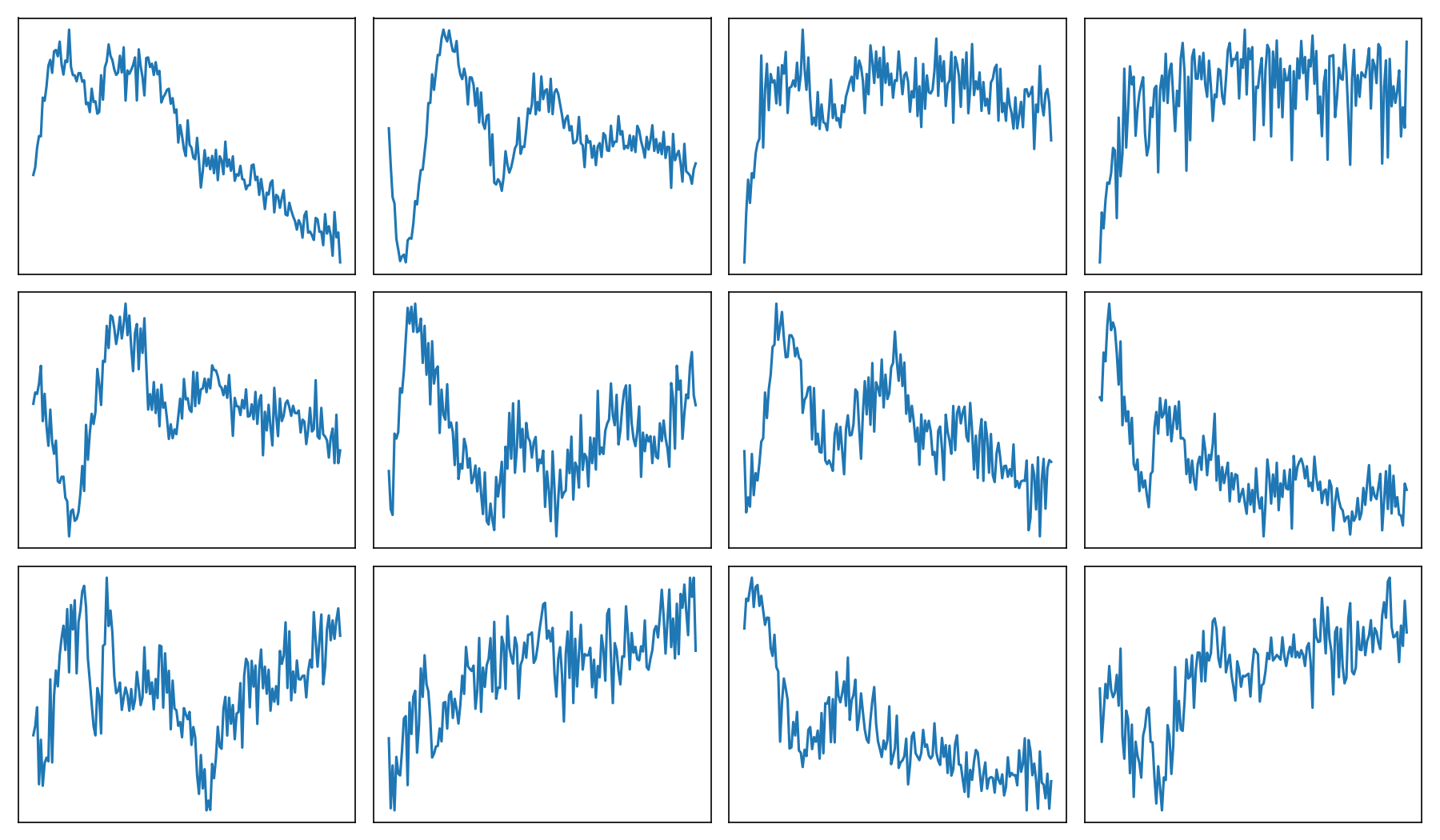}
	\end{subfigure}
	\hfill
\caption{Learned basis functions from an example model}
\label{fig:examplebases}
\end{figure}

\FloatBarrier
\subsection{Example additive basis functions}
\label{apx:additivebasis}
We found FSAM to require $\approx$1hr for training up to 15 bases on a given session. Since it is significantly slower than our simpler training method we may not want to apply it to all sessions. Instead it may be used on a handful of sessions to determine the basis count $b$ for future sessions.

To perform FSAM, we update our basis generator to be composed of $b$ separate MLPs, i.e., one sub-model for each basis. We continue to build the model using supervised learning, but on each learning step, we update only the sub-model of the basis we are presently learning. We update all parameters of our weight estimator on every learning step.

In one version of our proposal we instead attempted to build our bases using PCA and used supervised learning to predict the basis weights as we do now. However, we found that approach tended to result in unacceptably high overfit regardless of regularization used. See Appendix~\ref{apx:refpca} for details.

\begin{figure}[h!]
	\centering
  	\begin{subfigure}[c]{0.89\textwidth}
		\centering
		\includegraphics[width=\textwidth]{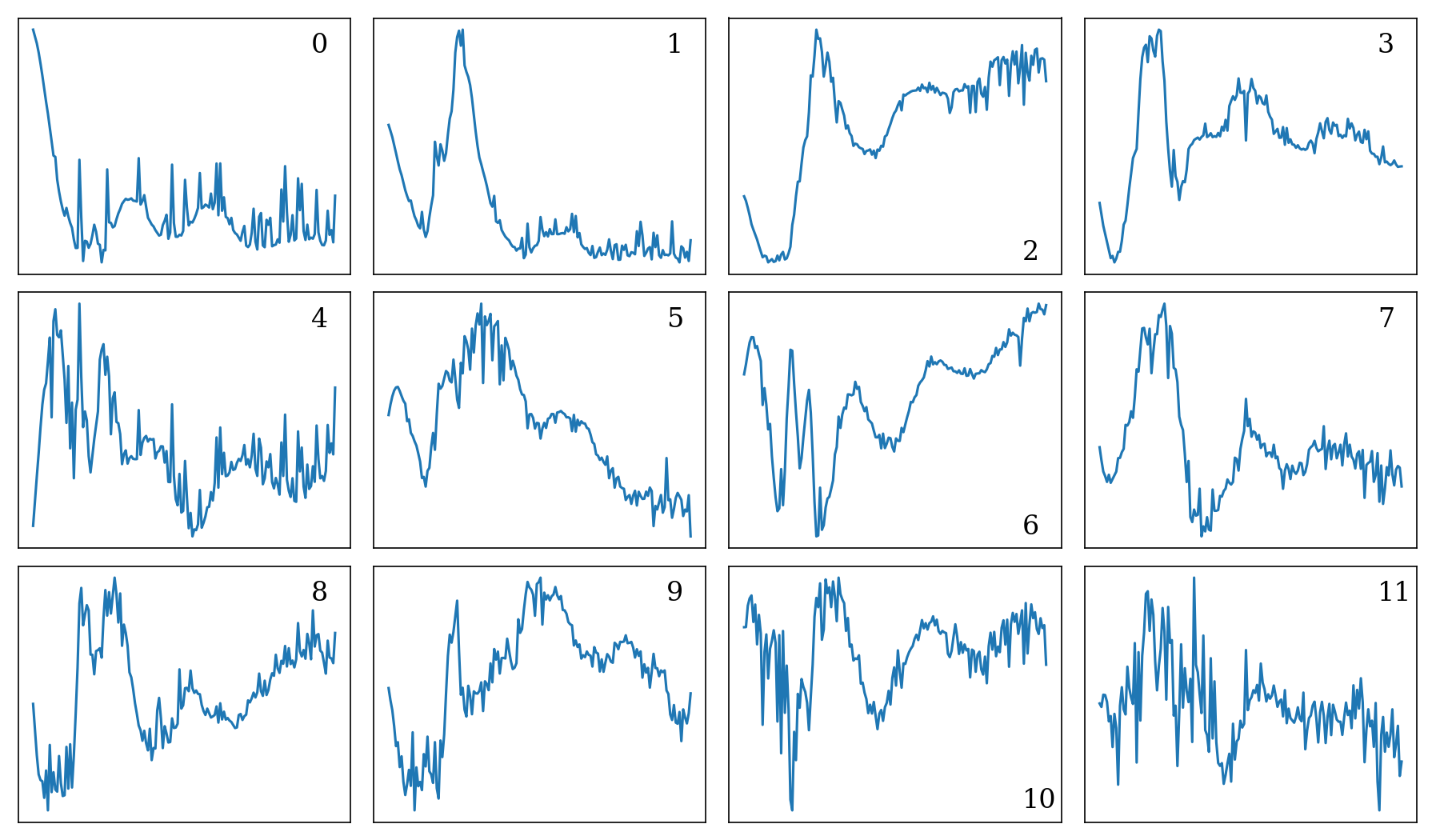}
	\end{subfigure}
	\hfill
\caption{Learned basis functions from an example additive model}
\label{fig:examplebasesadditive}
\end{figure}

As described in Section \ref{sec:results.additive} FSAM can result in some amount of interpretability of the basis functions. Detailed examples are depicted in Figure \ref{fig:examplebasesadditive}.

\FloatBarrier
\subsection{Example trials from Demonstrations 1, 2}
\label{apx:demoexamples}
\begin{figure}[h!]
	\centering
    \includegraphics[width=\textwidth]{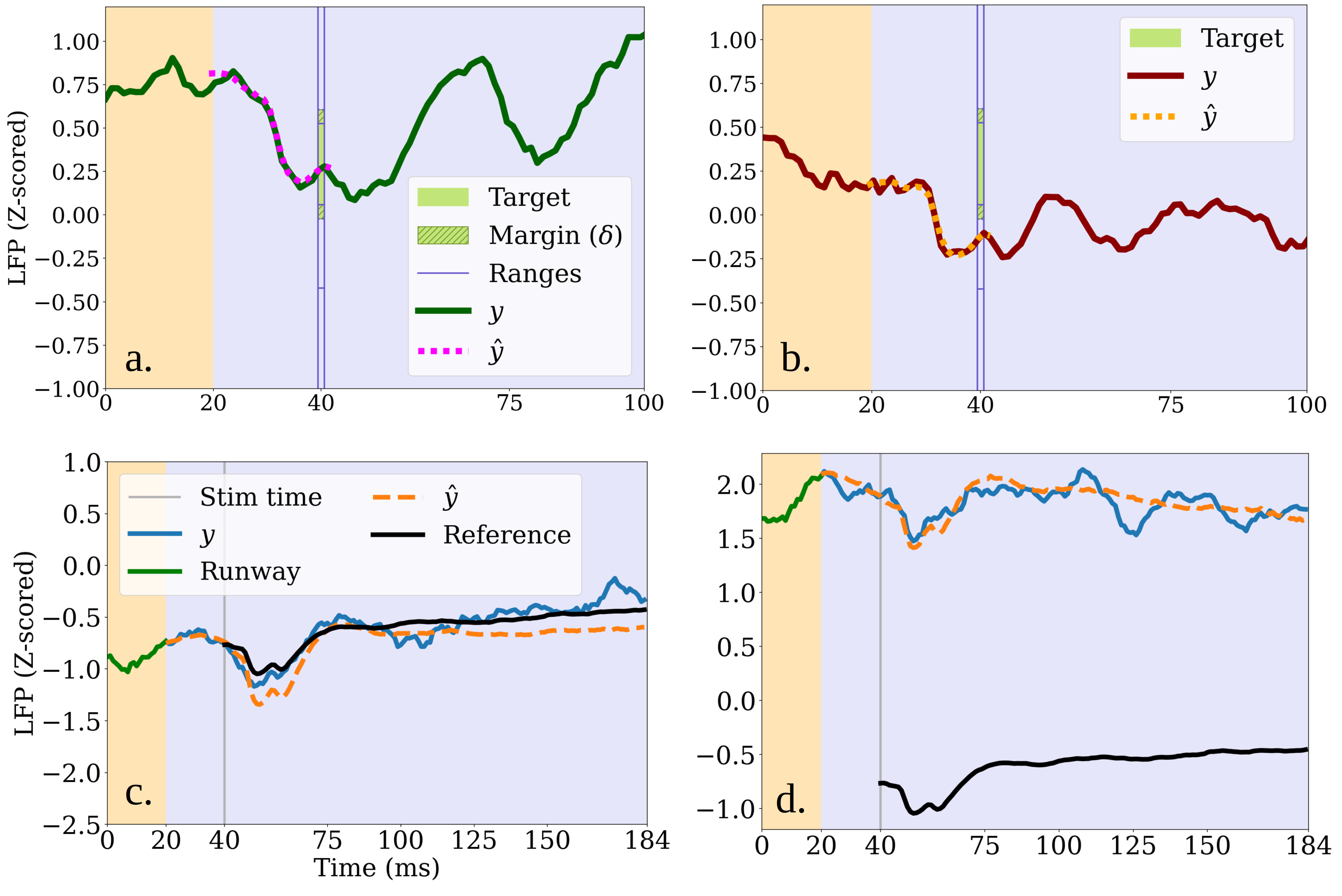}
	\hfill
\caption{\textbf{Example trials from Demonstrations 1, 2} (a) Demo 1, true positive. (b) Demo 1, true negative. (c) Demo 2, true positive. (d) Demo 2, true negative.}
\label{fig:demoexamples}
\end{figure}

See Figure \ref{fig:demoexamples}.

\end{document}